\documentclass{article} 
\usepackage[final]{colm2026_conference}

\usepackage{microtype}
\usepackage{hyperref}
\usepackage{url}
\usepackage{booktabs}
\usepackage{amsfonts}
\usepackage{nicefrac}
\usepackage[table]{xcolor}  
\usepackage{graphicx}
\usepackage{float}
\usepackage{makecell}
\usepackage{amsmath}
\usepackage{subcaption}
\usepackage{multirow}
\usepackage[normalem]{ulem}
\usepackage{wrapfig}
\usepackage{tablefootnote}

\usepackage{lineno}

\definecolor{darkblue}{rgb}{0, 0, 0.5}
\hypersetup{colorlinks=true, citecolor=darkblue, linkcolor=darkblue, urlcolor=darkblue}

\title{Judge a Book by Its Cover: Investigating Multi-Modal LLMs for Multi-Page Handwritten Document Transcription}

\author{Benjamin Gutteridge$^{1,2,*}$ \quad Matthew Jackson$^{1}$ \quad Toni Kukurin$^{2}$ \quad Xiaowen Dong$^{1}$ \\[2pt]
{\normalfont $^{1}$University of Oxford \quad $^{2}$QuantCo \quad $^{*}$Correspondence: \texttt{beng@robots.ox.ac.uk}}
}

\newcommand{\mytilde}{\raise.17ex\hbox{$\scriptstyle\mathtt{\sim}$}}
\newcommand{\bentham}{\texttt{Bentham}}
\newcommand{\iam}{\texttt{IAM}}
\newcommand{\mhills}{\texttt{Malvern-Hills}}
\newcommand{\firstpage}{\textsc{ocr+page1}}
\newcommand{\chosenpage}{\textsc{ocr+pageN}}
\newcommand{\vision}{\textsc{vision}}
\newcommand{\pbp}{\textsc{-pbp}}
\newcommand{\allpages}{\textsc{ocr+all-pages}}
\newcommand{\allocr}{\textsc{all-ocr}}
\newcommand{\ocr}{\textsc{ocr}}

\newcommand{\iamp}{\texttt{IAM-5}}
\newcommand{\iamr}{\texttt{IAM-5-Random}}
\newcommand{\mhillsp}{\texttt{Malvern-Hills-5+}}
\newcommand{\mhillspp}{\texttt{Malvern-Hills-10+}}
\newcommand{\casia}{\texttt{CASIA-5}}
\newcommand{\randompage}{\textsc{ocr+pageR}}
\newcommand{\pagex}{\textsc{ocr+pageX}}
\newcommand{\pylaia}{\textsc{pylaia}}
\newcommand{\trocr}{\textsc{trocr}}
\newcommand{\docowl}{\textsc{docowl2}}

\begin{document}

\maketitle

\begin{abstract}
Handwriting text recognition (HTR) remains a challenging task. Existing approaches require fine-tuning on labeled data, which is impractical to obtain for real-world problems, or rely on zero-shot tools such as OCR engines and multi-modal LLMs (MLLMs). MLLMs have shown promise both as end-to-end transcribers and as OCR post-processors, but to date there is little empirical research evaluating different MLLM prompting strategies for HTR, particularly for the case of \textit{multi-page documents}. Most handwritten documents are multi-page, and share context such as semantic content and handwriting style across pages, yet MLLMs are typically used for transcription at the page level, meaning they throw away this shared context. They are also typically used as either text-only post-processors or image-only OCR alternatives, rather than leveraging multiple modes.
This paper investigates a suite of methods combining OCR, LLM post-processing and MLLM end-to-end transcription, for the task of zero-shot multi-page handwritten document transcription.
We introduce a benchmark for this task from existing single-page datasets, including a new dataset, \mhills{}. Finally, we introduce \firstpage{} and \chosenpage{}, prompting strategies for multi-page transcription that outperform existing methods by sharing content across pages while minimizing prompt complexity.
\end{abstract}

\section{Introduction}

A significant proportion of human written material exists only as physical, handwritten documents. Accurate, cost-effective digitization of such documents would benefit many fields by improving information accessibility and ease of processing. Digitized handwriting could also provide a largely-untapped source of training data for future models.

While modern optical character recognition (OCR) software is now adept at transcribing machine-printed text, handwritten text recognition (HTR) remains challenging; handwritten documents are often multiple pages long, extremely noisy, and can vary enormously in handwriting style and document structure.
State-of-the-art HTR models \citep{li2023trocr, fujitake2024dtrocr} typically combine pre-trained vision Transformers \citep{dosovitskiy2020image, liu2021swin, huang2022layoutlmv3, xu2020layoutlm, kim2021donut} and (small) language models \citep{devlin2018bert, liu2019roberta}, and rely on fine-tuning with labeled data to perform well. Unfortunately, labeling this training data, i.e. manually transcribing documents, is usually too expensive and time-consuming to be practical in the real world. For this reason, we are interested in models that can be deployed \textit{zero-shot} --- without training/fine-tuning.

There are several zero-shot OCR tools for handwriting, such as Google's Vision API, Amazon's Textract, and Transkribus \citep{kahle2017transkribus, nockels2022understanding}. These can perform reasonably well, and are fairly cheap, but still frequently return noisy outputs. Furthermore, they operate at (at most) the page-level. HTR research and benchmarking is primarily focused on the character-, word-, line- or page-level, despite the fact that most real-world handwritten documents are \textit{multi-page}. This means that (after being scanned/photographed) a document is split over multiple images, which are \textit{highly inter-related}: handwriting patterns and quirks, structure and formatting, visual artifacts, and, of course, semantic text content. All of this shared context goes unused by the models and tools discussed above, as they process documents one page at a time.

Large language models (LLMs; \citet{floridi2020gpt, achiam2023gpt, zhao2023survey}) show promise for addressing these challenges. One example of this is using LLMs as a \textit{post-processor} for correcting cheap and noisy OCR output; we discuss this in Section~\ref{sec:related-work} and include experimental results in Section~\ref{sec:experiments}. Furthermore, by leveraging the long context capabilities of modern LLMs \citep{chen2023longlora, liu2023lost, kim2024fables, karpinska2024one}, this can even be done all-at-once for multi-page documents, theoretically enabling shared inter-page context to inform corrections.

Even more promising are multi-modal LLMs (MLLMs; \citet{wu2023multimodal}), LLMs that can accept both text and images as prompts. Despite being general models and not trained explicitly for HTR, commercial MLLMs such as \textsc{gpt-4o} are \textit{very} good at end-to-end handwriting transcription, often much better than OCR tools designed for the purpose (see Section~\ref{sec:experiments}). The main downsides of end-to-end transcription with MLLMs are (i) the risk of hallucination; an OCR engine may produce a noisy, erroneous signal, but it is less likely than an MLLM to corrupt a signal by introducing text that is `reasonable' but incorrect, and therefore more difficult to identify; and (ii) expense; the best commercial LLMs can be expensive at scale --- images consume many more tokens than raw text, and transcription is a task that produces a lot of output tokens, which are typically priced higher than input tokens. Using MLLMs as postprocessors rather than end-to-end can partially mitigate these issues, but without access to images they tend to underperform. 

\textbf{This paper} aims to address a lack of empirical research into the comparison and combination of OCR tools and MLLMs for HTR, specifically in the zero-shot, multi-page document setting; the setting most frequently faced by practitioners. 
The contributions of this paper are three-fold:
\begin{description}
    \item We investigate the capabilities of OCR, MLLMs, and combinations of both for \textit{zero-shot, multi-page, handwritten document transcription}, evaluating a suite of methods on documents ranging from 2 to 13 pages in length.
    \item We introduce \mhills{}, a new multi-page handwritten document dataset derived from previously undigitised archival sources, and complement it with three further multi-page datasets synthesized from existing single-page collections (including \casia{}, a Chinese handwriting dataset), forming a diverse four-task benchmark for multi-page transcription.\footnote{\small{Code \&{} datasets: \url{https://github.com/BenGutteridge/judge-a-book-by-its-cover/}}}
    \item We propose \firstpage{} and \chosenpage{}, simple but effective methods permitting MLLMs to extrapolate information from \textit{a single page image} to improve the transcription accuracy of OCR-generated text, leveraging multi-modality and shared inter-page context to improve transcription accuracy while balancing prompt complexity and token cost.
\end{description}

\section{Related work}
\label{sec:related-work}
\paragraph{Handwriting OCR.} Most OCR engines, including those that can be run locally like Tesseract, are designed for use with printed text and are nearly useless for handwriting.
Several commercial OCR engines, such as Google Cloud Vision, Azure Vision, Amazon Textract and Transkribus \citep{kahle2017transkribus, nockels2022understanding} are designed for use on handwritten text at the page-level scale.

SOTA HTR and OCR models \citep{fujitake2024dtrocr, li2023trocr, kim2021donut, huang2022layoutlmv3} are typically based on pre-trained vision Transformers (ViT; \citet{vaswani2017attention, dosovitskiy2020image}) and may include recurrent components like LSTMs or CNNs \citep{breuel2013high, azawi2013normalizing, bora2020handwritten, yang2019handwriting}; the commercial handwriting-capable OCR engines mentioned above are likely similar in architecture to the best of these models, leveraging massive, pre-trained ViTs and language models. In general, such models are somewhat effective on HTR tasks zero-shot, and SOTA is reached by fine-tuning 
on a specific task.
This is fine for benchmarks, but for real-world tasks obtaining labeled training data is often prohibitively expensive. 
Furthermore, most benchmarks are concerned only with recognition at the character- or line-level. This can, of course, be aggregated to return document-level transcriptions, but this neglects the task of text \textit{detection}, and does not consider incidentals that occur in real documents --- headings, figures, scribbles, margin notes, imperfections in image quality, distractors, etc. 
This paper is concerned with transcription over multi-page documents in a holistic manner.

\paragraph{LLMs for OCR post-processing.}  Several works have investigated improving OCR transcription accuracy with post-processing by a language model \citep{lund2011progressive, schaefer2020two, veninga2024llms, rigaud2019icdar}. LLM-aided OCR is a public tool that uses OCR output with an LLM post-processor to improve OCR transcription accuracy, but the authors do not provide any experimental results demonstrating improvement besides hand-picked examples. 
Similarly, BetterOCR is a tool that combines results from multiple OCR engines and passes them into an LLM, but only hand-picked examples are provided as experimental results. Furthermore, both tools are only designed for printed text, both operate at the page level (or at the finer-grained `page chunk' level), and neither process images directly with MLLMs, only using LLMs for post-processing. 

Several existing works investigate the use of OCR output, including text and bounding boxes, alongside images  as input to MLLMs for document understanding \citep{wang2024docllm, luo2024layoutllm, liao2025doclayllm, wang2025marten}, but these works also focus on the single page case only, are concerned with tasks like question answering rather than transcription, and do not use handwritten data (with the exception of short mathematical expressions).

\paragraph{Benchmarks.} Most OCR benchmarks are for machine-printed text, and only for single pages/images \citep{liu2023hidden}, such as receipts \citep{park2019cord, huang2019icdar2019}). Kleister is a pair of multi-page, long-context key entity extraction benchmark tasks, but consists of only machine-printed text \citep{gralinski2020kleister, stanislawek2021kleister}.

There are a number of HTR benchmarks, including historical documents, documents not written in English or with Latin characters  \citep{sanchez2019set, zhang2019icdar, causer2018making, dolfing2020scribblelens, serrano2010rodrigo, wigington2018start, carbonell2019end, yu2021benchmarking}, and transcription of numerical digits or mathematical expressions \citep{liu2023hidden, yuan2022syntax, diem2014icfhr}. None are explicitly concerned with multi-page documents, and most are at the line- or word-level.

\section{\firstpage{} and \chosenpage{}: correction from a single page}
\begin{figure}[h]
    \centering
    \begin{minipage}[b]{0.5\linewidth}
        \centering
        \includegraphics[width=0.76\linewidth]{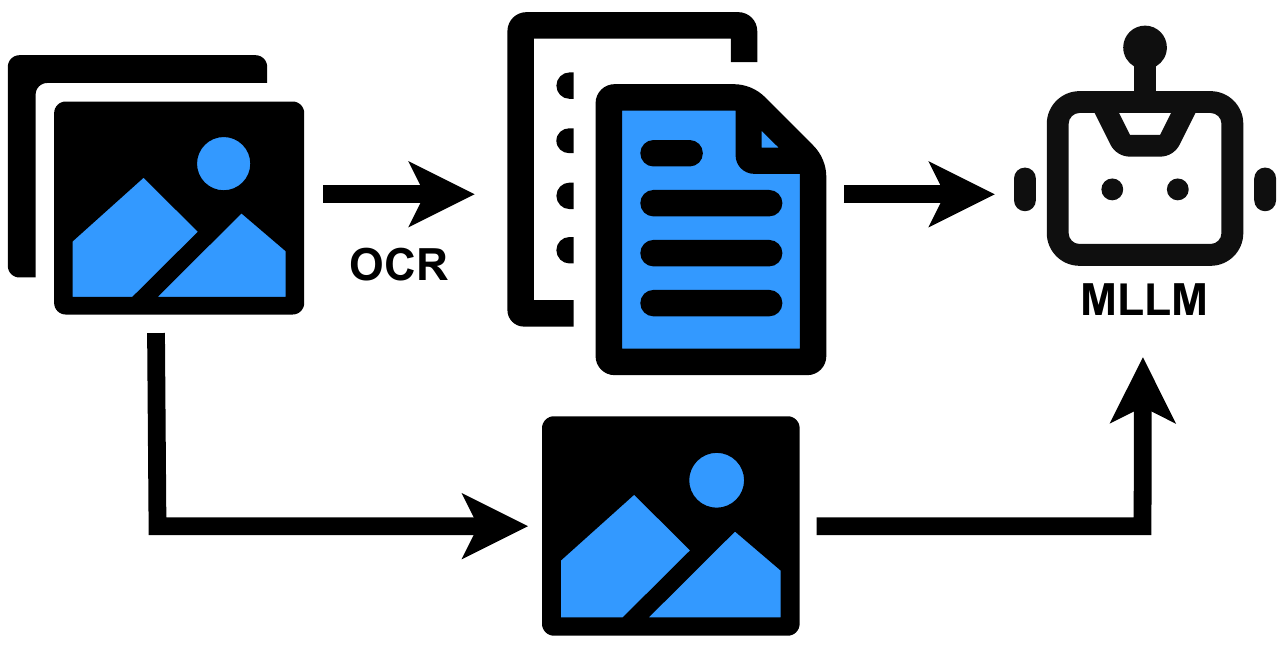}
        \subcaption{\firstpage{}}
        \label{fig:firstpage}
    \end{minipage}%
    \begin{minipage}[b]{0.5\linewidth}
        \centering
        \includegraphics[width=0.80\linewidth]{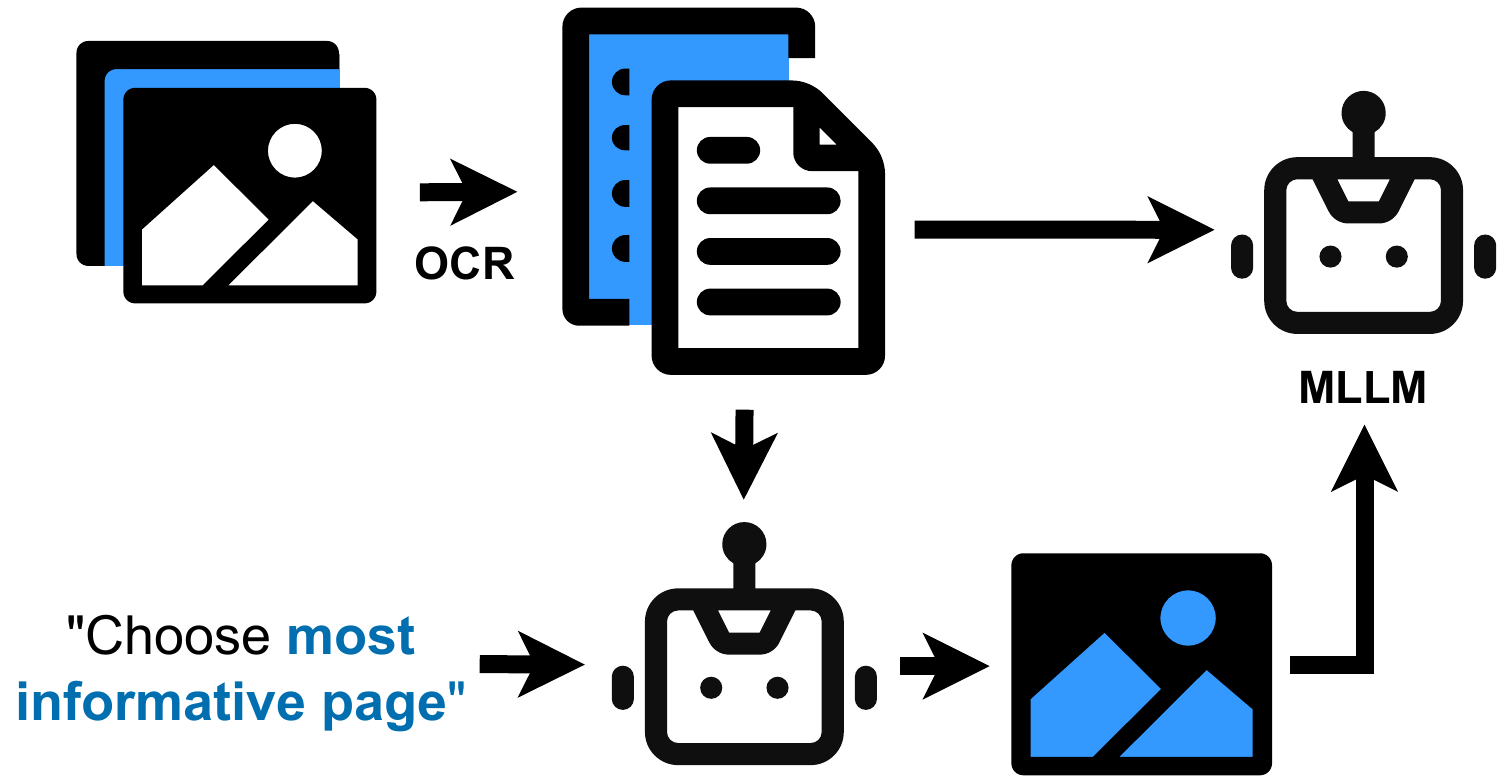}
        \subcaption{\chosenpage{}}
        \label{fig:chosenpage}
    \end{minipage}
    \caption{\small Illustrations of \pagex{} methods. In both, the full OCR text of the document is provided to an MLLM along with a single page image. \textsc{+page1} always uses the first page; \textsc{+pageN} uses the $N$th page, where $N$ is chosen by an upstream LLM given the OCR text. The page image/text passed to the MLLM are \textcolor[HTML]{3399FF}{highlighted}. For \textsc{+pageN}, in this case $N\!=\!2$, but it can be any page.}
    \label{fig:first-vs-chosen}
\end{figure}

We propose two instances of the following prompting strategy for OCR post-processing of multi-page documents: provide the MLLM with the \textit{OCR output for the entire document} as well as \textit{a single page image}. Figure~\ref{fig:first-vs-chosen} illustrates the two methods: \firstpage{} and \chosenpage{}.\footnote{We additionally evaluate \randompage{}, a variant of \chosenpage{} that provides a \textit{random} page image rather than one selected by an upstream LLM (see Appendix~\ref{sec:app:extended-experiments}).}

\firstpage{} is the simpler strategy: the page image chosen is always the first page of the document. For \chosenpage{}, a cheap LLM is prompted with the OCR text and asked to \textit{choose} the most promising page image to have access to, which may or may not be the first --- it should be clear from the OCR text whether a particular page has more or fewer errors, has a significant amount of reference text or relatively little, etc. \chosenpage{} adds a small amount of prompt complexity, but can reduce failures caused by unsuitable first pages, such as title pages with limited text. We would expect \chosenpage{} to perform at least as well as \firstpage{}, as \chosenpage{} can always fall back to \firstpage{} by default.

\paragraph{Motivation.} \firstpage{} and \chosenpage{} (collectively \pagex{}) are based on two assumptions. First, that multi-page documents are likely to be similar in many ways: handwriting is likely consistent for a single document, the semantic content is likely to be highly inter-related, and images of the original documents were likely obtained in a similar fashion (smartphone camera, scanner,  etc.) so image artifacts are likely similar across pages as well. We believe this is a reasonable assumption, indeed, a `multi-page document' without at least some shared traits across pages is better  characterized \textit{not} as a multi-page document, but a series of individual documents. The second assumption, which follows from the first, is that OCR errors are likely to be repeated across such similar pages. If a page contains a reasonable amount of text, it likely contains examples of many, if not most, handwriting quirks of the writer; after all, the most common 25 (100) words make up about one third (half) of all written English \citep{kress2015reading}, there are only 52 letter characters, and character combinations often repeat. Furthermore, more challenging words such as proper nouns (e.g. names, places) are reasonably likely to repeat across pages.

We stress that it is not necessary that \textit{all} OCR errors be present in a single page --- difficult names may appear once, an individual page may be damaged, etc. ---  our methods are based on the idea that \textbf{a single page provides more useful information than the second}, which provides more useful information than the third, and so on. A lot of page images means \textit{redundant information}, meaning more tokens, more expense and more prompt complexity. 

\begin{wrapfigure}{r}{0.5\textwidth}
    \centering
    \includegraphics[width=\linewidth]{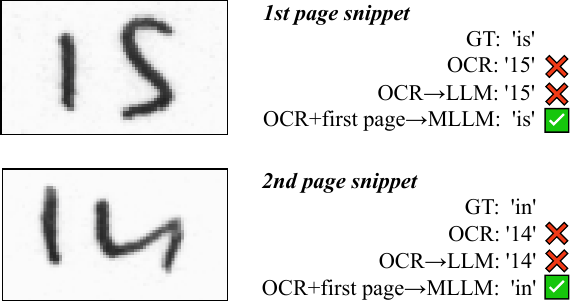}
    \caption{\small{Example of \firstpage{} propagating OCR error corrections across pages. Though the MLLM only has access to the first page image, it can use observed `i'$\rightarrow$1 \& word$\rightarrow$number errors to correctly transcribe `in' on the unseen second page.}}
    \label{fig:example_14_15}
\end{wrapfigure}

With these assumptions, we hypothesize that, with \firstpage{} or \chosenpage{}, an MLLM should be able to pick up on cues, from the provided image, about the relevant part of the noisy OCR input, and use this to improve on the post-processing of the entire text. Furthermore, as the OCR transcription for all pages is provided along with an image, semantic content can be shared between pages. This should provide a reasonable trade-off between useful inter-page information and prompt complexity/cost.

Figure~\ref{fig:example_14_15} illustrates a real example of \firstpage{} working in practice on the \iam{} dataset (see Section~\ref{sec:benchmark+exp-details}); see Appendix~\ref{sec:firstpage_examples} (Figures~\ref{fig:example_draws}--\ref{fig:example_columns}) for further examples.

We note a loose conceptual parallel to few-shot prompting \citep{dong2024survey}, in which one or more examples of expected input and desired output are provided within the prompt. The analogy is not exact, but we believe it aids intuition. The two input modalities for \firstpage{} and \chosenpage{}, the OCR text and page image, can be thought of as similar to the example input and target. In this case, however, rather than learning in-context to \textit{replicate} the OCR engine's output, the MLLM should (i) exercise its own judgement to identify OCR errors, (ii) identify how the OCR engine's choices should be corrected, and (iii) extrapolate from this learned image and the corresponding OCR text mapping to the remainder of the OCR output (i.e. the `unseen' text). 

\subsection{A comprehensive suite of methods for evaluation}
\label{sec:comprehensive-suite}
One of the main objectives of this work is to investigate the capabilities of OCR and MLLMs for multi-page transcription by evaluating a comprehensive suite of methods and MLLMs. To our knowledge, no existing works compare and evaluate such a range of methods and prompting strategies for handwriting transcription and OCR post-processing with (M)LLMs. We provide an overview of the methods evaluated in this paper below.

\paragraph{`All-at-once' and `page-by-page' processing.} Although we are interested in multi-page documents, it is an essential part of the task that proper page breaks are maintained in the final transcription; for this reason, in our experiments we evaluate transcription accuracy at the page- rather than document-level, and aggregate. As a result, the methods described below can be classified into `page-by-page' (\textsc{pbp}), where each page is processed separately,  and `all-at-once' (\textsc{aao}), where the relevant inputs for all pages are provided, and the MLLM must return output containing a transcription of each page keyed to its given page ID. 

\begin{wrapfigure}[30]{r}{0.38\linewidth}
    \centering
    \includegraphics[width=\linewidth]{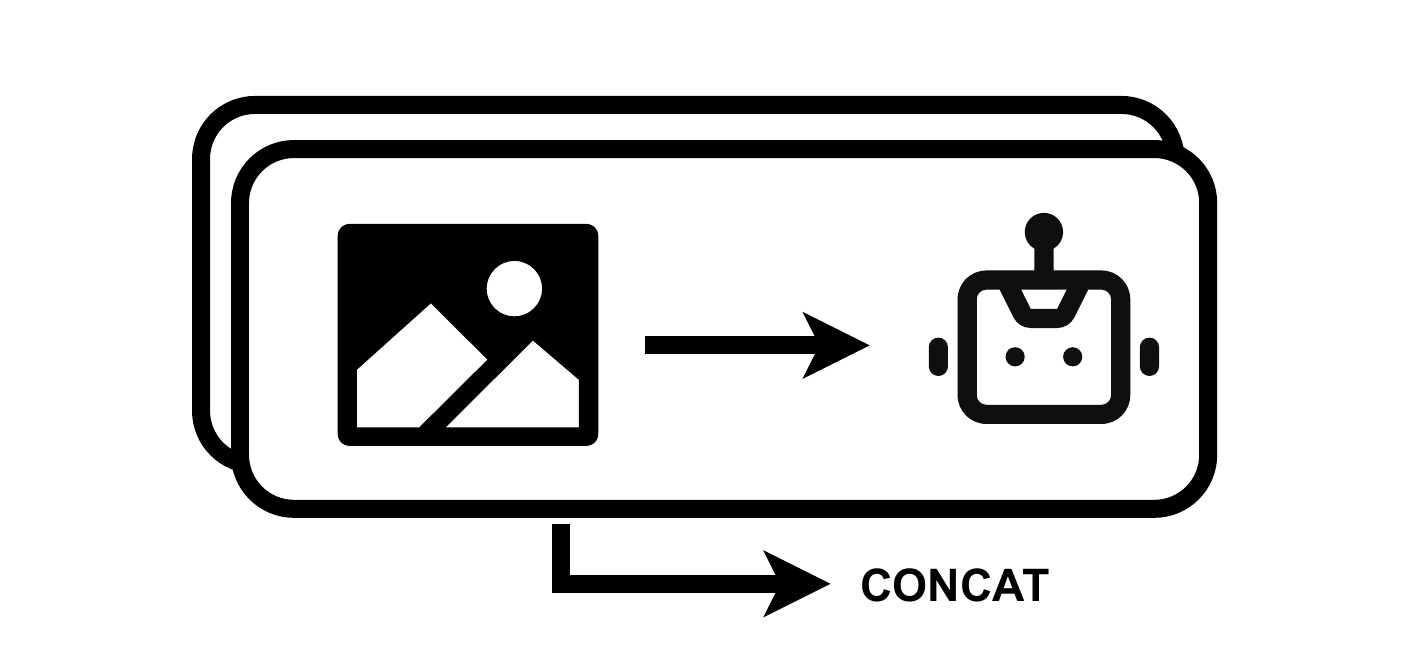}\\[-2pt]
    {\small (a)~\textsc{images $\rightarrow$ mllm}}\\[-2pt]
    \includegraphics[width=\linewidth]{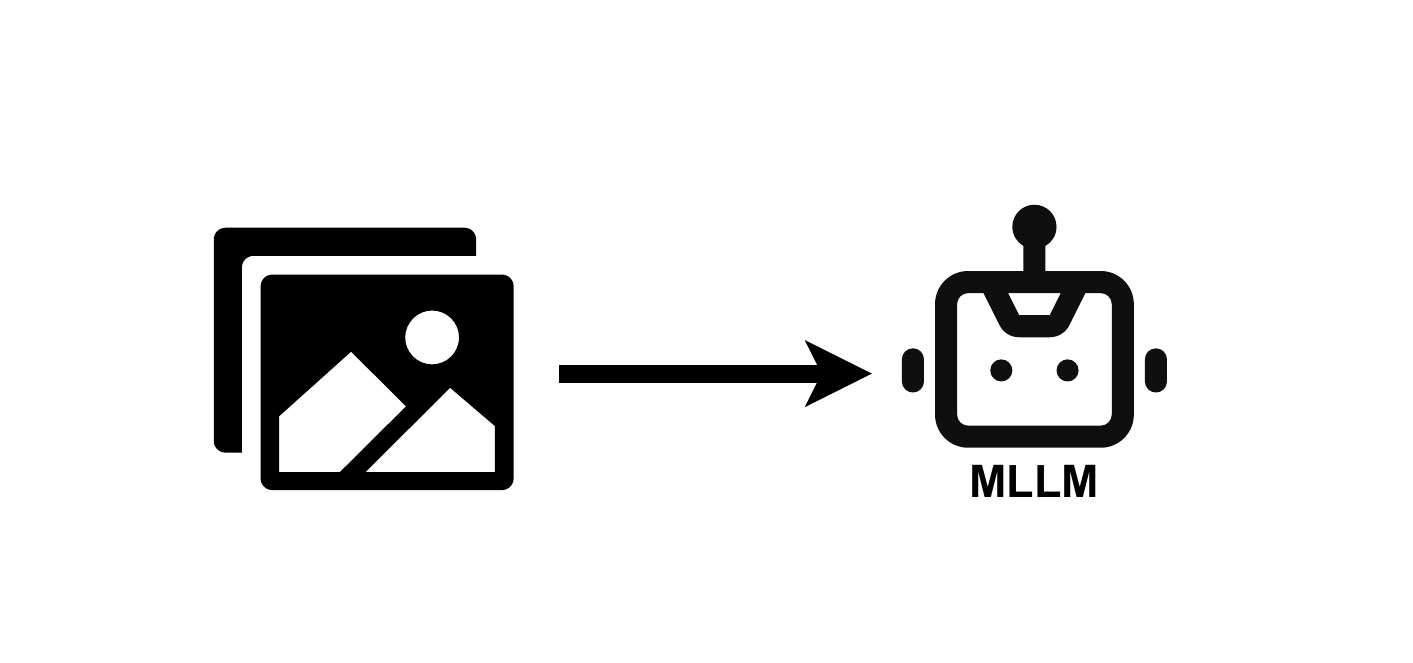}\\[-1pt]
    {\small (b)~\textsc{images-all-at-once $\rightarrow$ mllm}}\\[0pt]
    \includegraphics[width=\linewidth]{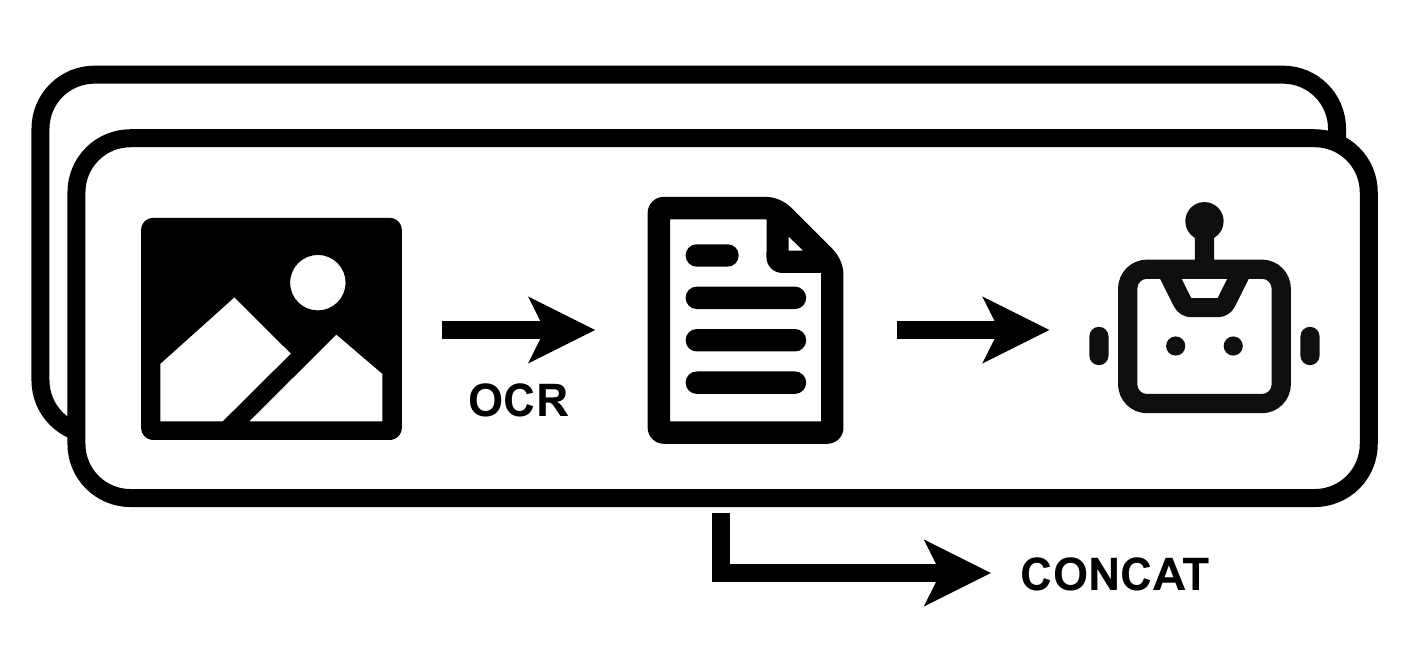}\\[-2pt]
    {\small (c)~\textsc{ocr $\rightarrow$ llm}}\\[3pt]
    \includegraphics[width=\linewidth]{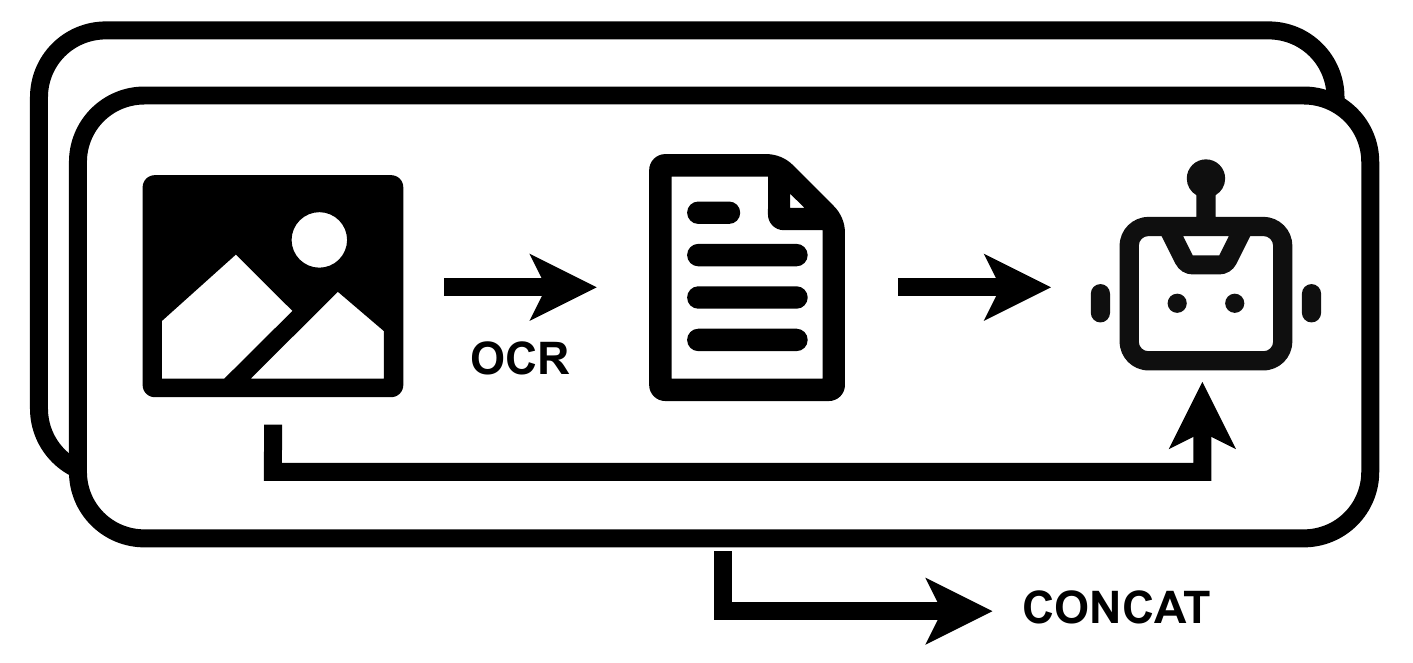}\\[-2pt]
    {\small (d)~\textsc{ocr+images $\rightarrow$ mllm}}
    \caption{\small MLLM-based transcription method illustrations.}
    \label{fig:grid_overview}
\end{wrapfigure}
For each method below we describe the \textit{input} to the (M)LLM that produces the final transcription. See Figures~\ref{fig:first-vs-chosen} \& \ref{fig:grid_overview} for illustrations.
\begin{description}
    \item[\textsc{images}] (\textsc{pbp}): image per-page; equivalent to using an MLLM for OCR end-to-end.
    \item[\textsc{images-all-at-once}] (\textsc{aao}): all images in the document; allows access to context across pages, but increases task difficulty and context length
    \item[\textsc{ocr}] (\textsc{pbp}): the output from an OCR engine for a given page; LLM has no access to the original image, but has a smaller token burden, and the potentially easier/more surgical task of \textit{correction} rather than full transcription.
    \item[\textsc{ocr+images}] (\textsc{pbp}): the page image \textit{and} the OCR transcription; we might expect this method to perform the best, as it has the most information, but it also consumes the most tokens.
    \item[\firstpage{}] (\textsc{aao}): the image of the \textit{first page} of the document, and input of the OCR transcription of every page (see Figure~\ref{fig:firstpage}); our hypothesis is that there is sufficient shared context between pages that a single page will permit effective, token-cheap OCR correction.
    \item[\chosenpage{}] (\textsc{aao}): the \textit{chosen page} image, and the OCR transcription of every page (see Figure~\ref{fig:chosenpage}). The chosen page ID is returned by an upstream (cheap) LLM call, asking, `given the OCR transcription of the full document, which page image would be most beneficial to a downstream LLM?' We might expect \hbox{\textsc{+pageN}} to be at least as good as \textsc{page1}.
\end{description}

\section{A benchmark for multi-page handwritten document transcription}
\label{sec:benchmark+exp-details}
In Section~\ref{sec:experiments} we evaluate the methods described above on three multi-page handwritten document datasets, including \mhills{}, a new dataset labeled by the authors and derived from public domain but not-previously-scanned documents obtained from a charity, the Malvern Hills Trust.
See Appendix~\ref{sec:app:example-docs} for example documents from each dataset.

\paragraph{\iam{}.}
The IAM Handwriting Database \citep{marti2002iam} is a handwriting benchmark of single pages, where each page contains a machine-typed passage and a handwritten copy of the same text beneath it, written by one of 657 English-speaking writers.
We crop the images to contain only the handwritten part (using provided metadata) and then combine them by writer ID to produce documents with consistent handwriting, and often related content.
We use a subset of 242 images to construct 107 multi-page documents, 79 with two pages and 28 with three, and henceforth refer to this multi-page dataset as \iam{}. We note that, though the multi-page documents produced are, in a sense, synthetic, the consistent handwriting means this is suitable for testing the cross-page extrapolation capabilities of our methods. Furthermore, many of the pages in the IAM Database are generated from splitting single text sources anyway (as Figure~\ref{fig:example-iam-side-by-side} demonstrates), so many of our multi-page documents do, in fact, flow from page to page and/or share related text content. 

\paragraph{\mhills{}.}
This dataset is composed of 161 images taken on a smartphone. From these we construct 70 multi-page documents, 49 with two pages and 21 with three. 
The documents include meeting minutes, correspondence and legal documents from the public archives of a British charity, the majority being meeting minutes from between 1889 and 1938. They are written in English (with instances of Latin), in multiple hands and styles and often use archaic language and handwriting conventions. As a result, this task is more challenging than \iam{}. The documents are publicly accessible archival materials held by the Malvern Hills Trust, obtained with the explicit permission of the Trust. Ground truth transcriptions were produced by a single annotator, with \textsc{gpt-4o} used as a noisy first-pass tool to accelerate the process; every document was subsequently manually checked and corrected, and every document's transcription differs from \textsc{gpt-4o}'s initial output to some degree.

The dataset is included in the linked code repository; see Appendix~\ref{sec:app:mhills-metadata} for \mhills{} statistics, and Appendix~\ref{sec:app:mhills-ablation} for ablations on document properties.

\paragraph{\bentham{}.}
The Bentham-R0 dataset consists of 433 images of handwritten notes by the 18th-century philosopher Jeremy Bentham with crowd-sourced transcriptions \citep{causer2012building, causer2018making}. Each page is identified by source and page number, but there are many gaps and isolated pages, so we create a multi-page version by extracting all groups of consecutive pages (239 total). The resulting multi-page dataset, \bentham{}, is the most challenging of the three, and consists of 52 two-page documents, 21 three-page documents and 18 four-page documents.

\paragraph{\casia{}.}
We derive \casia{} from (the test split of) CASIA-HWDB2.1 \citep{liu2011casia}, a Chinese handwritten dataset of 300 page images. As with \iam{}, we group pages by author ID to produce 30 \textit{five-page} documents with consistent handwriting. We use Google Cloud Vision as the OCR engine for this dataset, as it supports Chinese. Full results are in Appendix~\ref{sec:app:casia}.

\section{Experiments}
\label{sec:experiments}
\begin{figure}[h]
    \centering
    \includegraphics[width=1.0\linewidth]{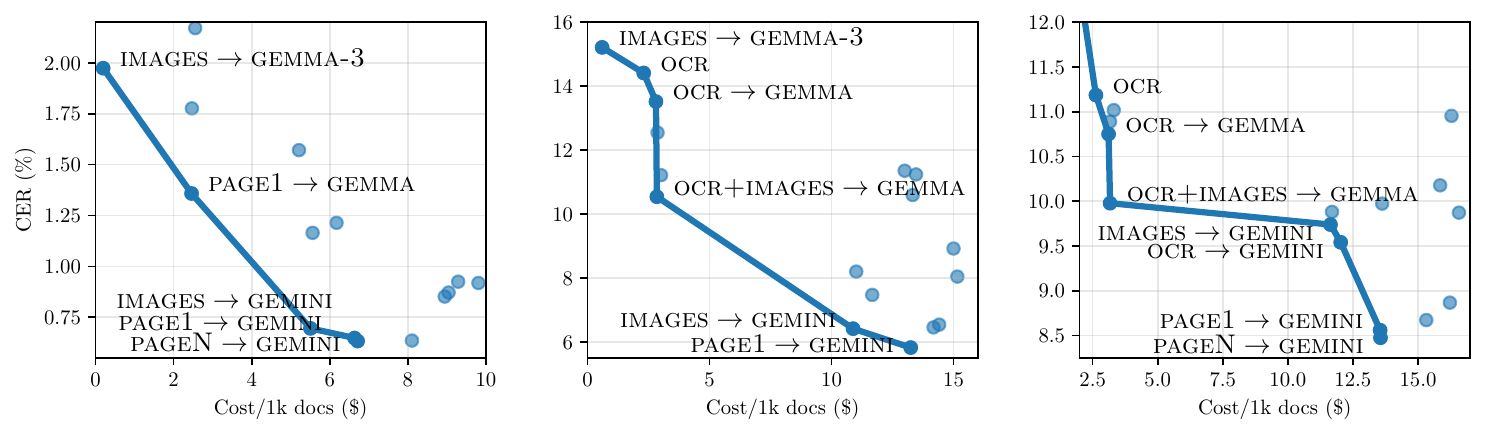}
    \caption{\small{Performance against cost for all methods on \iam{} (left), \mhills{} (center) and \bentham{} (right). Some outlier points are cut off for clarity. See Tables~\ref{tab:iam} and \ref{tab:mhills+bentham} for exact values. All Pareto frontier methods are labeled.
    We see that \firstpage{} and \chosenpage{} are Pareto frontier methods for all three tasks, and the best-performing overall.}}
    \label{fig:pareto_triptych}
\end{figure}

Full results are detailed in Tables~\ref{tab:iam} and \ref{tab:mhills+bentham}, with the best models trading off cost against transcription accuracy highlighted by Pareto frontier plots in Figure~\ref{fig:pareto_triptych}.

\begin{table}[ht]
\caption{\small{Transcription methods on the \iam{} dataset with Azure OCR. `Minor/Major\#Err.' are the average number of trivial/non-trivial errors per document for the given method, as assessed by an LLM; see Section~\ref{sec:llm-error-eval} for details. Our (\pagex{}) methods are bold. Score emphasis: \textbf{1st}, \textit{2nd}, \underline{3rd}.}}
\centering
\small
\label{tab:iam}
\begin{tabular}{@{}rlcccc@{}}
\toprule
\textbf{}                            & \textbf{Input} & \textbf{CER $\downarrow$ (\%)} & \textbf{Cost (\$) /1k docs} & \textbf{Minor\#Err.} & \textbf{Major\#Err.} \\ \midrule
\textsc{---} & \pylaia{}\tablefootnote{\cite{tarride2024improving}; an open-source non-MLLM baseline trained on \iam{} (see Appendix~\ref{sec:app:non-commercial-baselines}).}                    & 6.51          & 0.00  & ---  & ---           \\
\textsc{---} & \textsc{ocr}                 & 3.81          & 2.26  & 3.61 & 5.11          \\ \midrule
\multirow{6}{*}{\raisebox{-0.8em}[0pt][0pt]{\rotatebox[origin=c]{90}{\shortstack[c]{\textsc{gemma}\\\textsc{-3-27b}}}}}
    & \textsc{ocr}   & 2.93                           & 2.42      &          3.97        & 3.17                \\
             & \textsc{images}              & 2.31          & 0.19  & 2.65 & 4.32          \\
             & \textsc{images-all-at-once} & 1.97          & 0.19  & 3.45 & 4.05          \\
             & \textsc{ocr+images}          & 1.78          & 2.47  & 2.78 & 2.66          \\
             & \textsc{\textbf{ocr+pageN}}           & 2.17          & 2.55  & 3.44 & 2.22          \\
             & \textsc{\textbf{ocr+page1}}           & 1.36          & 2.46  & 2.41 & 2.03          \\ \midrule
\multirow{6}{*}[-0.7ex]{\rotatebox[origin=c]{90}{\textsc{gpt-4o}}}
         & \textsc{ocr}   & 1.21                           & 6.17                        & 2.81 & 1.86                \\
             & \textsc{images}              & 0.92          & 9.81  & 2.06 & 1.79          \\
             & \textsc{images-all-at-once} & 0.92          & 9.29  & 1.82 & 1.59          \\
             & \textsc{ocr+images}          & {0.72}    & 12.69 & 1.86 & 1.21          \\
             & \textsc{\textbf{ocr+pageN}}           & 0.87          & 9.04  & 1.95 & 1.29          \\
             & \textsc{\textbf{ocr+page1}}           & 0.85          & 8.94  & 1.66 & \underline{1.20}          \\ \midrule
\multirow{6}{*}{\raisebox{-0.8em}[0pt][0pt]{\rotatebox[origin=c]{90}{\shortstack[c]{\textsc{gemini}\\\textsc{-2.5-pro}}}}}

 & \textsc{ocr}   & 1.57                           & 5.21                        & 1.64 & 1.22                \\
             & \textsc{images}              & 1.16          & 5.56  & 1.66 & 1.47          \\
             & \textsc{images-all-at-once} & 0.70          & 5.50  & 1.77 & 1.37          \\
             & \textsc{ocr+images}          & \textit{0.64} & 8.10  & 1.63 & 1.32          \\
             & \textsc{\textbf{ocr+pageN}}           & \textbf{0.63} & 6.71  & 1.80 & \textit{1.04} \\
             & \textsc{\textbf{ocr+page1}}           & \underline{0.65} & 6.63  & 2.51 & \textbf{1.00} \\ \bottomrule
\end{tabular}
\end{table}

\subsection{Experimental details}
\label{sec:exp-details}
The task for each dataset is to produce transcriptions for each page in the given document, from page images, OCR output derived from those images, or some combination of the two. The methods are described in Section~\ref{sec:comprehensive-suite}. All MLLM prompts are included in the linked code repository. Prompts for each method were developed over four rounds of experimentation on a validation split of 100 multi-page documents from IAM Database images separate from the 242 images used to generate \iam{}.

\paragraph{OCR and MLLMs.} We use the Azure Vision OCR engine, as early experimentation comparing Azure Vision, Amazon Textract, Google Cloud Vision and Tesseract revealed Azure to be generally the best-performing OCR engine, as well as the cheapest (see Appendix~\ref{sec:app-additional-results}).

For MLLMs we use two leading commercial models at time of writing, OpenAI's \textsc{gpt-4o} and Google's  \textsc{gemini-2.5-pro}\footnote{We evaluate \textsc{gpt-4o} and \textsc{gemini-2.5-pro} as the respective flagship models of OpenAI and Google at the time our experiments were finalized (mid-2025); our primary goal in evaluating multiple MLLMs is to demonstrate that our methods generalize across model families, rather than to compare model quality directly. Extending evaluation to more recent models (e.g.\ GPT-5) is a natural direction for future work.}, as well as an open-source\footnote{Under Google's `open' Gemma license.} MLLM, \textsc{gemma-3-27b}. For each we use temperature of 0, minimal reasoning (for Gemini) and default parameters otherwise. Details of cost estimates for OCR and MLLMs are included in Appendix~\ref{sec:app-costs}, and per-stage token and cost breakdowns are in Appendices~\ref{sec:app:extended-tables}, \ref{sec:app:extended-experiments}~\&~\ref{sec:app:casia}.

We use a single evaluation run per method.\footnote{A single run keeps costs reasonable, and temperature-zero output is largely, though not perfectly, deterministic. We discuss this, and test it for a subset of methods, in Appendix \ref{sec:app:variance}.}
Evaluation is at the page level using Character Error Rate (CER; \citealt{morris2004and}), the most widely-used metric for OCR transcription.\footnote{We also experimented with ANLS \citep{peer2024anls} and WER; all three metrics produce very similar relative results.}

\subsection{Semantic evaluation with an LLM} 
\label{sec:llm-error-eval} 
CER measures character-level edit distance and therefore misses semantic quality: a formatting difference counts the same as a corrupted word, while an incorrect initial renders an entire name wrong despite costing only one character. We therefore supplement CER with an LLM-based semantic evaluation on \iam{}, classifying errors as \textit{minor} (formatting or phrasing differences that do not affect meaning) or \textit{major} (missing content, hallucinations, proper noun errors, etc.). Full methodology and error taxonomy are in Appendix~\ref{sec:app:semantic-error-eval}.

\begin{table}
\centering
\small
\caption{\small{Transcription methods on the \mhills{} and \bentham{} datasets. OCR engine is Azure Vision. Our methods are bold. Score emphasis: \textbf{1st}, \textit{2nd}, \underline{3rd}.}}
\label{tab:mhills+bentham}
\makebox[\textwidth][c]{%
  \begin{subtable}[t]{0.49\textwidth}
    \centering
    \caption{\mhills{} dataset}
    \label{tab:mhills}
    \begin{tabular}{@{}rlcc@{}}
    \toprule
    \textbf{} &
      \textbf{Input} &
      \textbf{\begin{tabular}[c]{@{}c@{}}CER $\downarrow$\\ (\%)\end{tabular}} &
      \textbf{\begin{tabular}[c]{@{}c@{}}Cost (\$)\\ /1k docs\end{tabular}} \\ \midrule
    \textsc{---} & \textsc{ocr}                & 14.41         & 2.30  \\ \midrule
    \multirow{6}{*}[-1.0ex]{\rotatebox[origin=c]{90}{\textsc{gemma-3-27b}}} &
      \textsc{ocr}                & 13.52         & 2.80  \\
     &
      \textsc{images} &
      27.19 &
      0.66 \\
     &
      \begin{tabular}[c]{@{}l@{}}\textsc{images}\\ \textsc{-all-at-once}\end{tabular} &
      15.21 &
      0.60 \\
                 & \textsc{ocr+images}         & 10.54         & 2.84  \\
                 & \textbf{\textsc{ocr+pageN}} & 11.22         & 3.01  \\
                 & \textbf{\textsc{ocr+page1}} & 12.55         & 2.87  \\ \midrule
    \multirow{6}{*}{\rotatebox[origin=c]{90}{\textsc{gpt-4o}}} &
      \textsc{ocr}                & 10.60         & 13.32 \\
                 & \textsc{images}             & 11.24         & 13.46 \\
                 & \begin{tabular}[c]{@{}l@{}}\textsc{images}\\ \textsc{-all-at-once}\end{tabular} &
      11.35 &
      13.00 \\
                 & \textsc{ocr+images}         & 7.11          & 17.54 \\
                 & \textbf{\textsc{ocr+pageN}} & 8.05          & 15.15 \\
                 & \textbf{\textsc{ocr+page1}} & 8.92          & 15.00 \\ \midrule
    \multirow{6}{*}[-0.6ex]{\rotatebox[origin=c]{90}{\textsc{gemini-2.5-pro}}} &
      \textsc{ocr}                & 7.47          & 11.67 \\
                 & \textsc{images}             & \textit{6.42} & 10.88 \\
                 & \begin{tabular}[c]{@{}l@{}}\textsc{images}\\ \textsc{-all-at-once}\end{tabular} &
      8.20 &
      11.01 \\
                 & \textsc{ocr+images}         & \underline{6.46} & 14.18 \\
                 & \textbf{\textsc{ocr+pageN}} & 6.54          & 14.41 \\
                 & \textbf{\textsc{ocr+page1}} & \textbf{5.83} & 13.24 \\ \bottomrule
    \end{tabular}
  \end{subtable}%
  \hspace{0.02\textwidth}
  \begin{subtable}[t]{0.49\textwidth}
    \centering
    \caption{\bentham{} dataset}
    \label{tab:bentham}
    \begin{tabular}{@{}rlcc@{}}
    \toprule
    \textbf{} &
      \textbf{Input} &
      \textbf{\begin{tabular}[c]{@{}c@{}}CER $\downarrow$\\ (\%)\end{tabular}} &
      \textbf{\begin{tabular}[c]{@{}c@{}}Cost (\$)\\ /1k docs\end{tabular}} \\ \midrule
    \textsc{---} & \textsc{ocr}        & 11.18            & 2.63  \\ \midrule
    \multirow{6}{*}[-1.0ex]{\rotatebox[origin=c]{90}{\textsc{gemma-3-27b}}} &
      \textsc{ocr}        & 10.75            & 3.11  \\
                 & \textsc{images}     & 15.12            & 0.59  \\
                 & \begin{tabular}[c]{@{}l@{}}\textsc{images}\\ \textsc{-all-at-once}\end{tabular} &
      27.88 &
      0.50 \\
                 & \textsc{ocr+images} & 9.98             & 3.17  \\
                 & \textsc{\textbf{ocr+pageN}}  & 11.02            & 3.31  \\
                 & \textsc{\textbf{ocr+page1}}  & 10.89            & 3.17  \\ \midrule
    \multirow{6}{*}{\rotatebox[origin=c]{90}{\textsc{gpt-4o}}} &
      \textsc{ocr}        & 9.97             & 13.63 \\
     &
      \textsc{images}     & 9.87             & 16.58 \\
                 & \begin{tabular}[c]{@{}l@{}}\textsc{images}\\ \textsc{-all-at-once}\end{tabular} &
      10.18 &
      15.85 \\
                 & \textsc{ocr+images} & 9.35             & 20.93 \\
                 & \textsc{\textbf{ocr+pageN}}  & {8.87} & 16.23 \\
                 & \textsc{ocr+page1} &
      10.95 &
      16.29 \\ \midrule
    \multirow{6}{*}[-0.6ex]{\rotatebox[origin=c]{90}{\textsc{gemini-2.5-pro}}} &
      \textsc{ocr}        & 9.54             & 12.03 \\
                 & \textsc{images}     & 9.74             & 11.64 \\
                 & \begin{tabular}[c]{@{}l@{}}\textsc{images}\\ \textsc{-all-at-once}\end{tabular} &
      9.88 &
      11.70 \\
                 & \textsc{ocr+images} & \underline{8.67}    & 15.32 \\
                 & \textsc{\textbf{ocr+pageN}}  & \textbf{8.48}    & 13.56 \\
                 & \textsc{\textbf{ocr+page1}}  & \textit{8.56}    & 13.55 \\ \bottomrule
    \end{tabular}
  \end{subtable}%
}

\end{table}

\subsection{Discussion}

Surprisingly, \textit{for all three datasets} we find that the best-performing method overall is either \firstpage{} or \chosenpage{}. 
Both of these methods have access to \textit{less than half}, and sometimes as little as a quarter, of the original source data, the document images. Yet they outperform methods that have access to all of the images, and those with access to both the OCR text \textit{and} all of the images, despite the fact that \textit{the performance of OCR \textbf{alone} is poor for all three tasks} --- for \iam{} it is the worst method outright, and for \mhills{} and \bentham{} it is outperformed by all \textsc{gemini}- and \textsc{gpt-4o}-based models.

The performance gap in each case is not large, but our intention is not to propose a SOTA method. It is to demonstrate that MLLMs can leverage common context from limited, expensive image input to improve correction of cheap text input. Put simply, a single image is often as good as the full document. Existing simplistic approaches to transcription based solely on OCR (with traditional OCR engines or MLLMs end-to-end), miss useful context, and methods that use both OCR and all images can overwhelm a model with redundant repeated context that can be found in a single image. 

\paragraph{Semantic accuracy.} Table~\ref{tab:iam} includes additional columns with information about document error types; this is described in Section~\ref{sec:llm-error-eval}. Counting MLLM errors semantically, rather than with strict CER, we see that the performance gap between \firstpage{} and \chosenpage{} and the next best method, \textsc{ocr+images$\rightarrow$gemini-2.5-pro}, is even larger. While all three methods have a similar average number of minor errors per document (semantic and formatting errors that do not affect meaning), our methods have over 20\% fewer major errors (genuine mistakes or hallucinations) than the next best method.

\paragraph{Scaling to longer documents.} We evaluate the same methods on longer-document variants of our benchmarks: \iamp{} (5 pages, 20 docs), \mhillsp{} (avg.\ 5.8 pages, 24 docs) and \mhillspp{} (avg.\ 11.5 pages, 10 docs). \firstpage{} remains the top or joint-top method across all three, despite having access to $<$20\% (or even $<$10\%) of document images, and are lower cost than next-best \textsc{ocr+images} methods (see Appendix~\ref{sec:app:extended-experiments}). Even when document pages have both different handwriting and unrelated context (\iamr{}; see Appendix~\ref{sec:app:iam-random}), \pagex{} methods remain competitive. Furthermore, ablations on document properties suggest that \pagex{} methods are most beneficial on \textit{more challenging documents}, e.g.\ those with archaic language (Appendix~\ref{sec:app:mhills-ablation}).

\paragraph{Generalization beyond English-language documents.} On \casia{}, a Chinese handwritten benchmark of 5-page documents (Table~\ref{tab:casia}; full results in Appendix~\ref{sec:app:casia}), our central finding generalizes to non-Latin script: with \textsc{gemini-2.5-pro}, \pagex{} achieves 8.94\% CER, nearly matching the more expensive \ocr{}+\textsc{images} (8.39\%) at lower token cost, and significantly improving on OCR alone (12.84\%). The result reinforces that a single well-chosen page captures most of the benefit of the full document image set.

\begin{table}[h]
\centering
\small
\caption{Transcription methods on \casia{}, a Chinese handwriting benchmark. OCR engine is Google Cloud Vision. Our (\pagex{}) methods are bold. Score emphasis: \textbf{1st}, \textit{2nd}, \underline{3rd}. Full results are in Appendix Table~\ref{tab:casia-rebuttal}.}
\label{tab:casia}

\begin{tabular}{@{}rlcc@{}}
\toprule
\textbf{} & \textbf{Input} & \textbf{CER $\downarrow$ (\%)} & \textbf{Cost (\$) /1k docs} \\ \midrule
\textsc{---} & \textsc{ocr} & 12.84 & 7.50 \\ \midrule
\multirow{6}{*}{\rotatebox[origin=c]{90}{\textsc{gpt-4o}}}
    & \textsc{ocr} & 18.32 & 26.75 \\
    & \textsc{images} & 42.40 & 30.81 \\
    & \textsc{images-all-at-once} & 43.64 & 29.30 \\
    & \textsc{ocr+images} & 14.16 & 41.58 \\
    & \textbf{\textsc{ocr+pageN}} & 12.87 & 28.68 \\
    & \textbf{\textsc{ocr+page1}} & 14.48 & 28.31 \\ \midrule
\multirow{6}{*}{\rotatebox[origin=c]{90}{\textsc{gemini}}}
    & \textsc{ocr} & 33.17 & 24.87 \\
    & \textsc{images} & 13.55 & 19.08 \\
    & \textsc{images-all-at-once} & 23.07 & 18.09 \\
    & \textsc{ocr+images} & \textbf{8.39} & 27.80 \\
    & \textbf{\textsc{ocr+pageN}} & \textit{8.94} & 23.42 \\
    & \textbf{\textsc{ocr+page1}} & \underline{12.61} & 23.25 \\ \bottomrule
\end{tabular}

\end{table}

\paragraph{Open-model baselines.} We also evaluate \pylaia{} \citep{tarride2024improving} on \iam{} (Table~\ref{tab:iam}), and \trocr{} \citep{li2023trocr} and \docowl{} \citep{hu2025mplug} on \mhillsp{} (Appendix~\ref{sec:app:extended-experiments}). All perform poorly in our zero-shot, document-level setting: \pylaia{} achieves 6.51\% CER on \iam{} (worse than OCR alone), \trocr{} achieves 31.4\% on \mhillsp{}, and \docowl{} is effectively unusable at 92\% CER.

\section{Limitations and future work}
\label{sec:limitations}
We stated that we would expect \chosenpage{} to be at least as good as \firstpage{}, yet in practice we find that neither method is consistently better than the other across datasets or MLLMs. We hypothesize that this relates to MLLMs' sensitivity to prompt ordering: \firstpage{}'s prompt keeps the selected page image adjacent to its corresponding OCR text, whereas \chosenpage{} may select a page far from the start of the prompt, and MLLMs are known to be better at local than long-range context reasoning. We find this effect to be more model- than document-length-dependent: on \mhills{}, \chosenpage{} consistently outperforms \firstpage{} for \textsc{gpt-4o}, and \firstpage{} consistently outperforms \chosenpage{} for \textsc{gemini-2.5-pro}, holding across all three document-length variants we test (2--3, 5+, and 10+ pages) --- suggesting this is a stable property of each model's handling of prompt order, rather than an artifact of any one dataset or document length. We discuss this further, with full results, in Appendix~\ref{sec:app:pageN-vs-page1}.

\paragraph{Cost and scaling.}
Although our methods use only a single page, they are sometimes more expensive than methods which use the full set of images. This is due to a combination of high text-density OCR transcriptions, as well as the added cost of OCR transcription itself. One way to address this could be the use of cheaper OCR; we note that some MLLMs, such as \textsc{gemma-3-27b}, are significantly cheaper than our chosen OCR engine, and can achieve improved or comparable results over OCR depending on the task. Further experimentation using cheap MLLMs as alternative OCR engines and more powerful MLLMs as post-processors is an area we leave for future work. 

Prompt caching \citep{shi2024keep} could further improve cost scaling for \pagex{} methods on long documents, as the shared page image tokens would be cached across per-page chunks; we leave this for future work.

\paragraph{Conclusion.} In this work we investigated the transcription of multi-page handwritten documents using various configurations of commercial OCR engines and MLLMs. We provide a set of multi-page transcription benchmarks, including a brand new dataset, \mhills{}, which we hope will serve as a useful evaluation tool for the community.
We also provide the first known evaluation of the effectiveness of different prompting strategies for the task of zero-shot, multi-page handwriting transcription, on our three benchmark tasks. We propose the \firstpage{} and \chosenpage{} methods, and empirically demonstrate that they improve transcription accuracy while balancing cost and performance. Notably, they are equally or more effective than end-to-end processing with leading MLLMs, despite not having access to all page images.

\section*{Acknowledgments}
BG and MJ are supported by the EPSRC Centre for Doctoral Training in Autonomous Intelligent Machines and Systems (EP/S024050/1). XD is supported by the Oxford-Man Institute of Quantitative Finance and EPSRC (EP/T023333/1). BG carried out part of this work during an internship at QuantCo.

\newpage

\bibliography{my}

@article{vaswani2017attention,
  title={Attention is all you need},
  author={Vaswani, Ashish and Shazeer, Noam and Parmar, Niki and Uszkoreit, Jakob and Jones, Llion and Gomez, Aidan N and Kaiser, {\L}ukasz and Polosukhin, Illia},
  journal={Advances in neural information processing systems},
  volume={30},
  year={2017}
}

@inproceedings{liu2021swin,
  title={Swin transformer: Hierarchical vision transformer using shifted windows},
  author={Liu, Ze and Lin, Yutong and Cao, Yue and Hu, Han and Wei, Yixuan and Zhang, Zheng and Lin, Stephen and Guo, Baining},
  booktitle={Proceedings of the IEEE/CVF international conference on computer vision},
  pages={10012--10022},
  year={2021}
}

@inproceedings{karpinska2024one,
  title={One thousand and one pairs: A “novel” challenge for long-context language models},
  author={Karpinska, Marzena and Thai, Katherine and Lo, Kyle and Goyal, Tanya and Iyyer, Mohit},
  booktitle={Proceedings of the 2024 Conference on Empirical Methods in Natural Language Processing},
  pages={17048--17085},
  year={2024}
}

@inproceedings{li2023trocr,
  title={Trocr: Transformer-based optical character recognition with pre-trained models},
  author={Li, Minghao and Lv, Tengchao and Chen, Jingye and Cui, Lei and Lu, Yijuan and Florencio, Dinei and Zhang, Cha and Li, Zhoujun and Wei, Furu},
  booktitle={Proceedings of the AAAI conference on artificial intelligence},
  volume={37},
  number={11},
  pages={13094--13102},
  year={2023}
}

@inproceedings{morris2004and,
  title={From WER and RIL to MER and WIL: improved evaluation measures for connected speech recognition.},
  author={Morris, Andrew Cameron and Maier, Viktoria and Green, Phil D},
  booktitle={Interspeech},
  number={4-8},
  pages={2004},
  year={2004}
}

@inproceedings{dong2024survey,
  title={A survey on in-context learning},
  author={Dong, Qingxiu and Li, Lei and Dai, Damai and Zheng, Ce and Ma, Jingyuan and Li, Rui and Xia, Heming and Xu, Jingjing and Wu, Zhiyong and Chang, Baobao and others},
  booktitle={Proceedings of the 2024 conference on empirical methods in natural language processing},
  pages={1107--1128},
  year={2024}
}

@article{liu2023hidden,
  title={On the hidden mystery of ocr in large multimodal models},
  author={Liu, Yuliang and Li, Zhang and Li, Hongliang and Yu, Wenwen and Huang, Mingxin and Peng, Dezhi and Liu, Mingyu and Chen, Mingrui and Li, Chunyuan and Jin, Lianwen and others},
  journal={arXiv preprint arXiv:2305.07895},
  volume={2},
  number={5},
  pages={6},
  year={2023}
}

@article{marti2002iam,
  title={The IAM-database: an English sentence database for offline handwriting recognition},
  author={Marti, U-V and Bunke, Horst},
  journal={International journal on document analysis and recognition},
  volume={5},
  number={1},
  pages={39--46},
  year={2002},
  publisher={Springer-Verlag Berlin Heidelberg}
}

@article{kim2024fables,
  title={Fables: Evaluating faithfulness and content selection in book-length summarization},
  author={Kim, Yekyung and Chang, Yapei and Karpinska, Marzena and Garimella, Aparna and Manjunatha, Varun and Lo, Kyle and Goyal, Tanya and Iyyer, Mohit},
  journal={arXiv preprint arXiv:2404.01261},
  year={2024}
}

@inproceedings{lund2011progressive,
  title={Progressive alignment and discriminative error correction for multiple OCR engines},
  author={Lund, William B and Walker, Daniel D and Ringger, Eric K},
  booktitle={2011 International Conference on Document Analysis and Recognition},
  pages={764--768},
  year={2011},
  organization={IEEE}
}

@inproceedings{schaefer2020two,
  title={A two-step approach for automatic OCR post-correction},
  author={Schaefer, Robin and Neudecker, Clemens},
  booktitle={Proceedings of the 4th Joint SIGHUM Workshop on Computational Linguistics for Cultural Heritage, Social Sciences, Humanities and Literature},
  pages={52--57},
  year={2020}
}

@inproceedings{rigaud2019icdar,
  title={ICDAR 2019 competition on post-OCR text correction},
  author={Rigaud, Christophe and Doucet, Antoine and Coustaty, Micka{\"e}l and Moreux, Jean-Philippe},
  booktitle={2019 international conference on document analysis and recognition (ICDAR)},
  pages={1588--1593},
  year={2019},
  organization={IEEE}
}

@inproceedings{fujitake2024dtrocr,
  title={Dtrocr: Decoder-only transformer for optical character recognition},
  author={Fujitake, Masato},
  booktitle={Proceedings of the IEEE/CVF winter conference on applications of computer vision},
  pages={8025--8035},
  year={2024}
}

@article{causer2018making,
  title={‘Making such bargain’: Transcribe Bentham and the quality and cost-effectiveness of crowdsourced transcription},
  author={Causer, Tim and Grint, Kris and Sichani, Anna-Maria and Terras, Melissa},
  journal={Digital Scholarship in the Humanities},
  volume={33},
  number={3},
  pages={467--487},
  year={2018},
  publisher={Oxford University Press}
}

@article{dosovitskiy2020image,
  title={An image is worth 16x16 words: Transformers for image recognition at scale},
  author={Dosovitskiy, Alexey and Beyer, Lucas and Kolesnikov, Alexander and Weissenborn, Dirk and Zhai, Xiaohua and Unterthiner, Thomas and Dehghani, Mostafa and Minderer, Matthias and Heigold, Georg and Gelly, Sylvain and others},
  journal={arXiv preprint arXiv:2010.11929},
  year={2020}
}

@article{liu2019roberta,
  title={Roberta: A robustly optimized bert pretraining approach},
  author={Liu, Yinhan and Ott, Myle and Goyal, Naman and Du, Jingfei and Joshi, Mandar and Chen, Danqi and Levy, Omer and Lewis, Mike and Zettlemoyer, Luke and Stoyanov, Veselin},
  journal={arXiv preprint arXiv:1907.11692},
  year={2019}
}

@article{devlin2018bert,
  title={Bert: Pre-training of deep bidirectional transformers for language understanding},
  author={Devlin, Jacob},
  journal={arXiv preprint arXiv:1810.04805},
  year={2018}
}

@article{achiam2023gpt,
  title={Gpt-4 technical report},
  author={Achiam, Josh and Adler, Steven and Agarwal, Sandhini and Ahmad, Lama and Akkaya, Ilge and Aleman, Florencia Leoni and Almeida, Diogo and Altenschmidt, Janko and Altman, Sam and Anadkat, Shyamal and others},
  journal={arXiv preprint arXiv:2303.08774},
  year={2023}
}

@article{floridi2020gpt,
  title={GPT-3: Its nature, scope, limits, and consequences},
  author={Floridi, Luciano and Chiriatti, Massimo},
  journal={Minds and machines},
  volume={30},
  number={4},
  pages={681--694},
  year={2020},
  publisher={Springer Netherlands Dordrecht}
}

@article{zhao2023survey,
  title={A survey of large language models},
  author={Zhao, Wayne Xin and Zhou, Kun and Li, Junyi and Tang, Tianyi and Wang, Xiaolei and Hou, Yupeng and Min, Yingqian and Zhang, Beichen and Zhang, Junjie and Dong, Zican and others},
  journal={arXiv preprint arXiv:2303.18223},
  volume={1},
  number={2},
  pages={1--124},
  year={2023}
}

@article{shi2024keep,
  title={Keep the cost down: A review on methods to optimize LLM's KV-cache consumption},
  author={Shi, Luohe and Zhang, Hongyi and Yao, Yao and Li, Zuchao and Zhao, Hai},
  journal={arXiv preprint arXiv:2407.18003},
  year={2024}
}

@article{veninga2024llms,
  title={LLMs for OCR post-correction},
  author={Veninga, Martijn},
  journal={URL http://essay. utwente. nl/102117/, master Thesis},
  year={2024}
}

@inproceedings{stanislawek2021kleister,
  title={Kleister: key information extraction datasets involving long documents with complex layouts},
  author={Stanis{\l}awek, Tomasz and Grali{\'n}ski, Filip and Wr{\'o}blewska, Anna and Lipi{\'n}ski, Dawid and Kaliska, Agnieszka and Rosalska, Paulina and Topolski, Bartosz and Biecek, Przemys{\l}aw},
  booktitle={International Conference on Document Analysis and Recognition},
  pages={564--579},
  year={2021},
  organization={Springer}
}

@article{gralinski2020kleister,
  title={Kleister: A novel task for information extraction involving long documents with complex layout},
  author={Grali{\'n}ski, Filip and Stanis{\l}awek, Tomasz and Wr{\'o}blewska, Anna and Lipi{\'n}ski, Dawid and Kaliska, Agnieszka and Rosalska, Paulina and Topolski, Bartosz and Biecek, Przemys{\l}aw},
  journal={arXiv preprint arXiv:2003.02356},
  year={2020}
}

@inproceedings{park2019cord,
  title={Cord: a consolidated receipt dataset for post-ocr parsing},
  author={Park, Seunghyun and Shin, Seung and Lee, Bado and Lee, Junyeop and Surh, Jaeheung and Seo, Minjoon and Lee, Hwalsuk},
  booktitle={Workshop on document intelligence at NeurIPS},
  volume={2019},
  number={1},
  pages={5},
  year={2019}
}

@inproceedings{huang2022layoutlmv3,
  title={Layoutlmv3: Pre-training for document ai with unified text and image masking},
  author={Huang, Yupan and Lv, Tengchao and Cui, Lei and Lu, Yutong and Wei, Furu},
  booktitle={Proceedings of the 30th ACM international conference on multimedia},
  pages={4083--4091},
  year={2022}
}

@inproceedings{xu2020layoutlm,
  title={Layoutlm: Pre-training of text and layout for document image understanding},
  author={Xu, Yiheng and Li, Minghao and Cui, Lei and Huang, Shaohan and Wei, Furu and Zhou, Ming},
  booktitle={Proceedings of the 26th ACM SIGKDD international conference on knowledge discovery \& data mining},
  pages={1192--1200},
  year={2020}
}

@article{kim2021donut,
  title={Donut: Document understanding transformer without ocr},
  author={Kim, Geewook and Hong, Teakgyu and Yim, Moonbin and Park, Jinyoung and Yim, Jinyeong and Hwang, Wonseok and Yun, Sangdoo and Han, Dongyoon and Park, Seunghyun},
  journal={arXiv preprint arXiv:2111.15664},
  volume={7},
  number={15},
  pages={2},
  year={2021}
}

@article{chen2023longlora,
  title={Longlora: Efficient fine-tuning of long-context large language models},
  author={Chen, Yukang and Qian, Shengju and Tang, Haotian and Lai, Xin and Liu, Zhijian and Han, Song and Jia, Jiaya},
  journal={arXiv preprint arXiv:2309.12307},
  year={2023}
}

@inproceedings{breuel2013high,
  title={High-performance OCR for printed English and Fraktur using LSTM networks},
  author={Breuel, Thomas M and Ul-Hasan, Adnan and Al-Azawi, Mayce Ali and Shafait, Faisal},
  booktitle={2013 12th international conference on document analysis and recognition},
  pages={683--687},
  year={2013},
  organization={IEEE}
}

@article{bora2020handwritten,
  title={Handwritten character recognition from images using CNN-ECOC},
  author={Bora, Mayur Bhargab and Daimary, Dinthisrang and Amitab, Khwairakpam and Kandar, Debdatta},
  journal={Procedia Computer Science},
  volume={167},
  pages={2403--2409},
  year={2020},
  publisher={Elsevier}
}

@article{sanchez2019set,
  title={A set of benchmarks for handwritten text recognition on historical documents},
  author={S{\'a}nchez, Joan Andreu and Romero, Ver{\'o}nica and Toselli, Alejandro H and Villegas, Mauricio and Vidal, Enrique},
  journal={Pattern Recognition},
  volume={94},
  pages={122--134},
  year={2019},
  publisher={Elsevier}
}

@inproceedings{dolfing2020scribblelens,
  title={The “ScribbleLens” Dutch historical handwriting corpus},
  author={Dolfing, Hans JGA and Bellegarda, Jerome and Chorowski, Jan and Marxer, Ricard and Laurent, Antoine},
  booktitle={2020 17th international conference on frontiers in handwriting recognition (ICFHR)},
  pages={67--72},
  year={2020},
  organization={IEEE}
}

@inproceedings{yuan2022syntax,
  title={Syntax-aware network for handwritten mathematical expression recognition},
  author={Yuan, Ye and Liu, Xiao and Dikubab, Wondimu and Liu, Hui and Ji, Zhilong and Wu, Zhongqin and Bai, Xiang},
  booktitle={Proceedings of the IEEE/CVF conference on computer vision and pattern recognition},
  pages={4553--4562},
  year={2022}
}

@inproceedings{diem2014icfhr,
  title={ICFHR 2014 competition on handwritten digit string recognition in challenging datasets (HDSRC 2014)},
  author={Diem, Markus and Fiel, Stefan and Kleber, Florian and Sablatnig, Robert and Saavedra, Jose M and Contreras, David and Barrios, Juan Manuel and Oliveira, Luiz S},
  booktitle={2014 14th International Conference on Frontiers in Handwriting Recognition},
  pages={779--784},
  year={2014},
  organization={IEEE}
}

@inproceedings{zhang2019icdar,
  title={Icdar 2019 robust reading challenge on reading chinese text on signboard},
  author={Zhang, Rui and Zhou, Yongsheng and Jiang, Qianyi and Song, Qi and Li, Nan and Zhou, Kai and Wang, Lei and Wang, Dong and Liao, Minghui and Yang, Mingkun and others},
  booktitle={2019 international conference on document analysis and recognition (ICDAR)},
  pages={1577--1581},
  year={2019},
  organization={IEEE}
}

@inproceedings{serrano2010rodrigo,
  title={The RODRIGO Database.},
  author={Serrano, Nicol{\'a}s and Castro, Francisco and Juan, Alfons},
  booktitle={LREC},
  pages={19--21},
  year={2010}
}

@inproceedings{wigington2018start,
  title={Start, follow, read: End-to-end full-page handwriting recognition},
  author={Wigington, Curtis and Tensmeyer, Chris and Davis, Brian and Barrett, William and Price, Brian and Cohen, Scott},
  booktitle={Proceedings of the European conference on computer vision (ECCV)},
  pages={367--383},
  year={2018}
}

@inproceedings{carbonell2019end,
  title={End-to-end handwritten text detection and transcription in full pages},
  author={Carbonell, Manuel and Mas, Joan and Villegas, Mauricio and Forn{\'e}s, Alicia and Llad{\'o}s, Josep},
  booktitle={2019 International conference on document analysis and recognition workshops (ICDARW)},
  volume={5},
  pages={29--34},
  year={2019},
  organization={IEEE}
}

@inproceedings{huang2019icdar2019,
  title={Icdar2019 competition on scanned receipt ocr and information extraction},
  author={Huang, Zheng and Chen, Kai and He, Jianhua and Bai, Xiang and Karatzas, Dimosthenis and Lu, Shijian and Jawahar, CV},
  booktitle={2019 International Conference on Document Analysis and Recognition (ICDAR)},
  pages={1516--1520},
  year={2019},
  organization={IEEE}
}

@article{yu2021benchmarking,
  title={Benchmarking chinese text recognition: Datasets, baselines, and an empirical study},
  author={Yu, Haiyang and Chen, Jingye and Li, Bin and Ma, Jianqi and Guan, Mengnan and Xu, Xixi and Wang, Xiaocong and Qu, Shaobo and Xue, Xiangyang},
  journal={arXiv preprint arXiv:2112.15093},
  year={2021}
}

@inproceedings{yang2019handwriting,
  title={Handwriting text recognition based on faster R-CNN},
  author={Yang, Junqing and Ren, Peng and Kong, Xiaoxiao},
  booktitle={2019 Chinese Automation Congress (CAC)},
  pages={2450--2454},
  year={2019},
  organization={IEEE}
}

@inproceedings{azawi2013normalizing,
  title={Normalizing historical orthography for OCR historical documents using LSTM},
  author={Azawi, Mayce Al and Afzal, Muhammad Zeshan and Breuel, Thomas M},
  booktitle={Proceedings of the 2nd International Workshop on Historical Document Imaging and Processing},
  pages={80--85},
  year={2013}
}

@article{peer2024anls,
  title={ANLS*--a universal document processing metric for generative large language models},
  author={Peer, David and Sch{\"o}pf, Philemon and Nebendahl, Volckmar and Rietzler, Alexander and Stabinger, Sebastian},
  journal={arXiv preprint arXiv:2402.03848},
  year={2024}
}

@article{liu2023lost,
  title={Lost in the middle: How language models use long contexts, 2023},
  author={Liu, Nelson F and Lin, Kevin and Hewitt, John and Paranjape, Ashwin and Bevilacqua, Michele and Petroni, Fabio and Liang, Percy},
  journal={URL https://arxiv. org/abs/2307.03172},
  volume={2},
  year={2023}
}

@inproceedings{wu2023multimodal,
  title={Multimodal large language models: A survey},
  author={Wu, Jiayang and Gan, Wensheng and Chen, Zefeng and Wan, Shicheng and Yu, Philip S},
  booktitle={2023 IEEE International Conference on Big Data (BigData)},
  pages={2247--2256},
  year={2023},
  organization={IEEE}
}

@article{nockels2022understanding,
  title={Understanding the application of handwritten text recognition technology in heritage contexts: a systematic review of Transkribus in published research},
  author={Nockels, Joe and Gooding, Paul and Ames, Sarah and Terras, Melissa},
  journal={Archival science},
  volume={22},
  number={3},
  pages={367--392},
  year={2022},
  publisher={Springer}
}

@inproceedings{kahle2017transkribus,
  title={Transkribus-a service platform for transcription, recognition and retrieval of historical documents},
  author={Kahle, Philip and Colutto, Sebastian and Hackl, G{\"u}nter and M{\"u}hlberger, G{\"u}nter},
  booktitle={2017 14th iapr international conference on document analysis and recognition (icdar)},
  volume={4},
  pages={19--24},
  year={2017},
  organization={IEEE}
}

@inproceedings{wang2025marten,
  title={Marten: Visual question answering with mask generation for multi-modal document understanding},
  author={Wang, Zining and Guan, Tongkun and Fu, Pei and Duan, Chen and Jiang, Qianyi and Guo, Zhentao and Guo, Shan and Luo, Junfeng and Shen, Wei and Yang, Xiaokang},
  booktitle={Proceedings of the Computer Vision and Pattern Recognition Conference},
  pages={14460--14471},
  year={2025}
}

@inproceedings{liao2025doclayllm,
  title={Doclayllm: An efficient multi-modal extension of large language models for text-rich document understanding},
  author={Liao, Wenhui and Wang, Jiapeng and Li, Hongliang and Wang, Chengyu and Huang, Jun and Jin, Lianwen},
  booktitle={Proceedings of the IEEE/CVF Conference on Computer Vision and Pattern Recognition},
  pages={4038--4049},
  year={2025}
}

@inproceedings{luo2024layoutllm,
  title={Layoutllm: Layout instruction tuning with large language models for document understanding},
  author={Luo, Chuwei and Shen, Yufan and Zhu, Zhaoqing and Zheng, Qi and Yu, Zhi and Yao, Cong},
  booktitle={Proceedings of the IEEE/CVF conference on computer vision and pattern recognition},
  pages={15630--15640},
  year={2024}
}

@inproceedings{wang2024docllm,
  title={Docllm: A layout-aware generative language model for multimodal document understanding},
  author={Wang, Dongsheng and Raman, Natraj and Sibue, Mathieu and Ma, Zhiqiang and Babkin, Petr and Kaur, Simerjot and Pei, Yulong and Nourbakhsh, Armineh and Liu, Xiaomo},
  booktitle={Proceedings of the 62nd Annual Meeting of the Association for Computational Linguistics (Volume 1: Long Papers)},
  pages={8529--8548},
  year={2024}
}

@article{causer2012building,
  title={Building a volunteer community: results and findings from Transcribe Bentham},
  author={Causer, Tim and Wallace, Valerie},
  journal={Digital Humanities Quarterly},
  volume={6},
  number={2},
  year={2012}
}

@article{guan2025order,
  title={The order effect: investigating prompt sensitivity to input order in LLMs},
  author={Guan, Bryan and Roosta, Tanya and Passban, Peyman and Rezagholizadeh, Mehdi},
  journal={arXiv preprint arXiv:2502.04134},
  year={2025}
}

@article{liu2025comprehensive,
  title={A comprehensive survey on long context language modeling},
  author={Liu, Jiaheng and Zhu, Dawei and Bai, Zhiqi and He, Yancheng and Liao, Huanxuan and Que, Haoran and Wang, Zekun and Zhang, Chenchen and Zhang, Ge and Zhang, Jiebin and others},
  journal={arXiv preprint arXiv:2503.17407},
  year={2025}
}

@book{kress2015reading,
  title={The reading teacher's book of lists},
  author={Kress, Jacqueline E and Fry, Edward B},
  year={2015},
  publisher={John Wiley \& Sons}
}

@inproceedings{tarride2024improving,
  title={Improving automatic text recognition with language models in the PyLaia open-source library},
  author={Tarride, Sol{\`e}ne and Schneider, Yoann and Generali-Lince, Marie and Boillet, M{\'e}lodie and Abadie, Bastien and Kermorvant, Christopher},
  booktitle={International Conference on Document Analysis and Recognition},
  pages={387--404},
  year={2024},
  organization={Springer Nature Switzerland Cham}
}

@inproceedings{liu2011casia,
  title={CASIA online and offline Chinese handwriting databases},
  author={Liu, Cheng-Lin and Yin, Fei and Wang, Da-Han and Wang, Qiu-Feng},
  booktitle={2011 international conference on document analysis and recognition},
  pages={37--41},
  year={2011},
  organization={IEEE}
}

@inproceedings{hu2025mplug,
  title={mplug-docowl2: High-resolution compressing for ocr-free multi-page document understanding},
  author={Hu, Anwen and Xu, Haiyang and Zhang, Liang and Ye, Jiabo and Yan, Ming and Zhang, Ji and Jin, Qin and Huang, Fei and Zhou, Jingren},
  booktitle={Proceedings of the 63rd Annual Meeting of the Association for Computational Linguistics (Volume 1: Long Papers)},
  pages={5817--5834},
  year={2025}
}
\bibliographystyle{colm2026_conference}

\appendix
\raggedbottom

\section{Example documents from datasets}
\label{sec:app:example-docs}
\begin{figure}[H]
  \centering
  \begin{subfigure}[b]{0.45\linewidth}
    \centering
    \includegraphics[width=\linewidth]{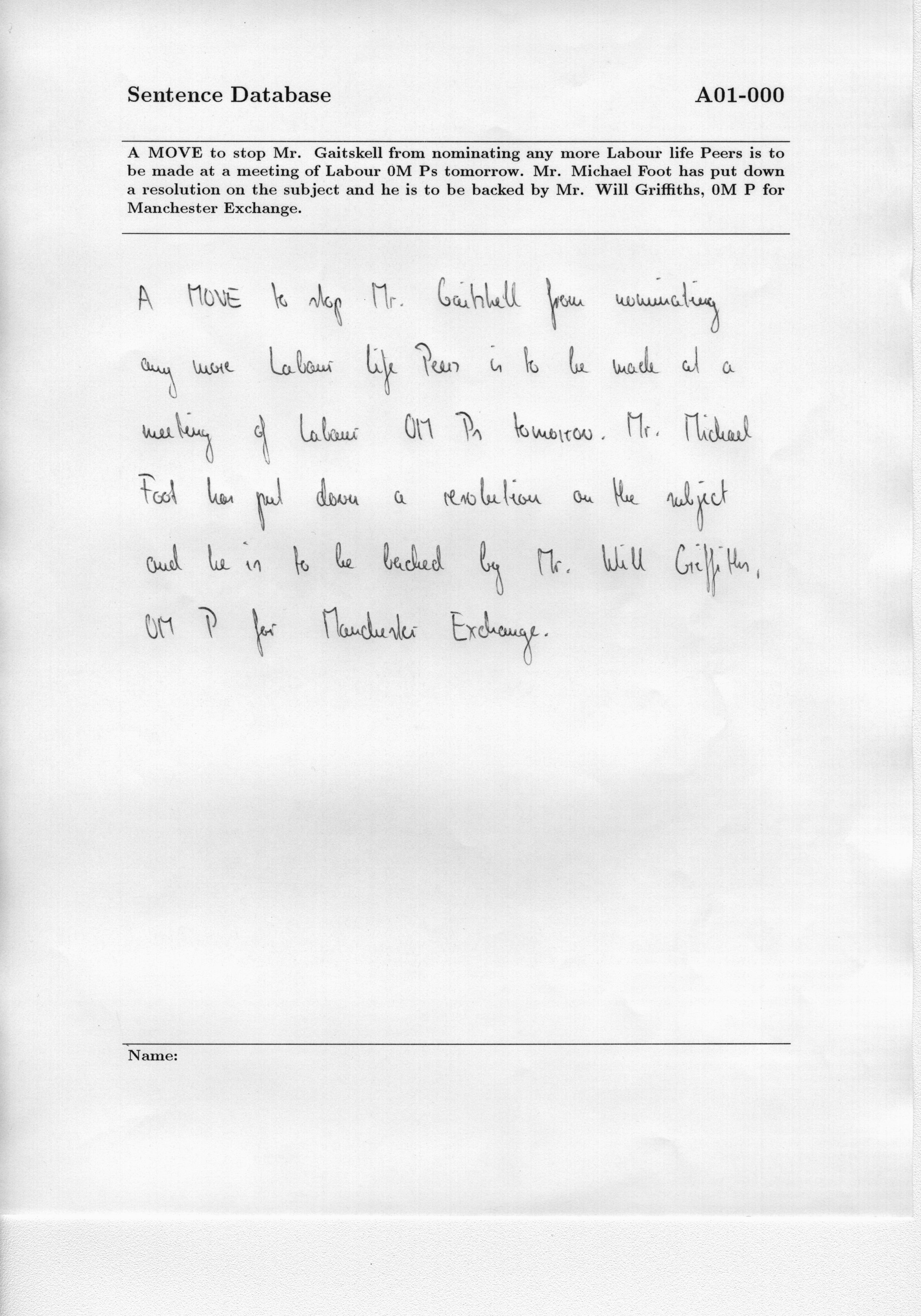}
    \caption{An example of a document from the IAM Handwriting Database.}
    \label{fig:example-iam-whole}
  \end{subfigure}\hfill
  \begin{subfigure}[b]{0.45\linewidth}
    \centering
    \includegraphics[width=\linewidth]{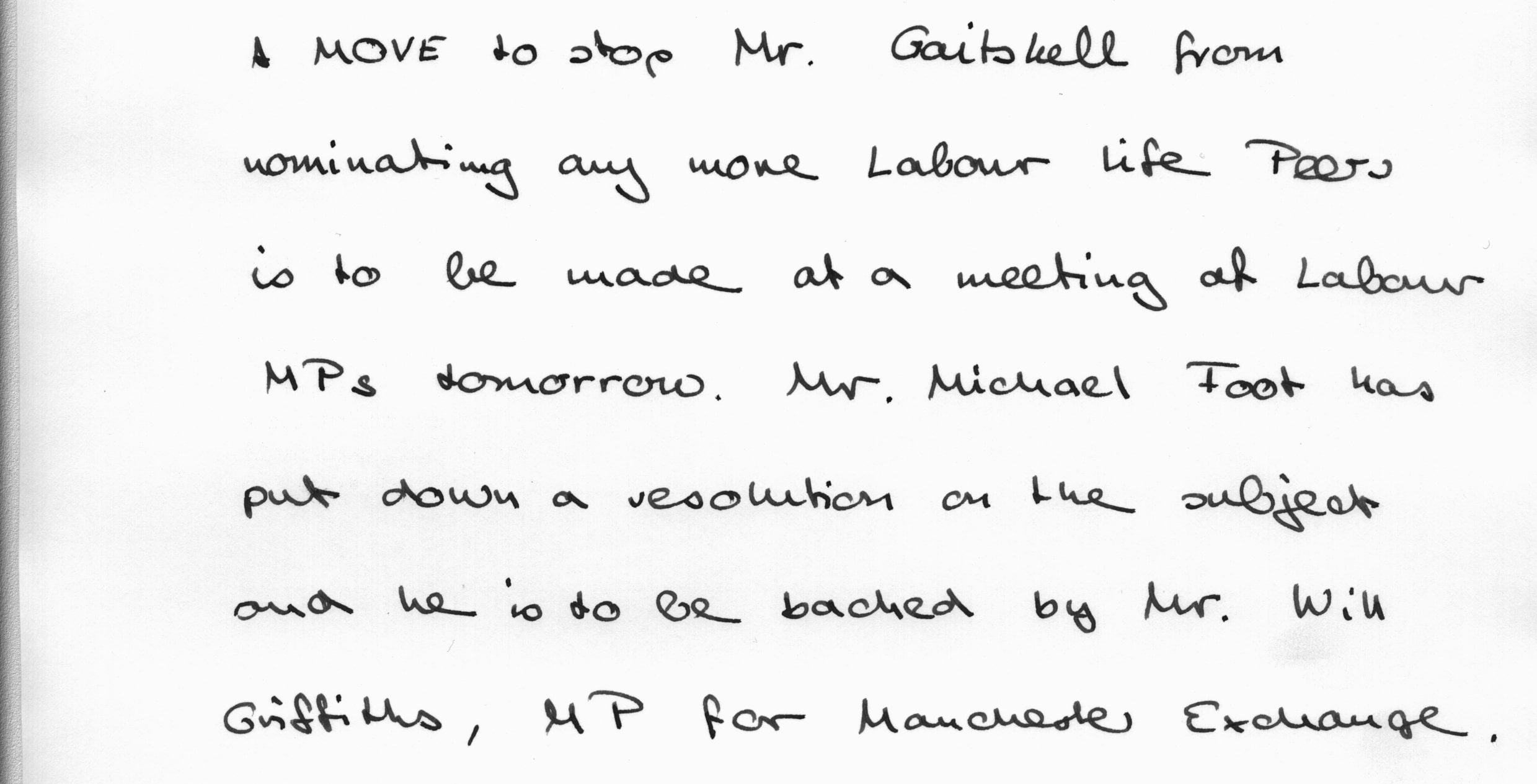}\\[0.5em]
    \includegraphics[width=\linewidth]{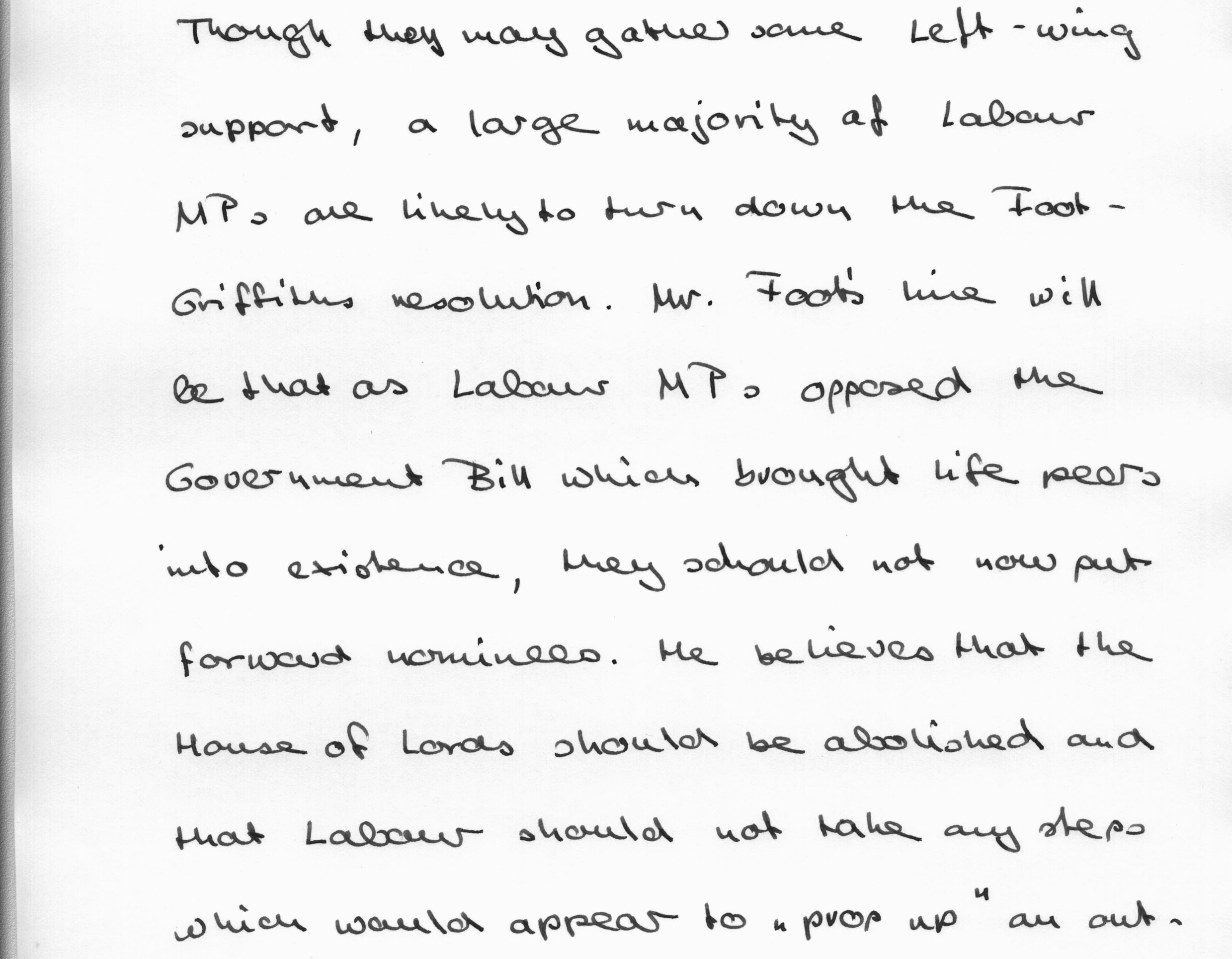}
    \caption{An example of a constructed multi-page \iam{} document, combining two pages from the same writer with the machine-printed text cropped out.}
    \label{fig:example-iam-multipage}
  \end{subfigure}
  \caption{Left: original IAM document. Right: constructed two-page \iam{} document.}
  \label{fig:example-iam-side-by-side}
\end{figure}

\begin{figure}[H]
  \centering
  \begin{subfigure}[t]{0.45\linewidth}
    \centering
    \includegraphics[width=\linewidth]{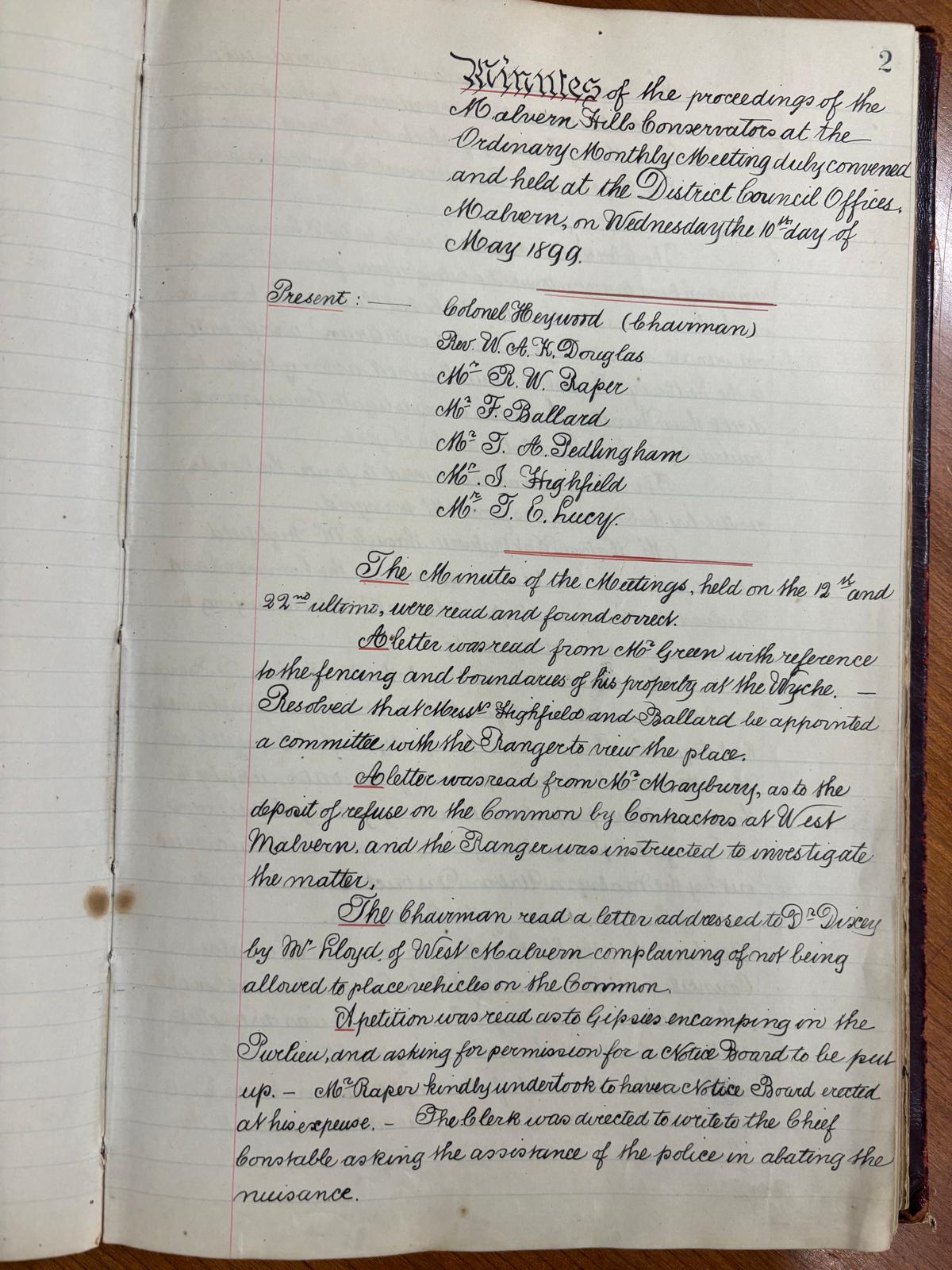} \\[0.5em]
    \caption{}
    \label{fig:example-mhills-1}  
  \end{subfigure}
  \begin{subfigure}[t]{0.4\linewidth}
    \centering
    \includegraphics[width=\linewidth]{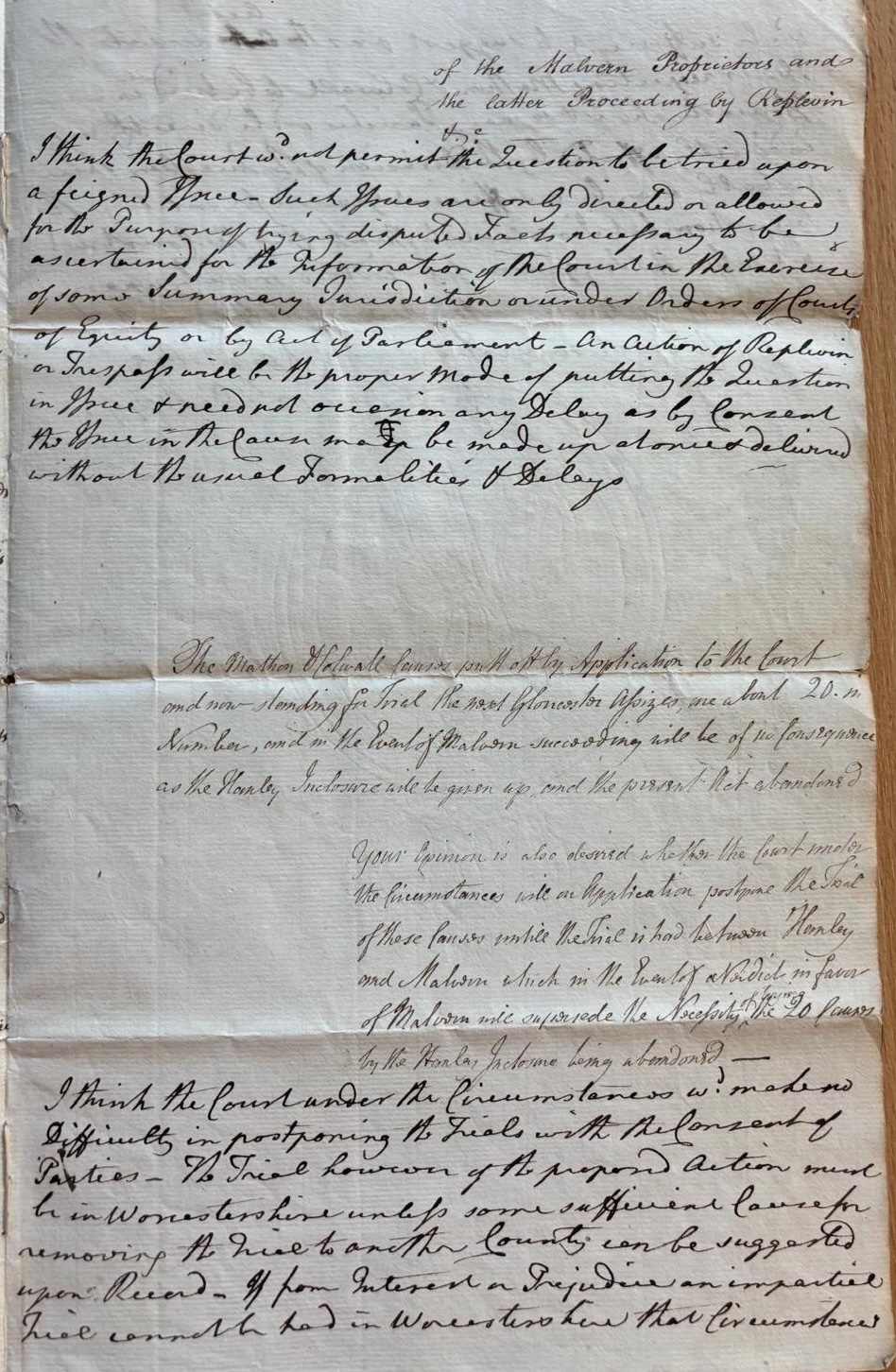} \\[0.5em]
    \caption{}
    \label{fig:example-mhills-2}
  \end{subfigure}
  \caption{Example pages from the \mhills{} dataset.}
  \label{fig:mhills-example-pages}
\end{figure}

\begin{figure}[H]
  \centering
  \begin{subfigure}[t]{0.45\linewidth}
    \centering
    \includegraphics[width=\linewidth]{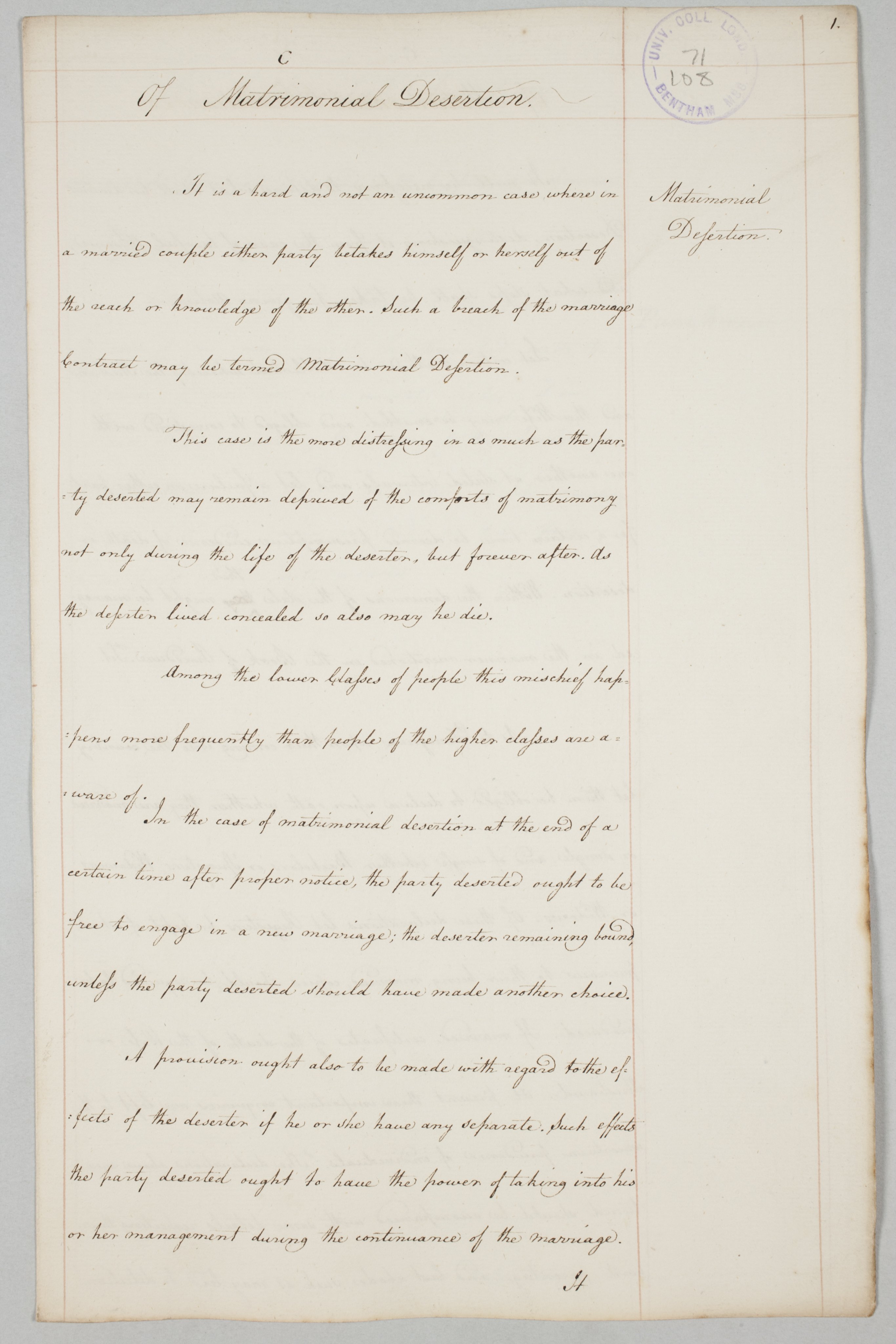} \\[0.5em]
    \caption{}
    \label{fig:example-bentham-1}  
  \end{subfigure}
  \begin{subfigure}[t]{0.45\linewidth}
    \centering
    \includegraphics[width=\linewidth]{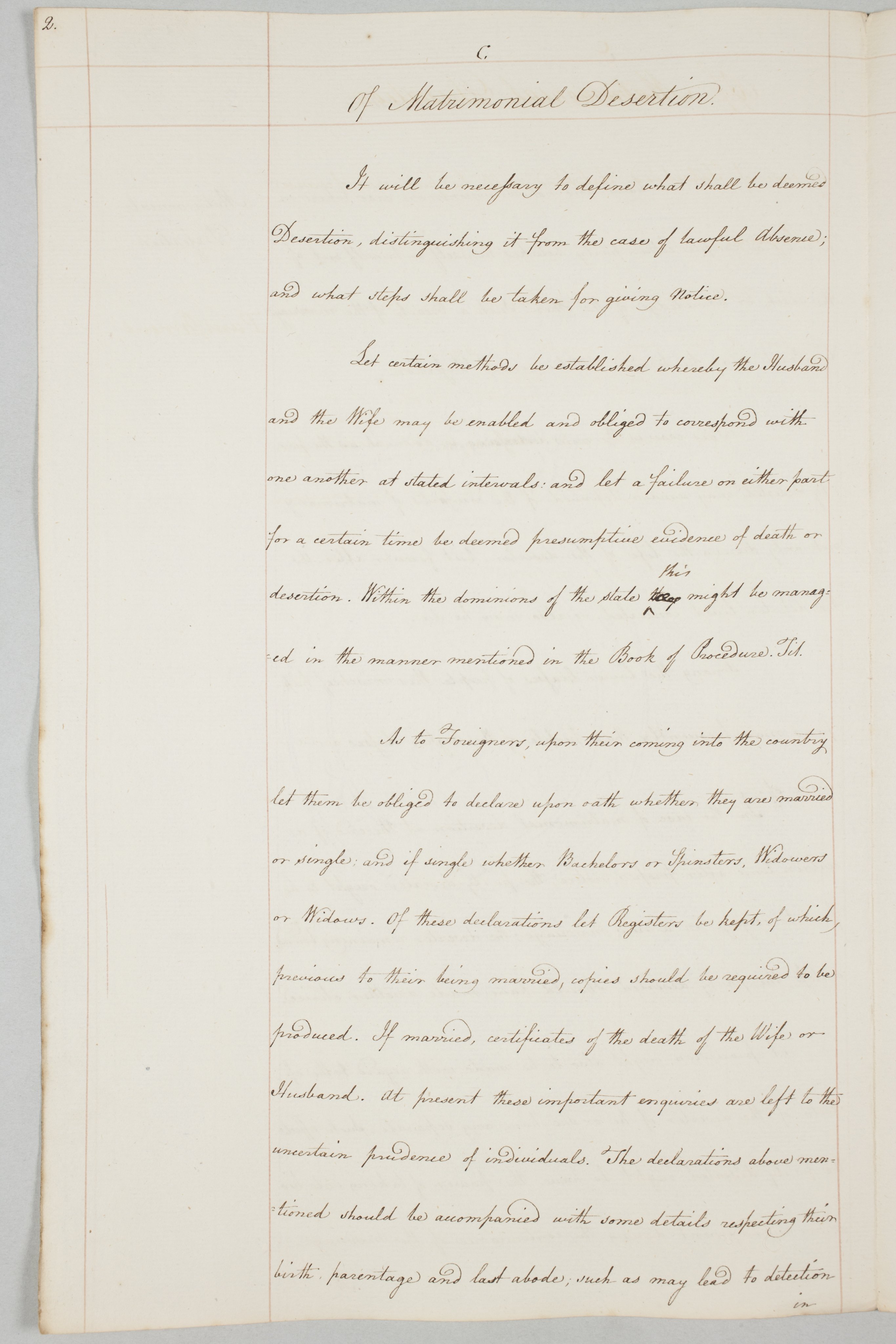} \\[0.5em]
    \caption{}
    \label{fig:example-bentham-2}
  \end{subfigure}
  \caption{Example pages from the \bentham{} dataset.}
  \label{fig:bentham-example-pages}
\end{figure}

Figures~\ref{fig:example-iam-side-by-side}, \ref{fig:mhills-example-pages} \& \ref{fig:bentham-example-pages} show example pages from the \iam{}, \mhills{} and \bentham{} datasets respectively.

\newpage

\section{Examples of \firstpage{} corrections}
\label{sec:firstpage_examples}
\begin{figure}[H]
    \centering
    \includegraphics[width=0.6\linewidth]{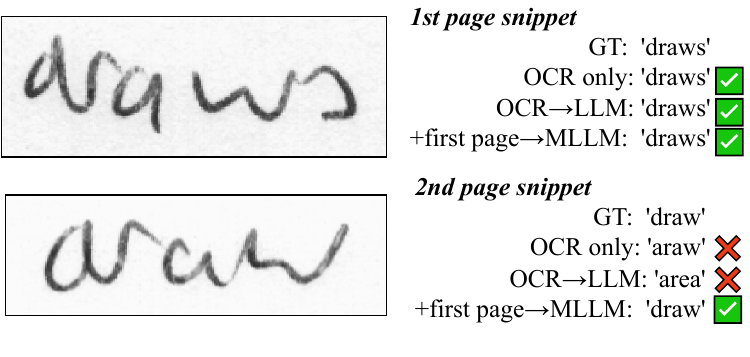}
    \caption{With \firstpage{}, the correctly-transcribed occurrence of `draws' in the first page can be extrapolated to the unseen `draw' on the second page.}
    \label{fig:example_draws}
\end{figure}

\begin{figure}[H]
    \centering
    \includegraphics[width=0.6\linewidth]{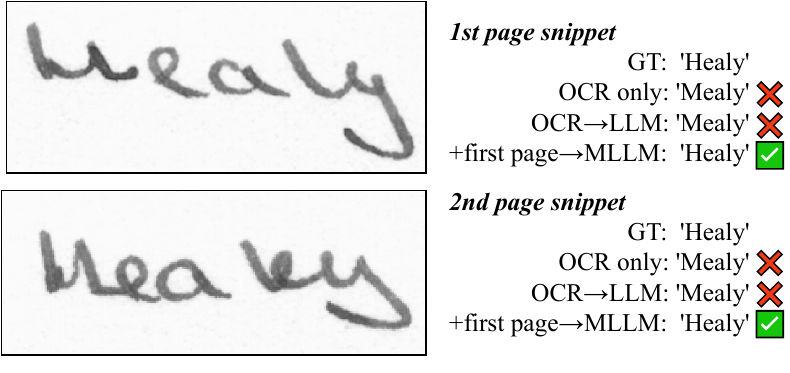}
    \caption{With \firstpage{}, the correctly-transcribed occurrence of the name `Mr Healy' in the first page can be extrapolated to the unseen occurrence on the second page.}
    \label{fig:example_healy}
\end{figure}

\begin{figure}[H]
    \centering
    \includegraphics[width=0.6\linewidth]{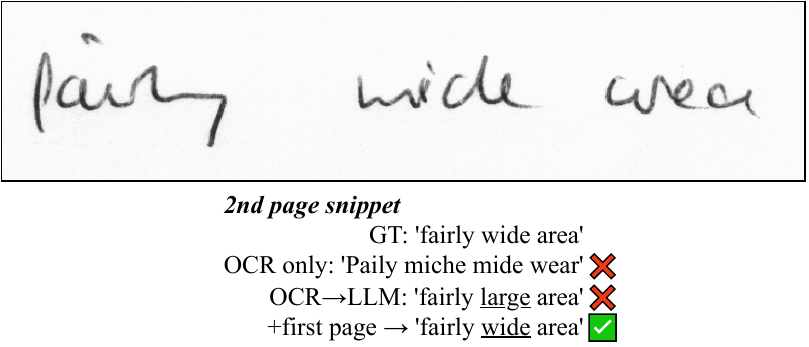}
    \caption{\firstpage{} corrects where \ocr{}$\rightarrow$LLM gets it wrong, despite \textit{only having access only to the garbled OCR output} and not the image of the word `wide' shown above. Suggests some degree of reasoning using the seemingly irrelevant first page text --- i.e. it can see that `m's on page 1 look similar to `w's and reason that `mide' could be `wide'.}
    \label{fig:example_wide}
\end{figure}

\begin{figure}[H]
    \centering
    \includegraphics[width=\linewidth]{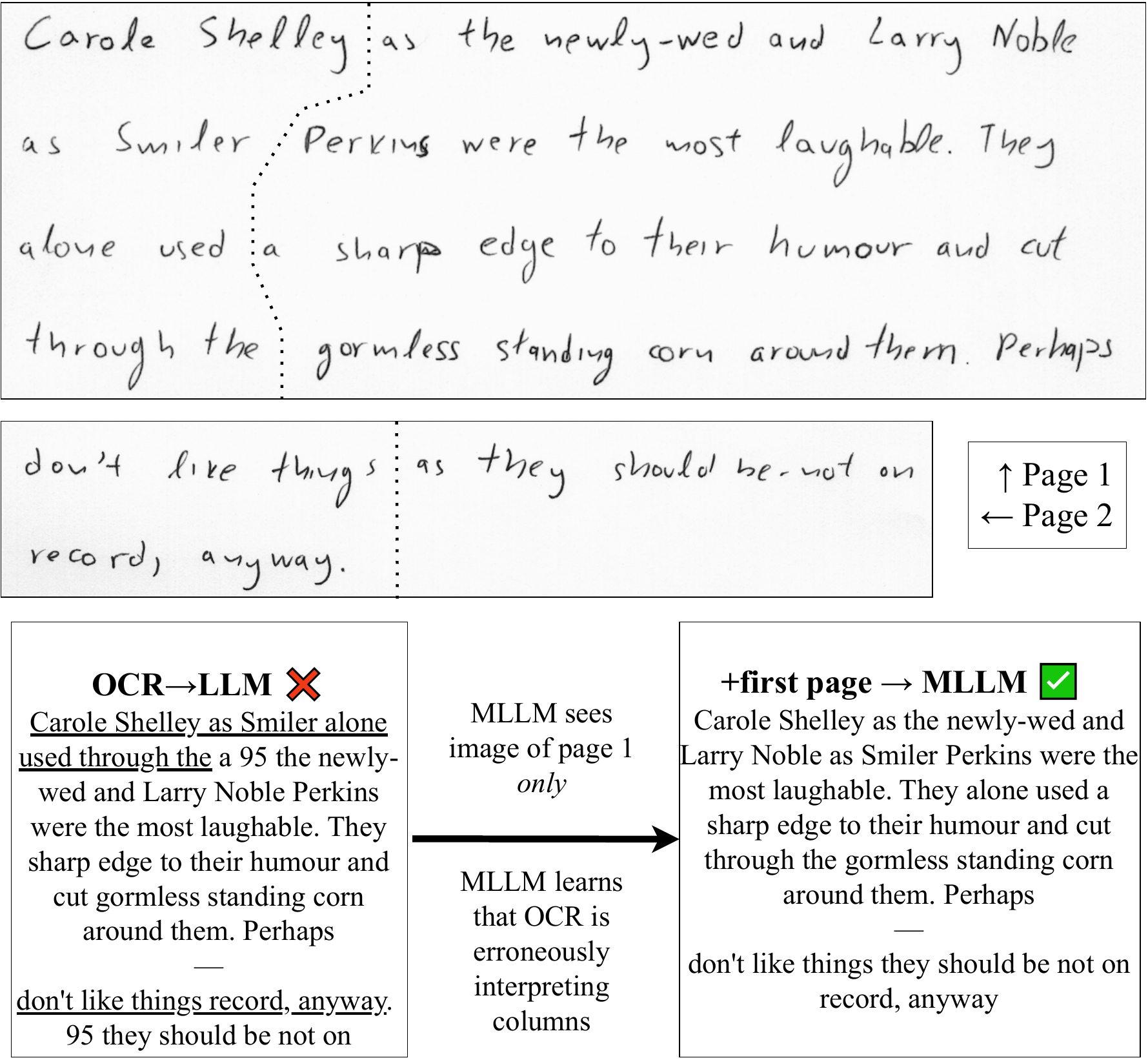}
    \caption{An unusual case where the OCR engine erroneously breaks the text into two columns for both pages. This is shown using the dotted line and underlining of the transcribed text. OCR$\rightarrow$LLM preserves this error. \firstpage{} trivially corrects it in the first page --- it has access to the image --- \textit{but also corrects it in the second page}, even though it has no access to that image. \firstpage{} correctly rearranges the text on page 2 using only context and the inferred formatting from page 1.}
    \label{fig:example_columns}
\end{figure}

See Figures~\ref{fig:example_draws}--\ref{fig:example_columns}. All examples are on \iam{}, use Google Cloud Vision as the OCR engine and \textsc{gpt-4o}.

\newpage
\section{Additional experimental details}

\subsection{Token counting and image tokenization policy}
\label{sec:app:tokenization}
Tokenization policy for OpenAI models is based on the \texttt{tiktoken} Python library. For \textsc{gemini} and \textsc{gemma}, text token counts are computed using the \texttt{count\_tokens} method packaged with the Gemini API, and image tokens are counted with a custom script that implements the image tokenization policy described in the Gemini API documentation.

\subsection{Semantic evaluation of transcription accuracy with an LLM}
\label{sec:app:semantic-error-eval}
For each transcribed page we generate a text diff from the ground-truth using \texttt{<ins>...<\textbackslash ins>} and \texttt{<del>...<\textbackslash del>} tags, and pass it to \textsc{gemini-2.5-flash} with a prompt asking to return a JSON-structured list of all errors, each classified into one of 7 error types. `Formatting' and `semantic' errors (those which do not affect meaning or remove information from the document) are `minor' errors; `missing content', `hallucination', `proper noun error', `numeric error' and `other' are `major' errors (those which impede understanding or remove information). This method is naturally imperfect, but we found that for a random sample of 5 documents, error classifications were accurate about 95\% of the time.

Table~\ref{tab:iam-errors} shows the full set of error classifications for \iam{}, as described in Section~\ref{sec:llm-error-eval} and summarised in Table~\ref{tab:iam}. For conciseness, the table combines the `proper noun' and `numerical' error types into one column, and combines the two minor error types, `formatting' and `semantic', into the `minor' error type.
The error types are defined in the prompt used to extract them as follows:

\begin{verbatim}
Error types (choose exactly one per error):
1) missing_content
  - A deletion with no suitable inserted counterpart. 
  Example: `<del>word</del>` with no matching `<ins>word</ins>`.
  - Minor missing content, such as single punctuation marks, 
  should be `formatting`.
2) hallucination
  - An insertion with no corresponding gt (including numbers 
  that appear only in pred).
  - Minor hallucinations, such as single punctuation marks, 
  should be `formatting`.
3) mistake_proper_noun
  - gt is a proper noun (possibly multi-word: full or partial 
  names, places, institutions, titles) and pred misrenders it 
  (letters/wording/order). 
  - Pure capitalization changes that don’t create a different name 
  are semantic.
  - Extended 'proper names', such as full names, should be 
  treated in their entirety, e.g. 'A. B. Smith'
4) mistake_numerical
  - Use only when gt contains a number/date/roman numeral/measure 
  and pred changes its value or structure or fails to include it. 
  - Combine obviously linked components (e.g., an entire date 
  like 18 March 1884) into one numerical error.
5) mistake_other
  - Non–proper-noun, non-numeric word/phrase error that affects 
  normal reading (e.g., genuine misspelling or wrong lemma/word).
  - Use sparingly, only if none of the other error types are 
  appropriate, not as a catch-all
6) semantic
  - Differences that are clearly meaning-preserving or trivial:
    - minor letter/spelling differences where the word couldn’t 
    reasonably be mistaken for another,
    - alternative/modernized spellings or contractions with 
    the same meaning
    - capitalization-only changes,
7) formatting
  - Minor erroneous punctuation, misplaced or missing newlines, 
  misplaced structural content including text that has been moved 
  from one place in the transcription to another
  - If a moved block contains internal real mistakes (e.g., a 
  wrong year), add separate errors for those internal pieces with 
  the appropriate types.
\end{verbatim}

\subsection{Why isn't \chosenpage{} always better than \firstpage{}?}
\label{sec:app:pageN-vs-page1}
While \chosenpage{} often performs better than \firstpage{}, this is not consistent over MLLMs or datasets. We attribute this to the added prompt complexity of extrapolating from an arbitrary Nth page rather than the 1st, as this is the only difference between the two methods. There is evidence that MLLMs are sensitive to prompt order \citep{guan2025order}, so it is not unreasonable to suppose that the prompt ordering for \firstpage{}, which is `page 1 image, page 1 OCR text, page 2 OCR text, ...', where the corresponding image and text are adjacent in the prompt, is easier for an MLLM to follow than `page N image, page 1 OCR text, ..., page N OCR text, ...'. The method relies on learning a mapping from the page N image to the page N OCR, and MLLMs are known to be better at local context reasoning than long-context \citep{liu2025comprehensive}. In the case where the first page is approximately as informative as the Nth, the complexity this prompt arrangement introduces may outweigh any marginal benefit from the page ID choice. It is possible that further prompt tuning, such as rearranging the image within the prompt, could mitigate this; we leave it for future work.

This effect appears to be more model-dependent than document-length-dependent: for \mhills{}, \chosenpage{} improves on \firstpage{} for \textsc{gpt-4o} and vice versa for \textsc{gemini-2.5-pro}, and this holds across the 2-3-, 5-plus-, and 10-plus-page versions of \mhills{} (Table~\ref{tab:pagen-vs-page1}), where \chosenpage{} has progressively more page-selection options. We also note that for two-page documents, \chosenpage{}'s upstream page choice is inherently limited, and it defaults to \firstpage{}'s behavior whenever the first page is selected; the choice can nonetheless be meaningful even here --- for example, when the first page is a sparse title page, selecting page 2 can provide substantially more useful handwriting context.

\begin{table}[h]
\centering
\small
\caption{\firstpage{} vs.\ \chosenpage{} CER (\%) on \mhills{}, by model and document length. Better of each pair is bold. Full results (all methods) are in Table~\ref{tab:mhills} (2--3 pages) and Appendix Tables~\ref{tab:mhills-5p-rebuttal} \& \ref{tab:mhills-10p-rebuttal} (5+, 10+ pages).}
\label{tab:pagen-vs-page1}

\begin{tabular}{@{}lcccc@{}}
\toprule
& \multicolumn{2}{c}{\textsc{gpt-4o}} & \multicolumn{2}{c}{\textsc{gemini-2.5-pro}} \\
\textbf{Document length} & \firstpage{} & \chosenpage{} & \firstpage{} & \chosenpage{} \\ \midrule
2--3 pages & 8.92 & \textbf{8.05} & \textbf{5.83} & 6.54 \\
5+ pages & 9.12 & \textbf{7.59} & \textbf{5.43} & 5.86 \\
10+ pages & 21.38 & \textbf{16.37} & \textbf{4.76} & 5.41 \\ \bottomrule
\end{tabular}

\end{table}

\subsection{Variance across API seeds}
\label{sec:app:variance}
To directly test the reviewer concern that single-run, temperature-zero evaluation may be unreliable, we repeated \firstpage{} and \chosenpage{} (\textsc{gemini-2.5-pro}) across 3 additional API seeds on \iam{} and \mhills{}, and additionally repeated \textsc{images} and \textsc{ocr+images} on \mhills{} to confirm the comparative ranking is not an artifact of a single run. Table~\ref{tab:variance} reports per-seed CER, the 4-seed mean (including the original published run as seed 0), and 95\% confidence intervals.

On \iam{}, variance is negligible: 95\% CI half-widths are $\pm$0.005\% (\firstpage{}) and $\pm$0.024\% (\chosenpage{}), and the published values sit within 0.02\% of the 4-seed mean.

On \mhills{}, our most challenging dataset, we observe real run-to-run variance (95\% CI half-widths of $\pm$0.09--0.54\%, depending on method) --- \textsc{gemini-2.5-pro} is not perfectly deterministic at temperature zero on harder inputs. Despite this, the 4-seed mean ranking is unchanged, confirming that our central finding --- that a single page image can capture most of the benefit of the full image set --- is not an artifact of single-run evaluation.

\begin{table}[h]
\centering
\small
\caption{Variance across 4 API seeds for \textsc{gemini-2.5-pro}. CIs are 95\%, $t$-distribution, 3 degrees of freedom.}
\label{tab:variance}

\begin{tabular}{@{}llcccccc@{}}
\toprule
\textbf{Dataset} & \textbf{Method} & \textbf{seed 0} & \textbf{1} & \textbf{2} & \textbf{3} & \textbf{Mean} & \textbf{95\% CI $\pm$} \\ \midrule
\iam{} (107 docs) & \firstpage{} & 0.648 & 0.645 & 0.642 & 0.648 & 0.646 & 0.005 \\
\iam{} (107 docs) & \chosenpage{} & 0.633 & 0.660 & 0.655 & 0.668 & 0.654 & 0.024 \\
\midrule
\mhills{} (70 docs) & \firstpage{} & 5.826 & 5.966 & 6.609 & 6.121 & 6.130 & 0.543 \\
\mhills{} (70 docs) & \chosenpage{} & 6.544 & 7.343 & 6.743 & 6.894 & 6.881 & 0.541 \\
\mhills{} (70 docs) & \textsc{images} & 6.417 & 6.545 & 6.527 & 6.527 & 6.504 & 0.093 \\
\mhills{} (70 docs) & \textsc{ocr+images} & 6.456 & 6.424 & 6.244 & 6.390 & 6.379 & 0.149 \\ \bottomrule
\end{tabular}

\end{table}

\begin{table}[h]
\caption{IAM error types for all runs in Table~\ref{tab:iam}, evaluated by \textsc{gemini-2.5-flash}. Each cell contains the average number of occurrences of that type of error in a transcription produced by the given method, and the percentage of total errors in a transcription that are that particular error type. E.g., Azure OCR output alone has an average of 5.1 major errors per transcription for \iam{} documents, but 41.4\% of total errors inn Azure transcription are minor, i.e. semantic or formatting-related only.}
\label{tab:iam-errors}
\centering
\small
\begin{tabular}{@{}rlcccccc@{}}
\toprule
\textbf{} &
  \textbf{Input} &
  \textbf{\begin{tabular}[c]{@{}c@{}}Major\\ (all)\end{tabular}} &
  \textbf{\begin{tabular}[c]{@{}c@{}}Missing\\ content\end{tabular}} &
  \textbf{\begin{tabular}[c]{@{}c@{}}Halluc\\ -ination\end{tabular}} &
  \textbf{\begin{tabular}[c]{@{}c@{}}Proper\\noun\\/Numeric\end{tabular}} &
  \textbf{\begin{tabular}[c]{@{}c@{}}Other\\ error\end{tabular}} &
  \textbf{\begin{tabular}[c]{@{}c@{}}Minor\\ error\end{tabular}} \\ \midrule
\textsc{---} &
  \textsc{ocr} &
  \begin{tabular}[c]{@{}c@{}}5.1\\ (58.6\%)\end{tabular} &
  \begin{tabular}[c]{@{}c@{}}0.3\\ (3.1\%)\end{tabular} &
  \begin{tabular}[c]{@{}c@{}}0.2\\ (1.9\%)\end{tabular} &
  \begin{tabular}[c]{@{}c@{}}0.8\\ (9.0\%)\end{tabular} &
  \begin{tabular}[c]{@{}c@{}}3.9\\ (44.6\%)\end{tabular} &
  \begin{tabular}[c]{@{}c@{}}3.6\\ (41.4\%)\end{tabular} \\ \midrule  
\multirow{6}{*}{\raisebox{-2.5em}[0pt][0pt]{\rotatebox[origin=c]{90}{\textsc{gemma-3-27b}}}}
 &
  \textsc{ocr} &
  \begin{tabular}[c]{@{}c@{}}3.2\\ (47.9\%)\end{tabular} &
  \begin{tabular}[c]{@{}c@{}}0.3\\ (4.2\%)\end{tabular} &
  \begin{tabular}[c]{@{}c@{}}0.2\\ (2.4\%)\end{tabular} &
  \begin{tabular}[c]{@{}c@{}}0.6\\ (8.8\%)\end{tabular} &
  \begin{tabular}[c]{@{}c@{}}2.1\\ (32.5\%)\end{tabular} &
  \begin{tabular}[c]{@{}c@{}}3.4\\ (52.1\%)\end{tabular} \\
 &
  \textsc{images} &
  \begin{tabular}[c]{@{}c@{}}4.3\\ (52.1\%)\end{tabular} &
  \begin{tabular}[c]{@{}c@{}}0.2\\ (2.7\%)\end{tabular} &
  \begin{tabular}[c]{@{}c@{}}0.2\\ (1.9\%)\end{tabular} &
  \begin{tabular}[c]{@{}c@{}}1.1\\ (13.0\%)\end{tabular} &
  \begin{tabular}[c]{@{}c@{}}2.9\\ (34.5\%)\end{tabular} &
  \begin{tabular}[c]{@{}c@{}}4.0\\ (47.9\%)\end{tabular} \\
 &
  \textsc{images-all-at-once} &
  \begin{tabular}[c]{@{}c@{}}4.0\\ (60.4\%)\end{tabular} &
  \begin{tabular}[c]{@{}c@{}}0.2\\ (3.3\%)\end{tabular} &
  \begin{tabular}[c]{@{}c@{}}0.1\\ (1.8\%)\end{tabular} &
  \begin{tabular}[c]{@{}c@{}}0.9\\ (13.8\%)\end{tabular} &
  \begin{tabular}[c]{@{}c@{}}2.8\\ (41.4\%)\end{tabular} &
  \begin{tabular}[c]{@{}c@{}}2.7\\ (39.6\%)\end{tabular} \\
 &
  \textsc{ocr+images} &
  \begin{tabular}[c]{@{}c@{}}2.7\\ (43.6\%)\end{tabular} &
  \begin{tabular}[c]{@{}c@{}}0.2\\ (3.5\%)\end{tabular} &
  \begin{tabular}[c]{@{}c@{}}0.1\\ (1.5\%)\end{tabular} &
  \begin{tabular}[c]{@{}c@{}}0.5\\ (8.2\%)\end{tabular} &
  \begin{tabular}[c]{@{}c@{}}1.9\\ (30.3\%)\end{tabular} &
  \begin{tabular}[c]{@{}c@{}}3.4\\ (56.4\%)\end{tabular} \\
 &
  \textsc{ocr+pageN} &
  \begin{tabular}[c]{@{}c@{}}2.2\\ (44.5\%)\end{tabular} &
  \begin{tabular}[c]{@{}c@{}}0.2\\ (4.7\%)\end{tabular} &
  \begin{tabular}[c]{@{}c@{}}0.1\\ (3.0\%)\end{tabular} &
  \begin{tabular}[c]{@{}c@{}}0.5\\ (9.7\%)\end{tabular} &
  \begin{tabular}[c]{@{}c@{}}1.4\\ (27.1\%)\end{tabular} &
  \begin{tabular}[c]{@{}c@{}}2.8\\ (55.5\%)\end{tabular} \\
 &
  \textsc{ocr+page1} &
  \begin{tabular}[c]{@{}c@{}}2.0\\ (45.7\%)\end{tabular} &
  \begin{tabular}[c]{@{}c@{}}0.3\\ (5.7\%)\end{tabular} &
  \begin{tabular}[c]{@{}c@{}}0.1\\ (3.2\%)\end{tabular} &
  \begin{tabular}[c]{@{}c@{}}0.4\\ (8.4\%)\end{tabular} &
  \begin{tabular}[c]{@{}c@{}}1.3\\ (28.4\%)\end{tabular} &
  \begin{tabular}[c]{@{}c@{}}2.4\\ (54.3\%)\end{tabular} \\ \midrule
\multirow{6}{*}{\raisebox{-2.5em}[0pt][0pt]{\rotatebox[origin=c]{90}{\textsc{gpt-4o}}}}
 &
  \textsc{ocr} &
  \begin{tabular}[c]{@{}c@{}}1.9\\ (39.8\%)\end{tabular} &
  \begin{tabular}[c]{@{}c@{}}0.2\\ (4.4\%)\end{tabular} &
  \textbf{\begin{tabular}[c]{@{}c@{}}0.0\\ (1.0\%)\end{tabular}} &
  \begin{tabular}[c]{@{}c@{}}0.5\\ (10.4\%)\end{tabular} &
  \begin{tabular}[c]{@{}c@{}}1.1\\ (24.0\%)\end{tabular} &
  \begin{tabular}[c]{@{}c@{}}2.8\\ (60.2\%)\end{tabular} \\
 &
  \textsc{images} &
  \begin{tabular}[c]{@{}c@{}}1.8\\ (49.6\%)\end{tabular} &
  \begin{tabular}[c]{@{}c@{}}0.2\\ (5.7\%)\end{tabular} &
  \begin{tabular}[c]{@{}c@{}}0.1\\ (1.6\%)\end{tabular} &
  \begin{tabular}[c]{@{}c@{}}0.5\\ (13.2\%)\end{tabular} &
  \begin{tabular}[c]{@{}c@{}}1.1\\ (29.2\%)\end{tabular} &
  \begin{tabular}[c]{@{}c@{}}1.8\\ (50.4\%)\end{tabular} \\
 &
  \textsc{images-all-at-once} &
  \begin{tabular}[c]{@{}c@{}}1.6\\ (43.6\%)\end{tabular} &
  \begin{tabular}[c]{@{}c@{}}0.2\\ (4.9\%)\end{tabular} &
  \begin{tabular}[c]{@{}c@{}}0.1\\ (1.5\%)\end{tabular} &
  \begin{tabular}[c]{@{}c@{}}0.4\\ (11.3\%)\end{tabular} &
  \begin{tabular}[c]{@{}c@{}}0.9\\ (25.9\%)\end{tabular} &
  \begin{tabular}[c]{@{}c@{}}2.1\\ (56.4\%)\end{tabular} \\
 &
  \textsc{ocr+images} &
  \textit{\begin{tabular}[c]{@{}c@{}}1.2\\ (42.2\%)\end{tabular}} &
  \begin{tabular}[c]{@{}c@{}}0.2\\ (7.1\%)\end{tabular} &
  \textbf{\begin{tabular}[c]{@{}c@{}}0.0\\ (1.3\%)\end{tabular}} &
  \begin{tabular}[c]{@{}c@{}}0.3\\ (8.7\%)\end{tabular} &
  \begin{tabular}[c]{@{}c@{}}0.7\\ (25.0\%)\end{tabular} &
  \begin{tabular}[c]{@{}c@{}}1.7\\ (57.8\%)\end{tabular} \\
 &
  \textsc{ocr+pageN} &
  \begin{tabular}[c]{@{}c@{}}1.3\\ (39.8\%)\end{tabular} &
  \begin{tabular}[c]{@{}c@{}}0.2\\ (5.5\%)\end{tabular} &
  \textbf{\begin{tabular}[c]{@{}c@{}}0.0\\ (0.9\%)\end{tabular}} &
  \begin{tabular}[c]{@{}c@{}}0.3\\ (7.8\%)\end{tabular} &
  \begin{tabular}[c]{@{}c@{}}0.8\\ (25.6\%)\end{tabular} &
  \begin{tabular}[c]{@{}c@{}}2.0\\ (60.2\%)\end{tabular} \\
 &
  \textsc{ocr+page1} &
  \textit{\begin{tabular}[c]{@{}c@{}}1.2\\ (39.1\%)\end{tabular}} &
  \textbf{\begin{tabular}[c]{@{}c@{}}0.1\\ (4.9\%)\end{tabular}} &
  \textbf{\begin{tabular}[c]{@{}c@{}}0.0\\ (1.5\%)\end{tabular}} &
  \begin{tabular}[c]{@{}c@{}}0.3\\ (8.6\%)\end{tabular} &
  \begin{tabular}[c]{@{}c@{}}0.7\\ (24.2\%)\end{tabular} &
  \begin{tabular}[c]{@{}c@{}}1.9\\ (60.9\%)\end{tabular} \\ \midrule
\multirow{6}{*}{\raisebox{-2.5em}[0pt][0pt]{\rotatebox[origin=c]{90}{\textsc{gemini-2.5-pro}}}}
 &
  \textsc{ocr} &
  \textit{\begin{tabular}[c]{@{}c@{}}1.2\\ (32.8\%)\end{tabular}} &
  \textbf{\begin{tabular}[c]{@{}c@{}}0.1\\ (4.0\%)\end{tabular}} &
  \begin{tabular}[c]{@{}c@{}}0.1\\ (2.8\%)\end{tabular} &
  \begin{tabular}[c]{@{}c@{}}0.2\\ (6.3\%)\end{tabular} &
  \begin{tabular}[c]{@{}c@{}}0.7\\ (19.8\%)\end{tabular} &
  \begin{tabular}[c]{@{}c@{}}2.5\\ (67.2\%)\end{tabular} \\
 &
  \textsc{images} &
  \begin{tabular}[c]{@{}c@{}}1.5\\ (47.1\%)\end{tabular} &
  \begin{tabular}[c]{@{}c@{}}0.2\\ (6.6\%)\end{tabular} &
  \begin{tabular}[c]{@{}c@{}}0.2\\ (5.4\%)\end{tabular} &
  \begin{tabular}[c]{@{}c@{}}0.3\\ (8.1\%)\end{tabular} &
  \begin{tabular}[c]{@{}c@{}}0.8\\ (27.0\%)\end{tabular} &
  \textbf{\begin{tabular}[c]{@{}c@{}}1.6\\ (52.9\%)\end{tabular}} \\
 &
  \textsc{images-all-at-once} &
  \begin{tabular}[c]{@{}c@{}}1.4\\ (45.2\%)\end{tabular} &
  \begin{tabular}[c]{@{}c@{}}0.2\\ (7.4\%)\end{tabular} &
  \begin{tabular}[c]{@{}c@{}}0.1\\ (2.8\%)\end{tabular} &
  \begin{tabular}[c]{@{}c@{}}0.2\\ (8.0\%)\end{tabular} &
  \begin{tabular}[c]{@{}c@{}}0.8\\ (27.1\%)\end{tabular} &
  \begin{tabular}[c]{@{}c@{}}1.7\\ (54.8\%)\end{tabular} \\
 &
  \textsc{ocr+images} &
  \begin{tabular}[c]{@{}c@{}}1.3\\ (42.7\%)\end{tabular} &
  \begin{tabular}[c]{@{}c@{}}0.2\\ (6.7\%)\end{tabular} &
  \begin{tabular}[c]{@{}c@{}}0.1\\ (3.9\%)\end{tabular} &
  \begin{tabular}[c]{@{}c@{}}0.2\\ (6.7\%)\end{tabular} &
  \begin{tabular}[c]{@{}c@{}}0.8\\ (25.5\%)\end{tabular} &
  \begin{tabular}[c]{@{}c@{}}1.8\\ (57.3\%)\end{tabular} \\
 &
  \textsc{ocr+pageN} &
  \textbf{\begin{tabular}[c]{@{}c@{}}1.0\\ (38.9\%)\end{tabular}} &
  \begin{tabular}[c]{@{}c@{}}0.2\\ (7.7\%)\end{tabular} &
  \begin{tabular}[c]{@{}c@{}}0.1\\ (2.5\%)\end{tabular} &
  \textbf{\begin{tabular}[c]{@{}c@{}}0.1\\ (4.6\%)\end{tabular}} &
  \textbf{\begin{tabular}[c]{@{}c@{}}0.6\\ (24.2\%)\end{tabular}} &
  \textbf{\begin{tabular}[c]{@{}c@{}}1.6\\ (61.1\%)\end{tabular}} \\
 &
  \textsc{ocr+page1} &
  \textbf{\begin{tabular}[c]{@{}c@{}}1.0\\ (35.7\%)\end{tabular}} &
  \begin{tabular}[c]{@{}c@{}}0.2\\ (6.7\%)\end{tabular} &
  \begin{tabular}[c]{@{}c@{}}0.1\\ (2.7\%)\end{tabular} &
  \textbf{\begin{tabular}[c]{@{}c@{}}0.1\\ (5.0\%)\end{tabular}} &
  \textbf{\begin{tabular}[c]{@{}c@{}}0.6\\ (21.3\%)\end{tabular}} &
  \begin{tabular}[c]{@{}c@{}}1.8\\ (64.3\%)\end{tabular} \\ \bottomrule
\end{tabular}
\end{table}

\newpage
\section{Chinese handwriting evaluation: \casia{}}
\label{sec:app:casia}
To synthesize \casia{}, we start with the CASIA-HWDB2.1 test split, which consists of 300 images of multiple lines of handwritten Chinese text. Similar to the process for IAM, we generate multi-page dataset \casia{} by randomly grouping pages with the same author ID, such that handwriting, but not semantic content, is consistent. We use a subset of 30 documents of 5 pages each in our experiments. We use Google Cloud Vision as our OCR engine.

See results in Table~\ref{tab:casia-rebuttal}; abridged results are in Table~\ref{tab:casia} in the main paper.

\begin{table}[h]
\centering
\caption{Results on \casia{} for all methods described in the main text, plus \randompage{}. Our methods, i.e. \pagex{} methods, are bold. Score emphasis: \textbf{1st}, \textit{2nd}, \underline{3rd}. `Fails (/doc)' is the average number of retried API calls required per method (all calls eventually succeeded). For \textsc{gpt-4o}, all retries were due to incomplete/truncated response errors. For \textsc{gemini-2.5-pro}, retries were due to invalid JSON output errors.}
\label{tab:casia-rebuttal}
\small
\begin{tabular}{@{}lcrrrrrrrcc@{}}
\toprule
\multirow{2}{*}{\textbf{Method}} &
  \multirow{2}{*}{\textbf{\begin{tabular}[c]{@{}c@{}}CER $\downarrow$\\ (\%)\end{tabular}}} &
  \multicolumn{4}{c}{\textbf{Cost (\$/1k docs)}} &
  \multicolumn{3}{c}{\textbf{Tokens (k/doc)}} &
  \multicolumn{1}{c}{\multirow{2}{*}{\textbf{\begin{tabular}[c]{@{}c@{}}Time\\ (s/doc)\end{tabular}}}} &
  \multicolumn{1}{c}{\multirow{2}{*}{\textbf{\begin{tabular}[c]{@{}c@{}}Fails\\ (/doc)\end{tabular}}}} \\
 &
   &
  \multicolumn{1}{c}{\textbf{Total}} &
  \multicolumn{1}{c}{\textbf{OCR}} &
  \multicolumn{1}{c}{\textbf{In}} &
  \multicolumn{1}{c}{\textbf{Out}} &
  \multicolumn{1}{c}{\textbf{Image}} &
  \multicolumn{1}{c}{\textbf{In}} &
  \multicolumn{1}{c}{\textbf{Out}} &
  \multicolumn{1}{c}{} &
  \multicolumn{1}{c}{} \\ \midrule
\textsc{ocr} &
  12.84 &
  7.50 &
  7.50 &
  0.00 &
  0.00 &
  0 &
  0 &
  0 &
  --- &
  --- \\ \midrule
\multicolumn{11}{l}{\textsc{gpt-4o}} \\ \midrule
\textsc{ocr} &
  18.32 &
  26.75 &
  7.50 &
  6.13 &
  13.12 &
  0.06 &
  2.40 &
  1.31 &
  36.6 &
  0.00 \\
\textsc{images} &
  42.40 &
  30.81 &
  0.00 &
  16.91 &
  13.90 &
  5.53 &
  1.18 &
  1.39 &
  84.4 &
  0.03 \\
\textsc{images-aao} &
  43.64 &
  29.30 &
  0.00 &
  14.86 &
  14.44 &
  5.53 &
  0.25 &
  1.44 &
  74.8 &
  0.10 \\
\textsc{ocr+images} &
  14.16 &
  41.58 &
  7.50 &
  20.58 &
  13.50 &
  5.58 &
  2.65 &
  1.35 &
  78.5 &
  0.00 \\
\textbf{\textsc{ocr+pageR}} &
  13.91 &
  28.41 &
  7.50 &
  7.51 &
  13.40 &
  1.11 &
  1.77 &
  1.34 &
  34.0 &
  0.00 \\
\textbf{\textsc{ocr+pageN}} &
  12.87 &
  28.68 &
  7.50 &
  7.64 &
  13.55 &
  1.11 &
  1.77 &
  1.35 &
  47.7 &
  0.00 \\
\textbf{\textsc{ocr+page1}} &
  14.48 &
  28.31 &
  7.50 &
  7.45 &
  13.36 &
  1.11 &
  1.74 &
  1.34 &
  44.2 &
  0.00 \\ \midrule
\multicolumn{11}{l}{\textsc{gemini}} \\ \midrule
\textsc{ocr} &
  33.17 &
  24.87 &
  7.50 &
  2.81 &
  14.56 &
  0 &
  2.25 &
  1.46 &
  30.7 &
  0.00 \\
\textsc{images} &
  13.55 &
  19.08 &
  0.00 &
  6.51 &
  12.58 &
  4.00 &
  1.21 &
  1.26 &
  31.1 &
  0.00 \\
\textsc{images-aao} &
  23.07 &
  18.09 &
  0.00 &
  5.32 &
  12.77 &
  4.00 &
  0.26 &
  1.28 &
  18.8 &
  0.00 \\
\textsc{ocr+images} &
  \textbf{8.39} &
  27.80 &
  7.50 &
  8.14 &
  12.16 &
  4.00 &
  2.52 &
  1.22 &
  40.2 &
  0.00 \\
\textbf{\textsc{ocr+pageR}} &
  \underline{10.13} &
  23.19 &
  7.50 &
  3.08 &
  12.61 &
  0.80 &
  1.66 &
  1.26 &
  18.0 &
  0.00 \\
\textbf{\textsc{ocr+pageN}} &
  \textit{8.94} &
  23.42 &
  7.50 &
  3.20 &
  12.72 &
  0.80 &
  1.66 &
  1.27 &
  27.5 &
  0.00 \\
\textbf{\textsc{ocr+page1}} &
  12.61 &
  23.25 &
  7.50 &
  3.05 &
  12.70 &
  0.80 &
  1.64 &
  1.27 &
  48.3 &
  0.07 \\ \bottomrule
\end{tabular}
\end{table}

\paragraph{Discussion.}
For this challenging Chinese handwriting dataset, we see that the best-performing model overall is \textsc{ocr+images}, i.e. a model that has access to both OCR prediction and the full set of document images. The OCR engine on its own is already quite effective, indeed, no method involving GPT-4o improves on it, and many significantly worsen it. The only methods that improve on OCR alone are those that use both OCR and at least one image as input to \textsc{gemini}. The fact that \chosenpage{} and \textsc{ocr+images} provide a similar performance boost over OCR alone supports our findings that a single well-chosen image can provide most of the benefit provided by the full set of images, and at a lower token/overall cost.

\newpage
\section{Additional experiments on longer documents}
\label{sec:app:extended-experiments}

\subsection{\iamp{}}
We generate 5-page documents by the same method we generated \iam{}; by grouping random individual pages by author ID so handwriting (but not necessarily semantic content) is consistent. We use 100 page images to generate 20 5-page documents.

See results in Table~\ref{tab:iam-5p-rebuttal}.

\begin{table}[h]
\centering
\caption{Results on \iamp{} for all methods described in the main text, plus \randompage{}. Our methods are bold. Score emphasis: \textbf{1st}, \textit{2nd}, \underline{3rd}. The single method with failed calls was due to \textsc{gemma} `free resource exhausted' errors.}
\label{tab:iam-5p-rebuttal}
\small
\begin{tabular}{@{}lcrrrrrrrcc@{}}
\toprule
\multirow{2}{*}{\textbf{Method}} &
  \multicolumn{1}{l}{\multirow{2}{*}{\textbf{CER $\downarrow$}}} &
  \multicolumn{4}{c}{\textbf{Cost (\$/1k docs)}} &
  \multicolumn{3}{c}{\textbf{Tokens (k/doc)}} &
  \multicolumn{1}{c}{\multirow{2}{*}{\textbf{\begin{tabular}[c]{@{}c@{}}Time\\ (s/doc)\end{tabular}}}} &
  \multicolumn{1}{c}{\multirow{2}{*}{\textbf{\begin{tabular}[c]{@{}c@{}}Fails\\ (/doc)\end{tabular}}}} \\
 &
  \multicolumn{1}{l}{} &
  \multicolumn{1}{c}{\textbf{Total}} &
  \multicolumn{1}{c}{\textbf{OCR}} &
  \multicolumn{1}{c}{\textbf{In}} &
  \multicolumn{1}{c}{\textbf{Out}} &
  \multicolumn{1}{c}{\textbf{Image}} &
  \multicolumn{1}{c}{\textbf{In}} &
  \multicolumn{1}{c}{\textbf{Out}} &
  \multicolumn{1}{c}{} &
  \multicolumn{1}{c}{} \\ \midrule
\textsc{ocr} &
  3.36 &
  5.00 &
  5.00 &
  0.00 &
  0.00 &
  0 &
  0 &
  0 &
  --- &
  --- \\ \midrule
\multicolumn{11}{l}{\textsc{gemma}} \\ \midrule
\textsc{ocr} &
  2.67 &
  5.34 &
  5.00 &
  0.11 &
  0.23 &
  0 &
  1.54 &
  0.47 &
  14.8 &
  0.00 \\
\textsc{images} &
  1.54 &
  0.65 &
  0.00 &
  0.40 &
  0.26 &
  4.44 &
  1.21 &
  0.51 &
  20.1 &
  0.00 \\
\textsc{images-aao} &
  1.35 &
  0.59 &
  0.00 &
  0.33 &
  0.26 &
  4.44 &
  0.26 &
  0.53 &
  12.7 &
  0.00 \\
\textsc{ocr+images} &
  1.47 &
  5.68 &
  5.00 &
  0.44 &
  0.24 &
  4.44 &
  1.81 &
  0.48 &
  19.5 &
  0.05 \\
\textbf{\textsc{ocr+pageR}} &
  2.69 &
  5.40 &
  5.00 &
  0.13 &
  0.27 &
  0.89 &
  0.94 &
  0.54 &
  12.2 &
  0.00 \\
\textbf{\textsc{ocr+pageN}} &
  2.52 &
  5.51 &
  5.00 &
  0.20 &
  0.31 &
  0.87 &
  0.94 &
  0.54 &
  18.3 &
  0.00 \\
\textbf{\textsc{ocr+page1}} &
  1.22 &
  5.39 &
  5.00 &
  0.12 &
  0.27 &
  0.86 &
  0.91 &
  0.53 &
  12.3 &
  0.00 \\ \midrule
\multicolumn{11}{l}{\textsc{gpt-4o}} \\ \midrule
\textsc{ocr} &
  0.99 &
  13.60 &
  5.00 &
  3.91 &
  4.69 &
  0.06 &
  1.50 &
  0.47 &
  21.6 &
  0.00 \\
\textsc{images} &
  0.54 &
  20.68 &
  0.00 &
  15.94 &
  4.75 &
  5.13 &
  1.18 &
  0.47 &
  60.6 &
  0.00 \\
\textsc{images-aao} &
  0.77 &
  18.60 &
  0.00 &
  13.88 &
  4.72 &
  5.13 &
  0.25 &
  0.47 &
  38.2 &
  0.00 \\
\textsc{ocr+images} &
  0.58 &
  27.04 &
  5.00 &
  17.38 &
  4.66 &
  5.19 &
  1.76 &
  0.47 &
  62.2 &
  0.00 \\
\textbf{\textsc{ocr+pageR}} &
  0.91 &
  14.93 &
  5.00 &
  5.11 &
  4.82 &
  1.02 &
  0.89 &
  0.48 &
  21.9 &
  0.00 \\
\textbf{\textsc{ocr+pageN}} &
  0.83 &
  15.05 &
  5.00 &
  5.19 &
  4.87 &
  1.02 &
  0.89 &
  0.48 &
  26.6 &
  0.00 \\
\textbf{\textsc{ocr+page1}} &
  0.98 &
  14.82 &
  5.00 &
  5.01 &
  4.81 &
  1.00 &
  0.87 &
  0.48 &
  24.3 &
  0.00 \\ \midrule
\multicolumn{11}{l}{\textsc{gemini}} \\ \midrule
\textsc{ocr} &
  1.37 &
  11.51 &
  5.00 &
  1.93 &
  4.58 &
  0 &
  1.54 &
  0.46 &
  11.3 &
  0.00 \\
\textsc{images} &
  \underline{0.52} &
  11.72 &
  0.00 &
  7.06 &
  4.67 &
  4.44 &
  1.21 &
  0.47 &
  21.4 &
  0.00 \\
\textsc{images-aao} &
  0.54 &
  11.18 &
  0.00 &
  5.87 &
  5.31 &
  4.44 &
  0.26 &
  0.53 &
  7.2 &
  0.00 \\
\textsc{ocr+images} &
  \textbf{0.47} &
  17.39 &
  5.00 &
  7.81 &
  4.57 &
  4.44 &
  1.81 &
  0.46 &
  18.1 &
  0.00 \\
\textbf{\textsc{ocr+pageR}} &
  \underline{0.52} &
  12.48 &
  5.00 &
  2.28 &
  5.20 &
  0.89 &
  0.94 &
  0.52 &
  8.3 &
  0.00 \\
\textbf{\textsc{ocr+pageN}} &
  \textit{0.51} &
  12.59 &
  5.00 &
  2.34 &
  5.25 &
  0.87 &
  0.94 &
  0.52 &
  13.5 &
  0.00 \\
\textbf{\textsc{ocr+page1}} &
  \textbf{0.47} &
  12.40 &
  5.00 &
  2.21 &
  5.19 &
  0.86 &
  0.91 &
  0.52 &
  7.9 &
  0.00 \\ \bottomrule
\end{tabular}
\end{table}

\paragraph{Discussion.} We see similar results for the longer document case of \iamp{} as we did for \iam{}. \firstpage{} remains the (joint) top-performing model and all OCR + single-image methods perform approximately as well as the full \textsc{ocr+images} method, despite lacking access to \textit{80\% of images} (and the comparatively poor performance of OCR alone), and correspondingly have lower costs and inference times.

\subsection{\iamr{}; inconsistent handwriting \textit{and} semantic content ablation}
\label{sec:app:iam-random}
\iamr{} is generated in the same way as \iamp{}, but we ensure that all 20 generated documents contain pages from 5 different authors.

See results in Table~\ref{tab:iam-5p-random-rebuttal}.

\begin{table}[h]
\centering
\caption{Results on \iamr{} for all methods described in the main text, plus \randompage{}. Our methods are bold. Score emphasis: \textbf{1st}, \textit{2nd}, \underline{3rd}.}
\label{tab:iam-5p-random-rebuttal}
\small
\begin{tabular}{@{}lcrrrrrrrcc@{}}
\toprule
\multirow{2}{*}{\textbf{Method}} &
  \multicolumn{1}{l}{\multirow{2}{*}{\textbf{CER $\downarrow$}}} &
  \multicolumn{4}{c}{\textbf{Cost (\$/1k docs)}} &
  \multicolumn{3}{c}{\textbf{Tokens (k/doc)}} &
  \multicolumn{1}{c}{\multirow{2}{*}{\textbf{\begin{tabular}[c]{@{}c@{}}Time\\ (s/doc)\end{tabular}}}} &
  \multicolumn{1}{c}{\multirow{2}{*}{\textbf{\begin{tabular}[c]{@{}c@{}}Fails\\(/doc)\end{tabular}}}} \\
 &
  \multicolumn{1}{l}{} &
  \multicolumn{1}{c}{\textbf{Total}} &
  \multicolumn{1}{c}{\textbf{OCR}} &
  \multicolumn{1}{c}{\textbf{In}} &
  \multicolumn{1}{c}{\textbf{Out}} &
  \multicolumn{1}{c}{\textbf{Image}} &
  \multicolumn{1}{c}{\textbf{In}} &
  \multicolumn{1}{c}{\textbf{Out}} &
  \multicolumn{1}{c}{} &
  \multicolumn{1}{c}{} \\ \midrule
\textsc{ocr} &
  3.78 &
  5.00 &
  5.00 &
  --- &
  --- &
  --- &
  --- &
  --- &
  --- &
  --- \\ \midrule
\multicolumn{11}{l}{\textsc{gemma}} \\ \midrule
\textsc{ocr} &
  2.88 &
  5.34 &
  5.00 &
  0.11 &
  0.23 &
  0 &
  1.55 &
  0.47 &
  13.8 &
  0.0 \\
\textsc{images} &
  2.34 &
  0.67 &
  0.00 &
  0.41 &
  0.26 &
  4.67 &
  1.21 &
  0.51 &
  20.7 &
  0.0 \\
\textsc{images-aao} &
  2.04 &
  0.61 &
  0.00 &
  0.35 &
  0.26 &
  4.67 &
  0.26 &
  0.53 &
  12.6 &
  0.0 \\
\textsc{ocr+images} &
  1.75 &
  5.70 &
  5.00 &
  0.45 &
  0.24 &
  4.67 &
  1.82 &
  0.48 &
  17.2 &
  0.0 \\
\textbf{\textsc{ocr+pageR}} &
  3.14 &
  5.40 &
  5.00 &
  0.13 &
  0.27 &
  0.91 &
  0.94 &
  0.54 &
  11.8 &
  0.0 \\
\textbf{\textsc{ocr+pageN}} &
  3.10 &
  5.51 &
  5.00 &
  0.20 &
  0.31 &
  0.90 &
  0.94 &
  0.54 &
  18.0 &
  0.0 \\
\textbf{\textsc{ocr+page1}} &
  1.53 &
  5.39 &
  5.00 &
  0.13 &
  0.27 &
  0.91 &
  0.92 &
  0.53 &
  11.7 &
  0.0 \\ \midrule
\multicolumn{11}{l}{\textsc{gpt-4o}} \\ \midrule
\textsc{ocr} &
  1.09 &
  13.66 &
  5.00 &
  3.92 &
  4.73 &
  0.06 &
  1.51 &
  0.47 &
  21.4 &
  0.0 \\
\textsc{images} &
  0.91 &
  21.47 &
  0.00 &
  16.66 &
  4.82 &
  5.46 &
  1.18 &
  0.48 &
  56.8 &
  0.0 \\
\textsc{images-aao} &
  1.20 &
  19.36 &
  0.00 &
  14.60 &
  4.76 &
  5.46 &
  0.25 &
  0.48 &
  35.6 &
  0.0 \\
\textsc{ocr+images} &
  \underline{0.71} &
  27.83 &
  5.00 &
  18.12 &
  4.71 &
  5.48 &
  1.77 &
  0.47 &
  55.0 &
  0.0 \\
\textbf{\textsc{ocr+pageR}} &
  1.01 &
  15.05 &
  5.00 &
  5.21 &
  4.84 &
  1.05 &
  0.90 &
  0.48 &
  20.3 &
  0.0 \\
\textbf{\textsc{ocr+pageN}} &
  0.92 &
  15.18 &
  5.00 &
  5.29 &
  4.90 &
  1.05 &
  0.90 &
  0.49 &
  25.7 &
  0.0 \\
\textbf{\textsc{ocr+page1}} &
  1.08 &
  15.03 &
  5.00 &
  5.19 &
  4.83 &
  1.07 &
  0.87 &
  0.48 &
  19.8 &
  0.0 \\ \midrule
\multicolumn{11}{l}{\textsc{gemini}} \\ \midrule
\textsc{ocr} &
  1.56 &
  11.54 &
  5.00 &
  1.93 &
  4.60 &
  0 &
  1.55 &
  0.46 &
  10.9 &
  0.0 \\
\textsc{images} &
  0.76 &
  12.08 &
  0.00 &
  7.35 &
  4.73 &
  4.67 &
  1.21 &
  0.47 &
  21.6 &
  0.0 \\
\textsc{images-aao} &
  0.75 &
  11.43 &
  0.00 &
  6.17 &
  5.27 &
  4.67 &
  0.26 &
  0.53 &
  6.7 &
  0.0 \\
\textsc{ocr+images} &
  \textbf{0.64} &
  17.72 &
  5.00 &
  8.11 &
  4.61 &
  4.67 &
  1.82 &
  0.46 &
  18.7 &
  0.0 \\
\textbf{\textsc{ocr+pageR}} &
  0.80 &
  12.55 &
  5.00 &
  2.32 &
  5.23 &
  0.91 &
  0.94 &
  0.52 &
  7.6 &
  0.0 \\
\textbf{\textsc{ocr+pageN}} &
  \textit{0.70} &
  12.66 &
  5.00 &
  2.38 &
  5.28 &
  0.90 &
  0.94 &
  0.52 &
  13.4 &
  0.0 \\
\textbf{\textsc{ocr+page1}} &
  1.31 &
  12.49 &
  5.00 &
  2.28 &
  5.21 &
  0.91 &
  0.92 &
  0.52 &
  7.0 &
  0.0 \\ \bottomrule
\end{tabular}
\end{table}

\paragraph{Discussion.} As expected, the benefit of OCR+single-image methods is less pronounced when there is neither consistent handwriting nor semantic content across pages --- \textit{but there is still some benefit}. Though \firstpage{} underperforms, \textsc{+pageR} and \textsc{+pageN} are perform comparably with methods that include access to all images, at lower cost and faster inference time. These results demonstrate, firstly, that our intuition about the benefit of \textsc{+pageX} methods is likely correct; the more similar pages in a document are (in content, writing, etc.) the more performance gain can be achieved with only a single page (and vice versa); secondly, that even in cases where pages are quite different, a single page can still provide performance benefit. This may be a result of other page similarities (e.g. image quality, document type), or an example of multi-modal inputs assisting with reasoning in general as a version of in-context learning.

\subsection{\mhillsp{}}
We group consecutive pages (i.e. pages that are continuous sections from the same book of minutes) from \mhills{} to produce a dataset of 24 documents from 140 images, each with a minimum of 5 pages: 11 documents have 5 pages, 9 have 6, 1 has 7 and 3 have 8 (average: 5.83 pages/doc).

In addition to the methods from the main text, we evaluate two open-source baselines on this dataset:
\begin{itemize}
    \item \trocr{} \citep{li2023trocr}: specifically \texttt{trocr-large-handwritten}, a pre-trained Transformer model fine-tuned on handwritten text. Since our setting is zero-shot, we do not additionally fine-tune \trocr{} on our datasets. As \trocr{} is intended for line images, we use the kraken command line tool to generate line sub-images of \mhills{} pages and concatenate results. We run \trocr{} on an M1 MacBook Pro, CPU only.
    \item \docowl{} \citep{hu2025mplug}: an open MLLM specialized for document processing. We use a temperature of zero and a sufficiently high output token limit to ensure no truncation events. We tested two versions: the original (full page images) and \textsc{docowl2-lines} (cropped line images, identical to \trocr{}). We run \docowl{} on an A10 GPU.\footnote{\docowl{} cannot be run on a CPU-only machine as the \texttt{flash-attn} python dependency requires CUDA.}
\end{itemize}

See results in Table~\ref{tab:mhills-5p-rebuttal}.

\begin{table}[h]
\centering
\caption{Results on \mhillsp{} for all methods described in the main text, plus \randompage{}, \trocr{}, \docowl{} and \textsc{docowl2-lines}. Our methods are bold. Score emphasis: \textbf{1st}, \textit{2nd}, \underline{3rd}. All failed calls (\textsc{gemma} only) were caused by API disconnection without a response.}
\label{tab:mhills-5p-rebuttal}
\small
\begin{tabular}{@{}lcrrrrrrrcc@{}}
\toprule
\multirow{2}{*}{\textbf{Method}} &
  \multicolumn{1}{l}{\multirow{2}{*}{\textbf{CER $\downarrow$}}} &
  \multicolumn{4}{c}{\textbf{Cost (\$/1k docs)}} &
  \multicolumn{3}{c}{\textbf{Tokens (k/doc)}} &
  \multicolumn{1}{c}{\multirow{2}{*}{\textbf{\begin{tabular}[c]{@{}c@{}}Time\\ (s/doc)\end{tabular}}}} &
  \multicolumn{1}{c}{\multirow{2}{*}{\textbf{\begin{tabular}[c]{@{}c@{}}Fails\\ (/doc)\end{tabular}}}} \\
 &
  \multicolumn{1}{l}{} &
  \multicolumn{1}{c}{\textbf{Total}} &
  \multicolumn{1}{c}{\textbf{OCR}} &
  \multicolumn{1}{c}{\textbf{In}} &
  \multicolumn{1}{c}{\textbf{Out}} &
  \multicolumn{1}{c}{\textbf{Image}} &
  \multicolumn{1}{c}{\textbf{In}} &
  \multicolumn{1}{c}{\textbf{Out}} &
  \multicolumn{1}{c}{} &
  \multicolumn{1}{c}{} \\ \midrule
\begin{tabular}[c]{@{}l@{}}\textsc{docowl2}\\ \textsc{-lines}\end{tabular} &
  93.01 &
  0.00 &
  0.00 &
  0.00 &
  0.00 &
  0 &
  0 &
  0 &
  2054 &
  0.00 \\
\textsc{docowl2} &
  92.08 &
  0.00 &
  0.00 &
  0.00 &
  0.00 &
  0 &
  0 &
  0 &
  234 &
  0.00 \\
\textsc{trocr} &
  31.43 &
  0.00 &
  0.00 &
  0.00 &
  0.00 &
  0 &
  0 &
  0 &
  990 &
  0.00 \\ \cmidrule(l){1-11}
\textsc{ocr} &
  13.96 &
  5.83 &
  5.83 &
  0.00 &
  0.00 &
  0 &
  0 &
  0 &
  --- &
  0.00 \\ \midrule
\multicolumn{11}{l}{\textsc{gemma}} \\ \midrule
\textsc{ocr} &
  12.92 &
  7.10 &
  5.83 &
  0.24 &
  1.03 &
  0 &
  3.36 &
  2.06 &
  44.0 &
  0.00 \\
\textsc{images} &
  16.00 &
  1.43 &
  0.00 &
  0.41 &
  1.01 &
  4.52 &
  1.41 &
  2.03 &
  85.4 &
  0.58 \\
\textsc{images-aao} &
  17.64 &
  1.32 &
  0.00 &
  0.33 &
  0.99 &
  4.52 &
  0.26 &
  1.98 &
  92.4 &
  0.75 \\
\textsc{ocr+images} &
  9.84 &
  7.40 &
  5.83 &
  0.57 &
  0.99 &
  4.52 &
  3.68 &
  1.99 &
  44.8 &
  0.00 \\
\textbf{\textsc{ocr+pageR}} &
  23.65 &
  7.06 &
  5.83 &
  0.24 &
  0.98 &
  0.77 &
  2.70 &
  1.97 &
  79.4 &
  0.58 \\
\textbf{\textsc{ocr+pageN}} &
  24.94 &
  7.22 &
  5.83 &
  0.44 &
  0.95 &
  0.77 &
  2.70 &
  1.81 &
  126 &
  1.12 \\
\textbf{\textsc{ocr+page1}} &
  21.35 &
  7.03 &
  5.83 &
  0.24 &
  0.95 &
  0.77 &
  2.68 &
  1.90 &
  89.8 &
  0.75 \\ \midrule
\multicolumn{11}{l}{\textsc{gpt-4o}} \\ \midrule
\textsc{ocr} &
  9.42 &
  33.70 &
  5.83 &
  8.35 &
  19.52 &
  0.07 &
  3.27 &
  1.95 &
  47.1 &
  0.00 \\
\textsc{images} &
  10.24 &
  33.84 &
  0.00 &
  14.77 &
  19.06 &
  4.46 &
  1.38 &
  1.91 &
  90.8 &
  0.00 \\
\textsc{images-aao} &
  12.38 &
  31.02 &
  0.00 &
  12.25 &
  18.77 &
  4.46 &
  0.25 &
  1.88 &
  80.2 &
  0.00 \\
\textsc{ocr+images} &
  6.49 &
  44.21 &
  5.83 &
  20.25 &
  18.13 &
  4.53 &
  3.57 &
  1.81 &
  83.1 &
  0.00 \\
\textbf{\textsc{ocr+pageR}} &
  10.17 &
  34.15 &
  5.83 &
  8.61 &
  19.71 &
  0.77 &
  2.53 &
  1.97 &
  46.6 &
  0.00 \\
\textbf{\textsc{ocr+pageN}} &
  7.59 &
  33.90 &
  5.83 &
  8.81 &
  19.26 &
  0.77 &
  2.53 &
  1.92 &
  68.2 &
  0.00 \\
\textbf{\textsc{ocr+page1}} &
  9.12 &
  33.32 &
  5.83 &
  8.55 &
  18.94 &
  0.77 &
  2.51 &
  1.89 &
  51.8 &
  0.00 \\ \midrule
\multicolumn{11}{l}{\textsc{gemini}} \\ \midrule
\textsc{ocr} &
  7.05 &
  29.46 &
  5.83 &
  4.20 &
  19.43 &
  0 &
  3.36 &
  1.94 &
  24.0 &
  0.00 \\
\textsc{images} &
  \textit{5.63} &
  27.37 &
  0.00 &
  7.41 &
  19.96 &
  4.52 &
  1.41 &
  2.00 &
  30.5 &
  0.00 \\
\textsc{images-aao} &
  6.14 &
  27.12 &
  0.00 &
  5.97 &
  21.15 &
  4.52 &
  0.26 &
  2.12 &
  19.2 &
  0.00 \\
\textsc{ocr+images} &
  \underline{5.69} &
  35.75 &
  5.83 &
  10.24 &
  19.67 &
  4.52 &
  3.68 &
  1.97 &
  27.5 &
  0.00 \\
\textbf{\textsc{ocr+pageR}} &
  5.99 &
  31.22 &
  5.83 &
  4.35 &
  21.04 &
  0.77 &
  2.70 &
  2.10 &
  20.9 &
  0.00 \\
\textbf{\textsc{ocr+pageN}} &
  5.86 &
  31.43 &
  5.83 &
  4.55 &
  21.05 &
  0.77 &
  2.70 &
  2.10 &
  31.2 &
  0.00 \\
\textbf{\textsc{ocr+page1}} &
  \textbf{5.43} &
  31.15 &
  5.83 &
  4.32 &
  21.00 &
  0.77 &
  2.68 &
  2.10 &
  21.4 &
  0.00 \\ \bottomrule
\end{tabular}
\end{table}

\paragraph{Discussion.}
As with \iamp{}, even with longer page counts, a \textsc{pageX} method remains the best-performing, despite having access to $<$20\% of the raw images per document, along with OCR which has a high error rate on its own. \textsc{images} alone performs quite well, and is slightly cheaper than \firstpage{} due to the high text density of the \mhills{} dataset, but it is significantly slower as each image in a multi-page document requires a separate API call.

Unfortunately, our \docowl{} and \trocr{} non-commercial-MLLM baselines perform poorly on this task zero-shot, and are quite slow. \trocr{}, at least, yields parsable text, but is prone to OCR-like errors such as the misreading of individual words or characters. Conversely, \docowl{} is occasionally accurate, but is extremely prone to a number of well-known LLM issues that destroy its overall performance: (i) early stopping (or simply outputting a single token), (ii) egregious hallucination often bearing no relation to the actual text, (iii) repetition of words or sentences ad infinitum, (iv) needless addition of code fences or explanatory text (e.g. double quotes, or `the document reads...'), (v) repeating the prompt, and (vi) ignoring the prompt completely. These behaviors persisted regardless of explicit instructions given against them during prompt iteration.

\subsection{\mhillspp{}}
\label{sec:app:mhills-10p}
We use the same generation method as \mhillsp{}. We obtain 10 documents in total from 115 images; 2 have 10 pages, 2 have 11, 5 have 12, 1 has 13 (average: 11.5 pages/doc).

See results in Table~\ref{tab:mhills-10p-rebuttal}.

\begin{table}[h]
\centering
\caption{Results on \mhillspp{} for all methods described in the main text, plus \randompage{}. Our methods are bold. Score emphasis: \textbf{1st}, \textit{2nd}, \underline{3rd}. All failed calls (\textsc{gemma} only) were a result of incomplete responses (i.e. failure or truncation).}
\label{tab:mhills-10p-rebuttal}
\small
\begin{tabular}{@{}lcrrrrrrrcc@{}}
\toprule
\multirow{2}{*}{\textbf{Method}} &
  \multicolumn{1}{l}{\multirow{2}{*}{\textbf{CER $\downarrow$}}} &
  \multicolumn{4}{c}{\textbf{Cost (\$/1k docs)}} &
  \multicolumn{3}{c}{\textbf{Tokens (k/doc)}} &
  \multicolumn{1}{c}{\multirow{2}{*}{\textbf{\begin{tabular}[c]{@{}c@{}}Time\\ (s/doc)\end{tabular}}}} &
  \multicolumn{1}{c}{\multirow{2}{*}{\textbf{\begin{tabular}[c]{@{}c@{}}Fails\\ (/doc)\end{tabular}}}} \\
 &
  \multicolumn{1}{l}{} &
  \multicolumn{1}{c}{\textbf{Total}} &
  \multicolumn{1}{c}{\textbf{OCR}} &
  \multicolumn{1}{c}{\textbf{In}} &
  \multicolumn{1}{c}{\textbf{Out}} &
  \multicolumn{1}{c}{\textbf{Image}} &
  \multicolumn{1}{c}{\textbf{In}} &
  \multicolumn{1}{c}{\textbf{Out}} &
  \multicolumn{1}{c}{} &
  \multicolumn{1}{c}{} \\ \midrule
\textsc{ocr} &
  12.92 &
  11.50 &
  11.50 &
  0.00 &
  0.00 &
  0 &
  0 &
  0 &
  0.0 &
  0.00 \\ \midrule
\multicolumn{11}{l}{\textsc{gpt-4o}} \\ \midrule
\textsc{ocr} &
  8.48 &
  67.28 &
  11.50 &
  16.65 &
  39.13 &
  0.13 &
  6.53 &
  3.91 &
  93.2 &
  0.00 \\
\textsc{images} &
  9.90 &
  67.13 &
  0.00 &
  29.12 &
  38.01 &
  8.80 &
  2.71 &
  3.80 &
  181.1 &
  0.00 \\
\textsc{images-aao} &
  11.64 &
  59.37 &
  0.00 &
  23.45 &
  35.92 &
  8.80 &
  0.25 &
  3.59 &
  100.8 &
  0.20 \\
\textsc{ocr+images} &
  6.44 &
  87.86 &
  11.50 &
  40.11 &
  36.26 &
  8.93 &
  7.12 &
  3.63 &
  165.4 &
  0.00 \\
\textbf{\textsc{ocr+pageR}} &
  17.11 &
  56.93 &
  11.50 &
  12.22 &
  33.21 &
  0.77 &
  4.70 &
  3.32 &
  103.9 &
  0.30 \\
\textbf{\textsc{ocr+pageN}} &
  16.37 &
  57.09 &
  11.50 &
  12.58 &
  33.01 &
  0.77 &
  4.70 &
  3.30 &
  123.7 &
  0.30 \\
\textbf{\textsc{ocr+page1}} &
  21.38 &
  56.48 &
  11.50 &
  12.17 &
  32.82 &
  0.77 &
  4.68 &
  3.28 &
  119.1 &
  0.40 \\ \midrule
\multicolumn{11}{l}{\textsc{gemini}} \\ \midrule
\textsc{ocr} &
  6.88 &
  59.29 &
  11.50 &
  8.40 &
  39.39 &
  0 &
  6.72 &
  3.94 &
  47.9 &
  0.00 \\
\textsc{images} &
  \underline{4.87} &
  54.71 &
  0.00 &
  14.61 &
  40.11 &
  8.90 &
  2.78 &
  4.01 &
  60.6 &
  0.00 \\
\textsc{images-aao} &
  6.05 &
  54.46 &
  0.00 &
  11.45 &
  43.00 &
  8.90 &
  0.26 &
  4.30 &
  40.9 &
  0.00 \\
\textsc{ocr+images} &
  \textit{4.80} &
  71.41 &
  11.50 &
  20.30 &
  39.61 &
  8.90 &
  7.34 &
  3.96 &
  54.9 &
  0.00 \\
\textbf{\textsc{ocr+pageR}} &
  5.50 &
  61.82 &
  11.50 &
  7.26 &
  43.06 &
  0.77 &
  5.03 &
  4.31 &
  40.1 &
  0.00 \\
\textbf{\textsc{ocr+pageN}} &
  5.41 &
  61.77 &
  11.50 &
  7.62 &
  42.65 &
  0.77 &
  5.03 &
  4.26 &
  63.8 &
  0.00 \\
\textbf{\textsc{ocr+page1}} &
  \textbf{4.76} &
  61.13 &
  11.50 &
  7.23 &
  42.41 &
  0.77 &
  5.01 &
  4.24 &
  40.6 &
  0.00 \\ \bottomrule
\end{tabular}
\end{table}

\paragraph{Discussion.}
Even for documents of at least 10 pages in length, \firstpage{} remains the top-performing method, with \textsc{ocr+images} being the second best, about equivalent in performance, but much slower and more expensive. This suggests that using a single page to improve performance does scale reasonably well, even for quite long documents.
This is good news, as it means that, for documents where the average number of text tokens inside each image is less than the average number of tokens produced by tokenizing the image (the case for \casia{}, \bentham{}, \iam{} and many other real-world datasets — \mhills{} is particularly token-dense), then \textsc{+pageX} methods will only be more cost and time effective in comparison to full-image methods as document length increases.

\newpage
\section{\mhills{} dataset metadata and ablations}
\label{sec:app:mhills-metadata}
This section includes information about \mhills{} at the image level: incidence of challenging OCR features (Table~\ref{tab:mhills-dataset-stats}) and information about writers (Table~\ref{tab:mhills-author-ids}), time-of-writing (Table~\ref{tab:mhills-years}), and breakdown of pages by primary and secondary content types (Table~\ref{tab:mhills-page-types}) for each page.

\begin{table}[h]
\centering
\caption{\mhills{} image statistics and prevalence of various OCR challenges. Particularly notable are the reasonably high frequencies of distractor text, tabular data, archaic language and multiple authors.}
\label{tab:mhills-dataset-stats}
\begin{tabular}{@{}lr@{}}
\toprule
Word count                        & 219 ± 83   \\
Character count                   & 1260 ± 473 \\
Includes tabular information      & 19.3\%     \\
Includes margin notes             & 15.7\%     \\
Includes distractor text          & 62.1\%     \\
Non-linear structure              & 0.7\%      \\
Archaic language                  & 31.4\%     \\
Poor quality image/damaged paper  & 2.1\%      \\
Handwriting from multiple authors & 18.6\%     \\
Includes crossed-out text         & 23.6\%     \\ \bottomrule
\end{tabular}
\end{table}

\begin{table}[h]
\centering
\begin{minipage}[t]{0.38\linewidth}
\centering
\caption{Unique writers for the \mhills{} dataset and number of documents containing that writer's handwriting for each. Note that some pages contain multiple hands.}
\label{tab:mhills-author-ids}
\begin{tabular}{@{}cc@{}}
\toprule
\textbf{Author ID} & \textbf{Count} \\ \midrule
0                  & 44             \\
1                  & 23             \\
2                  & 18             \\
3                  & 17             \\
4                  & 16             \\
5                  & 15             \\
6                  & 11             \\
7                  & 6              \\
8                  & 5              \\
9                  & 5              \\
10                 & 2              \\
11                 & 2              \\
12                 & 1              \\
13                 & 1              \\ \bottomrule
\end{tabular}
\end{minipage}
\hfill
\begin{minipage}[t]{0.55\linewidth}
\centering
\caption{Years of writing for document pages in \mhills{}. Some documents were written in one year but copied from a document originally written in another year; where known, this table includes both --- for example, some documents use archaic 17th century language but were copied by hand in the 19th century.}
\label{tab:mhills-years}
\begin{tabular}{@{}ccc@{}}
\toprule
\textbf{Year} & \textbf{Original} & \textbf{Written} \\ \midrule
1631          & 11                & 0                \\
1632          & 17                & 17               \\
1795          & 6                 & 6                \\
1899          & 15                & 15               \\
1915          & 17                & 17               \\
1925          & 22                & 22               \\
1932          & 5                 & 5                \\
1934          & 20                & 20               \\
1936          & 8                 & 8                \\
1938          & 11                & 11               \\
Unknown       & 8                 & 8                \\ \bottomrule
\end{tabular}
\end{minipage}
\end{table}

\begin{table}[h]
\centering
\caption{Breakdown of primary and secondary page types found in \mhills{}.}
\label{tab:mhills-page-types}
\begin{tabular}{@{}
>{\columncolor[HTML]{FFFFFF}}l 
>{\columncolor[HTML]{FFFFFF}}r 
>{\columncolor[HTML]{FFFFFF}}l 
>{\columncolor[HTML]{FFFFFF}}r @{}}
\toprule
\textbf{Primary Type}                                                   & \textbf{\# Pages}         & \textbf{Secondary Types}      & \textbf{\# Pages} \\ \midrule
\cellcolor[HTML]{FFFFFF}                                                & \cellcolor[HTML]{FFFFFF} & ---                           & 72               \\
\multirow{-2}{*}{\cellcolor[HTML]{FFFFFF}\begin{tabular}[c]{@{}l@{}}Historical\\ minutes\end{tabular}} & \multirow{-2}{*}{\cellcolor[HTML]{FFFFFF}98} & Tabular      & 26 \\ \midrule
\cellcolor[HTML]{FFFFFF}                                                & \cellcolor[HTML]{FFFFFF} & Statute                       & 17               \\
\cellcolor[HTML]{FFFFFF}                                                & \cellcolor[HTML]{FFFFFF} & Memoranda roll                & 10               \\
\cellcolor[HTML]{FFFFFF}                                                & \cellcolor[HTML]{FFFFFF} & Case memorandum               & 4                \\
\cellcolor[HTML]{FFFFFF}                                                & \cellcolor[HTML]{FFFFFF} & Memoranda roll, Tabular       & 1                \\
\cellcolor[HTML]{FFFFFF}                                                & \cellcolor[HTML]{FFFFFF} & Case memorandum, Legal letter & 1                \\
\multirow{-6}{*}{\cellcolor[HTML]{FFFFFF}\begin{tabular}[c]{@{}l@{}}Historical\\ legal\end{tabular}}   & \multirow{-6}{*}{\cellcolor[HTML]{FFFFFF}34} & Legal letter & 1  \\ \midrule
\begin{tabular}[c]{@{}l@{}}Historical\\ inventory/schedule\end{tabular} & 8                        & ---                           & 8                \\ \bottomrule
\end{tabular}
\end{table}

\newpage

\subsection{\mhillsp{} document type and feature ablation}
\label{sec:app:mhills-ablation}

Using primary document types from Table~\ref{tab:mhills-doc-types-rebuttal} and some choice dataset statistics from Table~\ref{tab:mhills-dataset-stats}, we ablate document types and features for \mhillsp{} in Tables~\ref{tab:mhills-minutes-rebuttal}--\ref{tab:mhills-inventory-rebuttal} (by document type) and Tables~\ref{tab:mhills-tabular-rebuttal}--\ref{tab:mhills-multiple-hands-rebuttal} (by document feature).

Each document of 5 or more pages has a single shared primary type over all individual page images. A document is \texttt{tabular}, \texttt{archaic} or has \texttt{multiiple\_hands} if any of its constituent pages has this property.

\begin{table}[h]
\centering
\caption{For the individual 5+ page \emph{documents} of \texttt{Malvern-Hills-5+} (not individual pages/images), the prevalence of tabular data, archaic language and multiple writers for each document. A document is treated as \texttt{tabular}/\texttt{archaic}/\texttt{multiple\_hands} if any pages have this feature. In general, only one or two pages per document will have tabular information or multiple hands, whereas archaic language will typically be throughout.}
\label{tab:mhills-doc-types-rebuttal}
\begin{tabular}{@{}lrrrrrl@{}}
\toprule
\begin{tabular}[c]{@{}l@{}}\textbf{Document}\\\textbf{type}\end{tabular} &
\begin{tabular}[c]{@{}r@{}}\textbf{Num.}\\\textbf{docs}\end{tabular} &
\begin{tabular}[c]{@{}r@{}}\textbf{Num.}\\\textbf{pages}\end{tabular} &
\begin{tabular}[c]{@{}r@{}}\textbf{\texttt{tabular}}\\\textbf{(\%)}\end{tabular} &
\begin{tabular}[c]{@{}r@{}}\textbf{\texttt{archaic}}\\\textbf{(\%)}\end{tabular} &
\begin{tabular}[c]{@{}r@{}}\textbf{\texttt{multiple}}\\\textbf{\texttt{\_hands} (\%)}\end{tabular} &
\begin{tabular}[c]{@{}l@{}}\textbf{\texttt{written\_year}}\\\textbf{min $\mid$ median $\mid$ max}\end{tabular} \\ \midrule
\begin{tabular}[c]{@{}l@{}}\texttt{historical}\\\texttt{\_minutes}\end{tabular} &
17 & 98 & 88.2 & 41.2 & 82.4 & 1899 $\mid$ 1925 $\mid$ 1938 \\
\begin{tabular}[c]{@{}l@{}}\texttt{historical}\\\texttt{\_legal}\end{tabular} &
6  & 34 & 16.7 & 100  & 33.3 & 1632 $\mid$ 1713 $\mid$ 1884 \\ \bottomrule
\end{tabular}
\end{table}

\begin{table}[h]
\centering
\caption{Performance for \texttt{historical\_minutes}-typed documents only from \texttt{Malvern-Hills-5+}. All methods use \textsc{gemini-2.5-pro} as the MLLM.}
\label{tab:mhills-minutes-rebuttal}
\begin{tabular}{@{}lr@{}}
\toprule
 \textbf{Method} & \textbf{CER (\%)} \\ \midrule
Azure OCR only & 11.8  \\
\textsc{ocr}          & 5.3   \\
\textsc{images-aao}   & 4.55  \\
\textsc{ocr+pageR}    & 3.88  \\
\textsc{ocr+pageN}    & 3.83  \\
\textsc{ocr+page1}    & 3.8   \\
\textsc{images}       & 3.16  \\
\textsc{ocr+images}   & 3.11  \\ \bottomrule
\end{tabular}
\end{table}

\begin{table}[h]
\centering
\caption{Performance for \texttt{historical\_legal}-typed documents only from \texttt{Malvern-Hills-5+}. All methods use \textsc{gemini-2.5-pro} as the MLLM.}
\label{tab:mhills-legal-rebuttal}
\begin{tabular}{@{}lr@{}}
\toprule
\textbf{Method} & \textbf{CER (\%)} \\ \midrule
Azure OCR only & 20.81 \\
\textsc{ocr+images}   & 13.56 \\
\textsc{images}       & 13.17 \\
\textsc{ocr}          & 12.53 \\
\textsc{ocr+pageR}    & 12.36 \\
\textsc{ocr+pageN}    & 12.03 \\
\textsc{images-aao}   & 11.11 \\
\textsc{ocr+page1}    & 10.36 \\ \bottomrule
\end{tabular}
\end{table}

\begin{table}[h]
\centering
\caption{Performance for \texttt{historical\_inventory/schedule}-typed documents only from \texttt{Malvern-Hills-5+}. All methods use \textsc{gemini-2.5-pro} as the MLLM.}
\label{tab:mhills-inventory-rebuttal}
\begin{tabular}{@{}lr@{}}
\toprule
\textbf{Method}      & \textbf{CER (\%)} \\ \midrule
Azure OCR only       & 9.58 \\
\textsc{ocr}         & 4.06 \\
\textsc{ocr+pageR}   & 3.56 \\
\textsc{ocr+page1}   & 3.46 \\
\textsc{ocr+pageN}   & 3.41 \\
\textsc{images-aao}  & 3.27 \\
\textsc{images}      & 2.42 \\
\textsc{ocr+images}  & 2.34 \\ \bottomrule
\end{tabular}
\end{table}

\begin{table}[h]
\centering
\caption{Performance for documents from \texttt{Malvern-Hills-5+} split by presence of tabular content. A document is \texttt{tabular} if any constituent page includes tabular layout. All methods use \textsc{gemini-2.5-pro} as the MLLM.}
\label{tab:mhills-tabular-rebuttal}
\begin{subtable}[t]{0.48\textwidth}
\centering
\caption{\texttt{tabular} documents}
\begin{tabular}{@{}lr@{}}
\toprule
\textbf{Method}      & \textbf{CER (\%)} \\ \midrule
Azure OCR only       & 12.18 \\
\textsc{ocr}         & 5.56  \\
\textsc{images-aao}  & 4.56  \\
\textsc{ocr+pageR}   & 4.00  \\
\textsc{ocr+page1}   & 3.98  \\
\textsc{ocr+pageN}   & 3.97  \\
\textsc{images}      & 3.28  \\
\textsc{ocr+images}  & 3.20  \\ \bottomrule
\end{tabular}
\end{subtable}\hfill
\begin{subtable}[t]{0.48\textwidth}
\centering
\caption{\texttt{non\_tabular} documents}
\begin{tabular}{@{}lr@{}}
\toprule
\textbf{Method}      & \textbf{CER (\%)} \\ \midrule
Azure OCR only       & 17.53 \\
\textsc{ocr+images}  & 10.68 \\
\textsc{images}      & 10.35 \\
\textsc{ocr}         & 10.05 \\
\textsc{ocr+pageR}   & 9.96  \\
\textsc{ocr+pageN}   & 9.63  \\
\textsc{images-aao}  & 9.28  \\
\textsc{ocr+page1}   & 8.31  \\ \bottomrule
\end{tabular}
\end{subtable}
\end{table}

\begin{table}[h]
\centering
\caption{Performance for documents from \texttt{Malvern-Hills-5+} split by presence of archaic language. A document is \texttt{archaic} if any constituent page is marked as such. All methods use \textsc{gemini-2.5-pro} as the MLLM.}
\label{tab:mhills-archaic-rebuttal}
\begin{subtable}[t]{0.48\textwidth}
\centering
\caption{\texttt{archaic} documents}
\begin{tabular}{@{}lr@{}}
\toprule
\textbf{Method}      & \textbf{CER (\%)} \\ \midrule
Azure OCR only       & 16.50 \\
\textsc{ocr}         & 8.61  \\
\textsc{ocr+images}  & 7.96  \\
\textsc{images}      & 7.86  \\
\textsc{ocr+pageR}   & 7.74  \\
\textsc{images-aao}  & 7.71  \\
\textsc{ocr+pageN}   & 7.58  \\
\textsc{ocr+page1}   & 6.95  \\ \bottomrule
\end{tabular}
\end{subtable}\hfill
\begin{subtable}[t]{0.48\textwidth}
\centering
\caption{\texttt{non\_archaic} documents}
\begin{tabular}{@{}lr@{}}
\toprule
\textbf{Method}      & \textbf{CER (\%)} \\ \midrule
Azure OCR only       & 10.96 \\
\textsc{ocr}         & 5.22  \\
\textsc{images-aao}  & 4.28  \\
\textsc{ocr+pageR}   & 3.92  \\
\textsc{ocr+pageN}   & 3.82  \\
\textsc{ocr+page1}   & 3.63  \\
\textsc{ocr+images}  & 3.01  \\
\textsc{images}      & 3.00  \\ \bottomrule
\end{tabular}
\end{subtable}
\end{table}

\begin{table}[h]
\centering
\caption{Performance for documents from \texttt{Malvern-Hills-5+} split by whether they contain handwriting from multiple authors. A document has \texttt{multiple\_hands} if any page is written by a different author. All methods use \textsc{gemini-2.5-pro} as the MLLM.}
\label{tab:mhills-multiple-hands-rebuttal}
\begin{subtable}[t]{0.48\textwidth}
\centering
\caption{\texttt{multiple\_hands} documents}
\begin{tabular}{@{}lr@{}}
\toprule
\textbf{Method}      & \textbf{CER (\%)} \\ \midrule
Azure OCR only       & 12.35 \\
\textsc{ocr}         & 5.73  \\
\textsc{images-aao}  & 5.12  \\
\textsc{ocr+pageR}   & 4.79  \\
\textsc{ocr+pageN}   & 4.76  \\
\textsc{ocr+page1}   & 4.15  \\
\textsc{images}      & 3.85  \\
\textsc{ocr+images}  & 3.55  \\ \bottomrule
\end{tabular}
\end{subtable}\hfill
\begin{subtable}[t]{0.48\textwidth}
\centering
\caption{\texttt{single\_hand} documents}
\begin{tabular}{@{}lr@{}}
\toprule
\textbf{Method}      & \textbf{CER (\%)} \\ \midrule
Azure OCR only       & 17.19 \\
\textsc{ocr+images}  & 9.97  \\
\textsc{ocr}         & 9.70  \\
\textsc{images}      & 9.20  \\
\textsc{ocr+pageR}   & 8.38  \\
\textsc{images-aao}  & 8.17  \\
\textsc{ocr+pageN}   & 8.07  \\
\textsc{ocr+page1}   & 7.98  \\ \bottomrule
\end{tabular}
\end{subtable}
\end{table}

\paragraph{Discussion.}
Overall we see that the \texttt{historical\_legal} documents are much more challenging, with $>3\times$ the proportion of errors compared to \texttt{historical\_minutes}. We can likely attribute this to (from examination) the ubiquity of archaic language and florid cursive (a product of the much earlier writing dates, $\sim 200$ years prior), which is much harder to transcribe and much more prone to errors in post-processing/correction. While \texttt{historical\_minutes} documents do include some archaic language, these are generally sporadic instances, rather than comprising the overall style of writing.

Comparing the two document types, we can see that \textsc{ocr+single-page} methods, and especially \textsc{ocr+page1}, are dominant on the \emph{more challenging archaic document type}, while full-image methods \textsc{images}, \textsc{ocr+images} struggle. The reverse is true for the overall \emph{easier} \texttt{historical\_minutes} document type, where \textsc{images}, \textsc{ocr+images} dominate (though \textsc{ocr+pageX} methods are still competitive, especially given their comparative image token dearth).

Considering the dominant features of each type of document; it is unsurprising that \textsc{ocr+page1} is less adept for transcribing documents where a single page includes tabular data, as it is likely that there will not be any on the ``seen'' page. Conversely, archaic language is likely to be consistent across pages, so single-page extrapolation is more effective in this case.

Overall though, what these results suggest is that \emph{\textsc{ocr+single-page} methods are most beneficial when the task is more challenging}. We can interpret these results in terms of the tradeoff between prompt complexity and task difficulty. If a task is relatively easier (\texttt{\_minutes}), the MLLM is less likely to become overwhelmed by a high-complexity or long prompt; it can leverage the additional detail (e.g. all images and OCR) to achieve incremental performance improvement on an almost-solved task. If a task is more challenging, this additional detail/complexity hurts overall performance, and a more optimal balance is achieved with a multi-modal prompt that reduces redundancy — i.e. OCR with a single page.

\paragraph{Relative importance of document type vs. document features} We can see that document type dominates the differences in performance, --- i.e. tables for \texttt{tabular}, \texttt{multiple\_writer}, \texttt{non\_archaic} generally follow \texttt{historical\_minutes} (and vice versa for \texttt{historical\_legal}).

\newpage
\section{Additional resources}

Below are links to several online tools mentioned in this paper:
\begin{itemize}
    \item Tesseract: \url{https://github.com/tesseract-ocr/tesseract}
    \item LLM-Aided OCR: \url{https://github.com/Dicklesworthstone/llm_aided_ocr}
    \item BetterOCR: \url{https://github.com/junhoyeo/BetterOCR}
\end{itemize}

\paragraph{Figures.} We acknowledge the use of (both original and edited) open-licensed SVG vectors from SVG Repo:
\begin{itemize}
\item Images clipart by Ionicons: \url{https://www.svgrepo.com/svg/327088/images-sharp}
\item Robot clipart by Konstantin Filatov \url{https://www.svgrepo.com/svg/521818/robot}
\item Documents clipart by SVG repo: \url{https://www.svgrepo.com/svg/139884/documents-papers}
\end{itemize}

\paragraph{\mhills{} dataset.} We are grateful to the Malvern Hills Trust for providing images.

\subsection{Commercial tools}
\label{sec:app-costs}
Costs of commercial OCR engines and LLMs for estimates used in this paper are given in Tables~\ref{table:cost_ocr_engines},~\ref{table:cost_llms}.
Most have pricing tiers based on scale, or a limited free allowance; for simplicity we take the lowest (non-batch) cost per run for each engine, and for the OpenAI API.

Costs for each tool were taken from their respective webpages:
\begin{itemize}
    \item Microsoft Azure: \url{https://azure.microsoft.com/en-gb/pricing/details/cognitive-services/computer-vision/}
    \item Amazon Textract: \url{https://aws.amazon.com/textract/pricing/}
    \item Google Cloud Vision: \url{https://cloud.google.com/vision/pricing}
    \item OpenAI: \url{https://openai.com/api/pricing/}
    \item Google Gemini: \url{https://ai.google.dev/gemini-api/docs/pricing}
    \item Google does not provide pricing information for \textsc{gemma-3-27b}, and only provide it for free with usage limits. For this reason we take our estimate of the price of \textsc{gemma-3-27b} from \url{https://openrouter.ai/google/gemma-3-27b-it}
\end{itemize}

\begin{table}[H]
\caption{Pricing for OCR Engines}
\centering
\begin{tabular}{@{}lc@{}}
\toprule
\textbf{OCR Engine}       & \textbf{Cost per 1k Calls (\$)} \\ \midrule
Azure Vision           & 1.00                             \\
Google Cloud Vision       & 1.50                             \\
Amazon Textract           & 1.50                             \\ \bottomrule
\end{tabular}
\label{table:cost_ocr_engines}
\end{table}

\begin{table}[H]
\caption{Pricing for LLMs}
\centering
\begin{tabular}{@{}lcc@{}}
\toprule
\textbf{LLM}              & \multicolumn{2}{c}{\textbf{Cost per 1M Tokens (\$)}}            \\ \cmidrule(lr){2-3}
                          & \textbf{Input} & \textbf{Output} \\ \midrule
\textsc{gemma-3-27b}                    & 0.07                        & 0.50                         \\
\textsc{gpt-4o}                    & 2.50                        & 10.00                         \\
\textsc{gemini-2.5-pro}               & 1.25                        & 10.00                          \\ 

\bottomrule
\end{tabular}
\label{table:cost_llms}
\end{table}

\newpage
\section{Additional experiments and extended tables}
\label{sec:app-additional-results}

\subsection{Extended tables with token and cost breakdowns}
\label{sec:app:extended-tables}
Tables~\ref{tab:iam-rebuttal}, \ref{tab:mhills-rebuttal} and \ref{tab:bentham-rebuttal} provide per-stage token and cost breakdowns for the \iam{}, \mhills{} and \bentham{} experiments reported in the main text, and correspond to Tables~\ref{tab:iam}, \ref{tab:mhills} \& \ref{tab:bentham} respectively.

\subsubsection{\iam{} with \pylaia{}}
\label{sec:app:non-commercial-baselines}
Table~\ref{tab:iam-rebuttal} also includes a row for \pylaia{} \citep{tarride2024improving}, an open-source baseline. As our problem setting is zero-shot and document-level, \pylaia{} poses a problem as it is (i) largely dependent on fine-tuning to achieve reasonable performance on out-of-sample data, and (ii) operates on the line level, rather than the page or document level. We run \pylaia{} on an M1 MacBook Pro, CPU only.

We use \pylaia{} out-of-the-box with the public \texttt{Teklia/pylaia-iam} model, a CNN–BLSTM–CTC recognizer trained on IAM. We enable \pylaia{}'s lexicon-aware decoding and use the bundled files from the model repository (\texttt{tokens.txt,  lexicon.txt, language\_model.arpa.gz}). The language model is a 6-gram character LM trained on IAM.

\useunder{\uline}{\ul}{}
\begin{table}[h]
\centering
\caption{Results on \iam{}. Much of this table is reproduced from Table~\ref{tab:iam}; but new columns with token and cost breakdowns have been added, as well as a new row for \pylaia{}. See Tables~\ref{tab:mhills-rebuttal}~\&~\ref{tab:bentham-rebuttal} for similar reproductions of Tables~\ref{tab:mhills}~\&~\ref{tab:bentham} respectively.}
\label{tab:iam-rebuttal}
\small
\begin{tabular}{@{}rlcrrrrrrr@{}}
\toprule
\multirow{2}{*}{\textbf{}} &
  \multirow{2}{*}{\textbf{Method}} &
  \multicolumn{1}{l}{\multirow{2}{*}{\textbf{CER $\downarrow$}}} &
  \multicolumn{4}{c}{\textbf{Cost (\$/1k docs)}} &
  \multicolumn{3}{c}{\textbf{Tokens (/doc)}} \\
 &
   &
  \multicolumn{1}{l}{} &
  \multicolumn{1}{c}{\textbf{Total}} &
  \multicolumn{1}{c}{\textbf{OCR}} &
  \multicolumn{1}{c}{\textbf{In}} &
  \multicolumn{1}{c}{\textbf{Out}} &
  \multicolumn{1}{c}{\textbf{Image}} &
  \multicolumn{1}{c}{\textbf{In}} &
  \multicolumn{1}{c}{\textbf{Out}} \\ \midrule
\textsc{---} &
  \textsc{pylaia} &
  6.51 &
  0.00 &
  0.00 &
  0.00 &
  0.00 &
  0 &
  0 &
  0 \\ \midrule
\textsc{---} &
  \textsc{ocr} &
  3.81 &
  2.26 &
  2.26 &
  0.00 &
  0.00 &
  0 &
  0 &
  0 \\ \midrule
\multirow{6}{*}{\raisebox{0.0em}[0pt][0pt]{\rotatebox[origin=c]{90}{\textsc{gemma}}}} &
  \textsc{ocr} &
  2.93 &
  2.42 &
  2.26 &
  0.05 &
  0.11 &
  0 &
  699 &
  211 \\
 &
  \textsc{images} &
  2.31 &
  0.31 &
  0.00 &
  0.19 &
  0.12 &
  2,192 &
  547 &
  230 \\
 &
  \textsc{images-all-at-once} &
  1.97 &
  0.29 &
  0.00 &
  0.17 &
  0.12 &
  2,192 &
  262 &
  242 \\
 &
  \textsc{ocr+images} &
  1.78 &
  2.58 &
  2.26 &
  0.21 &
  0.11 &
  2,192 &
  821 &
  216 \\
 &
  \textsc{ocr+pageN} &
  2.17 &
  2.59 &
  2.26 &
  0.17 &
  0.16 &
  945 &
  648 &
  246 \\
 &
  \textsc{ocr+page1} &
  1.36 &
  2.50 &
  2.26 &
  0.11 &
  0.12 &
  980 &
  623 &
  243 \\ \midrule
\multirow{6}{*}{\raisebox{0.0em}[0pt][0pt]{\rotatebox[origin=c]{90}{\textsc{gpt-4o}}}} &
  \textsc{ocr} &
  1.21 &
  6.17 &
  2.26 &
  1.77 &
  2.14 &
  26 &
  682 &
  213 \\
 &
  \textsc{images} &
  0.92 &
  9.81 &
  0.00 &
  7.63 &
  2.17 &
  2,515 &
  533 &
  217 \\
 &
  \textsc{images-all-at-once} &
  0.92 &
  9.29 &
  0.00 &
  7.11 &
  2.18 &
  2,515 &
  252 &
  218 \\
 &
  \textsc{ocr+images} &
  {\ul 0.72} &
  12.69 &
  2.26 &
  8.29 &
  2.13 &
  2,519 &
  797 &
  213 \\
 &
  \textsc{ocr+pageN} &
  0.87 &
  9.04 &
  2.26 &
  4.52 &
  2.26 &
  1,095 &
  613 &
  222 \\
 &
  \textsc{ocr+page1} &
  0.85 &
  8.94 &
  2.26 &
  4.47 &
  2.21 &
  1,124 &
  589 &
  221 \\ \midrule
\multirow{6}{*}{\raisebox{0.0em}[0pt][0pt]{\rotatebox[origin=c]{90}{\textsc{gemini}}}} &
  \textsc{ocr} &
  1.57 &
  5.21 &
  2.26 &
  0.87 &
  2.08 &
  0 &
  699 &
  207 \\
 &
  \textsc{images} &
  1.16 &
  5.56 &
  0.00 &
  3.43 &
  2.13 &
  2,192 &
  547 &
  213 \\
 &
  \textsc{images-all-at-once} &
  0.70 &
  5.50 &
  0.00 &
  3.07 &
  2.44 &
  2,192 &
  262 &
  243 \\
 &
  \textsc{ocr+images} &
  \textbf{0.64} &
  8.10 &
  2.26 &
  3.77 &
  2.08 &
  2,192 &
  821 &
  207 \\
 &
  \textsc{ocr+pageN} &
  \textbf{0.63} &
  6.71 &
  2.26 &
  2.05 &
  2.40 &
  945 &
  648 &
  236 \\
 &
  \textsc{ocr+page1} &
  \textit{0.65} &
  6.63 &
  2.26 &
  2.00 &
  2.37 &
  980 &
  623 &
  236 \\ \bottomrule
\end{tabular}
\end{table}

\subsubsection{\mhills{} and \bentham{}}
See Tables~\ref{tab:mhills-rebuttal}~\&~\ref{tab:bentham-rebuttal} for extended versions of Tables~\ref{tab:mhills}~\&~\ref{tab:bentham} respectively, with token and cost breakdowns added.

\begin{table}[h]
\centering
\caption{Results on \mhills{}. Much of this table is reproduced from Table~\ref{tab:mhills}; but new columns with token and cost breakdowns have been added.}
\label{tab:mhills-rebuttal}
\small
\begin{tabular}{@{}rlcrrrrrrr@{}}
\toprule
\multirow{2}{*}{\textbf{}} &
  \multirow{2}{*}{\textbf{Method}} &
  \multicolumn{1}{l}{\multirow{2}{*}{\textbf{CER $\downarrow$}}} &
  \multicolumn{4}{c}{\textbf{Cost (\$/1k docs)}} &
  \multicolumn{3}{c}{\textbf{Tokens (/doc)}} \\
 &
   &
  \multicolumn{1}{l}{} &
  \multicolumn{1}{c}{\textbf{Total}} &
  \multicolumn{1}{c}{\textbf{OCR}} &
  \multicolumn{1}{c}{\textbf{In}} &
  \multicolumn{1}{c}{\textbf{Out}} &
  \multicolumn{1}{c}{\textbf{Image}} &
  \multicolumn{1}{c}{\textbf{In}} &
  \multicolumn{1}{c}{\textbf{Out}} \\ \midrule
\textsc{---} &
  \textsc{ocr} &
  14.41 &
  2.30 &
  2.30 &
  0.00 &
  0.00 &
  0 &
  0 &
  0 \\ \midrule
\multirow{6}{*}{\raisebox{0.0em}[0pt][0pt]{\rotatebox[origin=c]{90}{\textsc{gemma}}}} &
  \textsc{ocr} &
  13.52 &
  2.80 &
  2.30 &
  0.09 &
  0.41 &
  0 &
  1,331 &
  818 \\
 &
  \textsc{images} &
  27.19 &
  0.74 &
  0.00 &
  0.16 &
  0.58 &
  1,791 &
  556 &
  1,152 \\
 &
  \textsc{images-all-at-once} &
  15.21 &
  0.66 &
  0.00 &
  0.14 &
  0.52 &
  1,791 &
  262 &
  1,037 \\
 &
  \textsc{ocr+images} &
  10.54 &
  2.92 &
  2.30 &
  0.23 &
  0.40 &
  1,791 &
  1,455 &
  791 \\
 &
  \textbf{\textsc{ocr+pageN}} &
  11.22 &
  3.03 &
  2.30 &
  0.25 &
  0.48 &
  781 &
  1,317 &
  879 \\
 &
  \textbf{\textsc{ocr+page1}} &
  12.55 &
  2.89 &
  2.30 &
  0.14 &
  0.44 &
  781 &
  1,285 &
  888 \\ \midrule
\multirow{6}{*}{\raisebox{0.0em}[0pt][0pt]{\rotatebox[origin=c]{90}{\textsc{gpt-4o}}}} &
  \textsc{ocr} &
  10.60 &
  13.32 &
  2.30 &
  3.31 &
  7.71 &
  25 &
  1,298 &
  771 \\
 &
  \textsc{images} &
  11.24 &
  13.46 &
  0.00 &
  5.86 &
  7.60 &
  1,774 &
  542 &
  760 \\
 &
  \textsc{images-all-at-once} &
  11.35 &
  12.97 &
  0.00 &
  5.31 &
  7.66 &
  1,774 &
  252 &
  765 \\
 &
  \textsc{ocr+images} &
  7.11 &
  17.54 &
  2.30 &
  8.04 &
  7.20 &
  1,799 &
  1,415 &
  720 \\
 &
  \textbf{\textsc{ocr+pageN}} &
  8.05 &
  15.30 &
  2.30 &
  5.35 &
  7.65 &
  774 &
  1,235 &
  761 \\
 &
  \textbf{\textsc{ocr+page1}} &
  8.92 &
  15.00 &
  2.30 &
  5.16 &
  7.53 &
  774 &
  1,203 &
  753 \\ \midrule
\multirow{6}{*}{\raisebox{0.0em}[0pt][0pt]{\rotatebox[origin=c]{90}{\textsc{gemini}}}} &
  \textsc{ocr} &
  7.47 &
  11.67 &
  2.30 &
  1.66 &
  7.70 &
  0 &
  1,331 &
  770 \\
 &
  \textsc{images} &
  \textit{6.42} &
  10.88 &
  0.00 &
  2.93 &
  7.95 &
  1,791 &
  556 &
  794 \\
 &
  \textsc{images-all-at-once} &
  8.20 &
  11.01 &
  0.00 &
  2.57 &
  8.45 &
  1,791 &
  262 &
  844 \\
 &
  \textsc{ocr+images} &
  6.46 &
  14.18 &
  2.30 &
  4.06 &
  7.82 &
  1,791 &
  1,455 &
  782 \\
 &
  \textbf{\textsc{ocr+pageN}} &
  6.54 &
  14.56 &
  2.30 &
  2.73 &
  9.53 &
  781 &
  1,317 &
  948 \\
 &
  \textbf{\textsc{ocr+page1}} &
  \textbf{5.83} &
  13.24 &
  2.30 &
  2.58 &
  8.36 &
  781 &
  1,285 &
  835 \\ \bottomrule
\end{tabular}
\end{table}

\begin{table}[h]
\centering
\caption{Results on \bentham{}. Much of this table is reproduced from Table~\ref{tab:bentham}; but new columns with token and cost breakdowns have been added.}
\label{tab:bentham-rebuttal}
\small
\begin{tabular}{@{}rlcrrrrrrr@{}}
\toprule
\multirow{2}{*}{\textbf{}} &
  \multirow{2}{*}{\textbf{Method}} &
  \multicolumn{1}{l}{\multirow{2}{*}{\textbf{CER $\downarrow$}}} &
  \multicolumn{4}{c}{\textbf{Cost (\$/1k docs)}} &
  \multicolumn{3}{c}{\textbf{Tokens (/doc)}} \\
 &
   &
  \multicolumn{1}{l}{} &
  \multicolumn{1}{c}{\textbf{Total}} &
  \multicolumn{1}{c}{\textbf{OCR}} &
  \multicolumn{1}{c}{\textbf{In}} &
  \multicolumn{1}{c}{\textbf{Out}} &
  \multicolumn{1}{c}{\textbf{Image}} &
  \multicolumn{1}{c}{\textbf{In}} &
  \multicolumn{1}{c}{\textbf{Out}} \\ \midrule
\textsc{---} &
  \textsc{ocr} &
  11.18 &
  2.63 &
  2.63 &
  0.00 &
  0.00 &
  0 &
  0 &
  0 \\ \midrule
\multirow{6}{*}{\raisebox{0.0em}[0pt][0pt]{\rotatebox[origin=c]{90}{\textsc{gemma}}}} &
  \textsc{ocr} &
  10.75 &
  3.11 &
  2.63 &
  0.10 &
  0.39 &
  0 &
  1,367 &
  781 \\
 &
  \textsc{images} &
  15.60 &
  0.70 &
  0.00 &
  0.20 &
  0.50 &
  2,287 &
  635 &
  990 \\
 &
  \textsc{images-all-at-once} &
  25.06 &
  0.55 &
  0.00 &
  0.18 &
  0.37 &
  2,287 &
  262 &
  747 \\
 &
  \textsc{ocr+images} &
  9.98 &
  3.28 &
  2.63 &
  0.27 &
  0.39 &
  2,287 &
  1,509 &
  781 \\
 &
  \textbf{\textsc{ocr+pageN}} &
  11.02 &
  3.34 &
  2.63 &
  0.25 &
  0.47 &
  871 &
  1,269 &
  851 \\
 &
  \textbf{\textsc{ocr+page1}} &
  10.89 &
  3.20 &
  2.63 &
  0.15 &
  0.42 &
  871 &
  1,244 &
  847 \\ \midrule
\multirow{6}{*}{\raisebox{0.0em}[0pt][0pt]{\rotatebox[origin=c]{90}{\textsc{gpt-4o}}}} &
  \textsc{ocr} &
  9.97 &
  13.63 &
  2.63 &
  3.42 &
  7.58 &
  29 &
  1,338 &
  758 \\
 &
  \textsc{images} &
  9.87 &
  16.58 &
  0.00 &
  8.88 &
  7.69 &
  2,902 &
  619 &
  769 \\
 &
  \textsc{images-all-at-once} &
  10.18 &
  15.80 &
  0.00 &
  8.15 &
  7.65 &
  2,902 &
  252 &
  764 \\
 &
  \textsc{ocr+images} &
  9.35 &
  20.93 &
  2.63 &
  11.01 &
  7.30 &
  2,931 &
  1,472 &
  729 \\
 &
  \textbf{\textsc{ocr+pageN}} &
  8.87 &
  16.37 &
  2.63 &
  6.13 &
  7.61 &
  1,105 &
  1,213 &
  757 \\
 &
  \textbf{\textsc{ocr+page1}} &
  10.95 &
  16.29 &
  2.63 &
  5.97 &
  7.69 &
  1,105 &
  1,189 &
  769 \\ \midrule
\multirow{6}{*}{\raisebox{0.0em}[0pt][0pt]{\rotatebox[origin=c]{90}{\textsc{gemini}}}} &
  \textsc{ocr} &
  9.54 &
  12.03 &
  2.63 &
  1.71 &
  7.70 &
  0 &
  1,367 &
  769 \\
 &
  \textsc{images} &
  9.74 &
  11.64 &
  0.00 &
  3.65 &
  7.99 &
  2,287 &
  635 &
  798 \\
 &
  \textsc{images-all-at-once} &
  9.88 &
  11.70 &
  0.00 &
  3.19 &
  8.51 &
  2,287 &
  262 &
  851 \\
 &
  \textsc{ocr+images} &
  8.67 &
  15.32 &
  2.63 &
  4.75 &
  7.95 &
  2,287 &
  1,509 &
  794 \\
 &
  \textbf{\textsc{ocr+pageN}} &
  \textbf{8.48} &
  13.70 &
  2.63 &
  2.77 &
  8.30 &
  871 &
  1,269 &
  825 \\
 &
  \textbf{\textsc{ocr+page1}} &
  \textit{8.56} &
  13.55 &
  2.63 &
  2.64 &
  8.28 &
  871 &
  1,244 &
  827 \\ \bottomrule
\end{tabular}
\end{table}

\newpage

\subsection{Early validation experiments}
\label{sec:app:early-validation}
Tables \ref{tab:azure_iam},~\ref{tab:google_ocr_improvement_iam_multipage_minpages=02_split=0.50_seed=00_checked}~\&~\ref{tab:textract_improvement_iam_multipage_minpages=02_split=0.50_seed=00_checked} show results of early experiments with \textsc{gpt-4o} and \textsc{gpt-4o-mini} for earlier prompt iterations of our methods, on a multi-page validation split derived from the IAM Database. Each table uses a different baseline OCR engine: Azure, Google Cloud Vision and Amazon Textract. As Azure performs the best of the three, and is the cheapest (see Table~\ref{table:cost_ocr_engines}), we use it for final experiments in the main text. We also tested Tesseract, but found it to be completely incapable of producing any meaningful transcription of handwritten text.

\begin{table}[H]
\small
\caption{\iam{}: relative performance of transcription methods. Rel(ative) imp(rovement) is against the baseline OCR (Azure), and cost is for processing the entire dataset with the given method.}
\centering
\begin{tabular}{@{}rrccc@{}}
\toprule
Method & $\rightarrow$ MLLM & CER &  \makecell{Rel.\\Imp.} & \makecell{Cost\\(\$)} \\
\midrule
\midrule
\ocr{} & \textsc{-} & 0.036 & \cellcolor[RGB]{254,254,254}0.00 & 0.59 \\

\ocr{} & \textsc{gpt-4o-mini} & 0.032 & \cellcolor[RGB]{233,245,233}0.11 & 0.64 \\
\ocr{} & \textsc{gpt-4o} & 0.033 & \cellcolor[RGB]{239,248,239}0.08 & 1.42 \\

\ocr{}\pbp{} & \textsc{gpt-4o-mini} & 0.025 & \cellcolor[RGB]{195,229,196}0.31 & 0.65 \\
\ocr{}\pbp{} & \textsc{gpt-4o} & 0.029 & \cellcolor[RGB]{214,237,215}0.21 & 1.50 \\

\chosenpage{} & \textsc{gpt-4o-mini} & 0.029 & \cellcolor[RGB]{218,239,218}0.19 & 2.12 \\
\chosenpage{} & \textsc{gpt-4o} & 0.025 & \cellcolor[RGB]{199,231,200}0.29 & 2.24 \\

\vision{} & \textsc{gpt-4o} & 0.027 & \cellcolor[RGB]{208,235,209}0.24 & 2.32 \\
\vision{}\pbp{} & \textsc{gpt-4o} & \textbf{0.010} & \cellcolor[RGB]{115,195,117}0.73 & 2.43 \\

\allpages{} & \textsc{gpt-4o} & 0.027 & \cellcolor[RGB]{208,235,209}0.24 & 3.10 \\
\allpages{}\pbp{} & \textsc{gpt-4o} & \underline{0.011} & \cellcolor[RGB]{123,199,126}0.68 & 3.24 \\

\allocr{}\pbp{} & \textsc{gpt-4o-mini} & 0.020 & \cellcolor[RGB]{166,217,168}0.46 & 2.48 \\
\allocr{}\pbp{} & \textsc{gpt-4o} & 0.021 & \cellcolor[RGB]{172,219,173}0.43 & 4.07 \\
\midrule
\firstpage{} & \textsc{gpt-4o-mini} & \textit{0.015} & \cellcolor[RGB]{141,206,143}0.59 & 2.10 \\
\firstpage{} & \textsc{gpt-4o} & 0.027 & \cellcolor[RGB]{205,233,205}0.26 & 2.20 \\
\bottomrule
\end{tabular}
\label{tab:azure_iam}
\end{table}

\begin{table}[H]
\small
\caption{Relative performance of MLLMs and prompting strategies compared to the baseline \textbf{Google} OCR engine on the \texttt{IAM} dataset.}
\centering
\begin{tabular}{@{}rrccc@{}}
\toprule
Method & $\rightarrow$ MLLM & CER &  \makecell{Rel.\\Imp.} & \makecell{Cost\\(\$)} \\
\midrule
\midrule
\ocr{} & \textsc{-} & 0.095 & \cellcolor[RGB]{254,254,254}0.00 & 0.89 \\
\ocr{} & \textsc{gpt-4o-mini} & 0.074 & \cellcolor[RGB]{219,239,219}0.23 & 0.94 \\
\ocr{}\pbp{} & \textsc{gpt-4o-mini} & 0.071 & \cellcolor[RGB]{214,237,215}0.26 & 0.95 \\
\ocr{} & \textsc{gpt-4o} & 0.064 & \cellcolor[RGB]{204,233,204}0.33 & 1.73 \\
\ocr{}\pbp{} & \textsc{gpt-4o} & 0.064 & \cellcolor[RGB]{204,233,204}0.33 & 1.81 \\
\vision{} & \textsc{gpt-4o} & 0.027 & \cellcolor[RGB]{144,208,146}0.71 & 2.32 \\
\firstpage{} & \textsc{gpt-4o-mini} & 0.047 & \cellcolor[RGB]{175,221,176}0.51 & 2.40 \\
\chosenpage{} & \textsc{gpt-4o-mini} & 0.060 & \cellcolor[RGB]{197,230,198}0.37 & 2.40 \\
\vision{}\pbp{} & \textsc{gpt-4o} & \textbf{0.010} & \cellcolor[RGB]{115,195,117}0.90 & 2.43 \\
\allocr{}\pbp{} & \textsc{gpt-4o-mini} & 0.020 & \cellcolor[RGB]{130,202,132}0.80 & 2.48 \\
\firstpage{} & \textsc{gpt-4o} & 0.042 & \cellcolor[RGB]{167,217,169}0.56 & 2.50 \\
\chosenpage{} & \textsc{gpt-4o} & 0.044 & \cellcolor[RGB]{171,219,172}0.54 & 2.53 \\
\allpages{} & \textsc{gpt-4o} & 0.035 & \cellcolor[RGB]{156,213,158}0.63 & 3.39 \\
\allpages{}\pbp{} & \textsc{gpt-4o} & \textit{0.019} & \cellcolor[RGB]{130,202,132}0.80 & 3.54 \\
\allocr{}\pbp{} & \textsc{gpt-4o} & 0.021 & \cellcolor[RGB]{133,203,135}0.78 & 4.07 \\\bottomrule
\end{tabular}
\label{tab:google_ocr_improvement_iam_multipage_minpages=02_split=0.50_seed=00_checked}
\end{table}

\begin{table}[H]
\small
\caption{Relative performance of MLLMs and prompting strategies compared to the baseline \textbf{Textract} OCR engine on the \texttt{IAM} dataset.}
\centering
\begin{tabular}{@{}rrccc@{}}
\toprule
Method & $\rightarrow$ MLLM & CER &  \makecell{Rel.\\Imp.} & \makecell{Cost\\(\$)} \\
\midrule
\midrule
\ocr{} & \textsc{-} & 0.050 & \cellcolor[RGB]{254,254,254}0.00 & 0.89 \\
\ocr{} & \textsc{gpt-4o-mini} & 0.051 & \cellcolor[RGB]{254,249,248}-0.03 & 0.94 \\
\ocr{}\pbp{} & \textsc{gpt-4o-mini} & 0.045 & \cellcolor[RGB]{238,247,238}0.10 & 0.95 \\
\ocr{} & \textsc{gpt-4o} & 0.046 & \cellcolor[RGB]{241,249,241}0.08 & 1.71 \\
\ocr{}\pbp{} & \textsc{gpt-4o} & 0.041 & \cellcolor[RGB]{223,241,224}0.18 & 1.79 \\
\vision{} & \textsc{gpt-4o} & 0.027 & \cellcolor[RGB]{176,221,177}0.45 & 2.32 \\
\firstpage{} & \textsc{gpt-4o-mini} & 0.027 & \cellcolor[RGB]{176,221,177}0.45 & 2.40 \\
\chosenpage{} & \textsc{gpt-4o-mini} & 0.046 & \cellcolor[RGB]{241,249,241}0.08 & 2.41 \\
\vision{}\pbp{} & \textsc{gpt-4o} & \textbf{0.010} & \cellcolor[RGB]{115,195,117}0.81 & 2.43 \\
\allocr{}\pbp{} & \textsc{gpt-4o-mini} & 0.020 & \cellcolor[RGB]{149,210,151}0.61 & 2.48 \\
\firstpage{} & \textsc{gpt-4o} & 0.030 & \cellcolor[RGB]{185,225,186}0.40 & 2.50 \\
\chosenpage{} & \textsc{gpt-4o} & 0.027 & \cellcolor[RGB]{175,221,176}0.46 & 2.54 \\
\allpages{} & \textsc{gpt-4o} & 0.029 & \cellcolor[RGB]{182,224,183}0.42 & 3.39 \\
\allpages{}\pbp{} & \textsc{gpt-4o} & \textit{0.011} & \cellcolor[RGB]{119,197,121}0.78 & 3.54 \\
\allocr{}\pbp{} & \textsc{gpt-4o} & 0.021 & \cellcolor[RGB]{152,211,154}0.59 & 4.07 \\\bottomrule
\end{tabular}
\label{tab:textract_improvement_iam_multipage_minpages=02_split=0.50_seed=00_checked}
\end{table}

\end{document}